\documentclass[10pt,journal,compsoc]{IEEEtran}

\ifCLASSOPTIONcompsoc
  \usepackage[nocompress]{cite}
\else
  \usepackage{cite}
\fi

\ifCLASSINFOpdf
\else
\fi
\usepackage{graphicx}
\usepackage{booktabs}  
\usepackage{amsmath}
\usepackage{multirow}
\usepackage{amssymb}
\usepackage{amsfonts} 
\usepackage{xcolor}
\usepackage{comment}
\usepackage{enumitem}
\usepackage[switch, mathlines]{lineno}
\usepackage{ulem}

\usepackage[ruled]{algorithm2e}

\newcommand\ourmodel{\textcolor{black}{P$^3$Net}}

\newcommand\XP[1]{\textcolor{black}{#1}}

\newcommand\HUI[1]{\textcolor{black}{#1}}

\begin{document}

\title{Semi-supervised Counting via Pixel-by-pixel Density Distribution Modelling}

\author{Hui Lin,
        Zhiheng Ma,
        Rongrong Ji,~\IEEEmembership{Senior Member,~IEEE,} 
        Yaowei Wang, 
        Zhou Su,~\IEEEmembership{Senior Member,~IEEE,} 
        Xiaopeng Hong\IEEEauthorrefmark{2},~\IEEEmembership{Senior Member,~IEEE,} 
        Deyu  Meng
\IEEEcompsocitemizethanks{\IEEEcompsocthanksitem Hui Lin, Zhou Su, and Deyu Meng are with Xi'an Jiaotong University. Zhiheng Ma is with Shenzhen Institute of Advanced Technology Chinese Academy of Science. Rongrong Ji is with Xiamen University. Yaowei Wang is with Peng Cheng Laboratory. Xiaopeng Hong is with the Faculty of Computing, Harbin Institute of Technology. E-mail: \{linhuixjtu@gmail.com / hongxiaopeng@ieee.org\}. 
\IEEEcompsocthanksitem {$^\dagger$Corresponding author: Xiaopeng Hong.}
}}

\markboth{Technical Report, ~2024}{}

\IEEEtitleabstractindextext{%
\begin{abstract}
This paper focuses on semi-supervised crowd counting, where only a small portion of the training data are labeled. We formulate the pixel-wise density value to regress as a probability distribution, instead of a single deterministic value. 
On this basis, we propose a semi-supervised crowd counting model. Firstly, we design a pixel-wise distribution matching loss to measure the differences in the pixel-wise density distributions between the prediction and the ground-truth; Secondly, we enhance the transformer decoder by using \emph{density tokens} to {specialize the forwards of decoders w.r.t. different density intervals;} Thirdly, \XP{we design the  \emph{interleaving consistency} self-supervised  learning mechanism to learn from unlabeled data efficiently.}
Extensive experiments on four datasets are performed to show that our method clearly outperforms the competitors by a large margin under various labeled ratio settings. \emph{\XP{Code will be released at https://github.com/LoraLinH/Semi-supervised-Counting-via-Pixel-by-pixel-Density-Distribution-Modelling.}}
\end{abstract}

}

\maketitle

\IEEEdisplaynontitleabstractindextext
\IEEEpeerreviewmaketitle

\IEEEraisesectionheading{\section{Introduction}\label{sec:introduction}}
Crowd counting~\cite{zhang2016single, cao2018scale, ma2019bayesian} is becoming increasingly important in computer vision. It has wide applications such as congestion estimation and crowd management. {A lot of fully-supervised crowd counting models have been proposed, which require a large number of labeled data to train an accurate and stable model. However, considering the density of the crowd, it is laborious and time-consuming to annotate the center of each person's head in a dataset of all dense crowd images. To alleviate the requirement for large amounts of labeled data,}
this paper focuses on \emph{semi-supervised counting} where only a small portion of training data are labeled~\cite{liu2018leveraging}.

Traditional semi-supervised counting methods target density regression and then leverage self-supervised criteria~\cite{liu2018leveraging, liu2019exploiting} or pseudo-label generation~\cite{sindagi2020learning, meng2021spatial} to exploit supervision signals under unlabeled data. These methods are designed to directly generate density maps,
where each pixel is associated with a definite value. However, it is still extremely difficult to learn {a good model} due to the uncertainty of pixel labels.
Firstly, there are commonly erroneous head locations in the annotations~\cite{wan2020modeling,bai2020adaptive}; Secondly, {the pseudo labels for unlabeled training data assigned by the models are pervasively noisy.}

To address these challenges, 
we propose a new semi-supervised counting model, termed by
the Pixel-by-Pixel Probability distribution modelling Network (\ourmodel{}). {Unlike traditional methods which generate a deterministic pixel density value, we 
model the targeted density value of a pixel as a probability distribution.} 
On this premise, \XP{we contribute to semi-supervised counting in four ways.}

\begin{itemize}[noitemsep,topsep=0pt,parsep=2pt,partopsep=0pt,leftmargin=1em]

    \item  \XP{We propose a Pixel-wise probabilistic Distribution (PDM) loss to match the distributions of the predicted density values and the targeted ones pixel by pixel.} The PDM loss, designed in line with the Cumulative Distribution Function distance, measures the cumulative gap between the predicted distribution and the ground-truth one along the density (interval) dimension. \XP{By  modeling the density intervals probabilistically, our method responds well to the uncertainty in the labels. It thus surpasses traditional methods that regard the \XP{density values} as deterministic.} 

    \item {We \XP{incorporate the transformer decoder structure with a} density-token scheme to modulate the features and generate high-quality density maps.  A density token encodes the semantic information of a specific density interval. In prediction, these density-specific tokens specialize the forwards of the transformer decoder  with respect to the corresponding density intervals.}

    \item \XP{We create two discrete representations of the pixel-wise density probability function and shift one to be interleaved, which are modelled by a dual-branch network structure. Then we propose an inter-branch Expectation Consistency Regularization term to reconcile the expectation of the predictions made by the two branches.}

    \item {We set up new state-of-the-art performance for semi-supervised crowd counting on four challenging crowd counting datasets, i.e. UCF-QNRF~\cite{idrees2018composition}, JHU-Crowd++~\cite{sindagi2020jhu}, ShanghaiTech A and B~\cite{zhang2016single}. Our method outperforms previous state-of-the-art methods under all three  settings of labeled ratio.}
\end{itemize}

\section{Related Works}
\subsubsection{Fully-supervised Crowd Counting.}
Early methods tackle the crowd counting problem by exhaustively detecting every individual in the image~\cite{liu2019point}~\cite{liu2018decidenet}. However, these methods are sensitive to occlusion and require additional annotations like bounding boxes. With the introduction of density map~\cite{lempitsky2010learning}, numerous CNN-based approaches are proposed to treat crowd counting as a regression problem.
MCNN~\cite{zhang2016single} employs multi-column network with adaptive Gaussian kernels to extract multi-scale features. Switch-CNN~\cite{babu2017switching} handles the variation of crowd density by training a switch classifier to relay a patch to a particular regressor.
SANet~\cite{cao2018scale} proposes a local pattern consistency loss with scale aggregation modules and transposed convolutions. 
CSRnet~\cite{li2018csrnet} uses dilated kernels to enlarge receptive fields and perform accurate count estimation of highly congested scenes. 
BL~\cite{ma2019bayesian} introduces the loss under Bayesian assumption to calculate the expected count of pixels. Furthermore, methods based on multi-scale mechanisms~\cite{zeng2017multi,sindagi2019multi,ma2020learning}, perspective estimation~\cite{shi2019revisiting,yan2019perspective} and optimal transport~\cite{wang2020distribution, ma2021learning, lin2021direct} are proposed to overcome  \XP{the problem caused by large scale variations in crowd images.}

Recently, to alleviate the problem of inaccurate annotations in crowd counting, \XP{a few} studies begin to find solutions by quantizing the count values within each local patch into a set of intervals and learning to classify. S-DCNet proposes a classifier and a division decider to decide which sub-region should be divided and transform the open-set counting into a closed-set problem~\cite{xiong2019open}. A block-wise count level classification framework is \XP{introduced} to address the problems of inaccurately generated regression targets and serious sample imbalances~\cite{liu2019counting}. The work~\cite{liu2020adaptive} proposes an adaptive mixture regression framework and leverages on local counting map to reduce the inconsistency between training targets and evaluation criteria. UEPNet~\cite{wang2021uniformity} \XP{uses} two criteria to minimize the prediction risk and discretization errors of a classification model. Our method is distinct from most existing approaches. We revisit the paradigm of density classification from the perspective of semi-supervised learning and \XP{reveal} that the interleaving quantization interval has a natural consistency self-supervision mechanism.

\subsubsection{\XP{Semi and Weakly}-Supervised Crowd Counting.}
As labeling crowd images is expensive,
recent studies gradually focus on semi- and weakly-supervised crowd counting. 
For \emph{semi-supervised counting}, various techniques have been proposed to make efficient use of unlabeled data. For instance, L2R~\cite{liu2018leveraging} introduces an auxiliary sorting task by learning containment relationships to exploit unlabeled images. Sindagi et al.~\cite{sindagi2020learning} propose a Gaussian Process-based learning mechanism to generate pseudo-labels for unlabeled data. Zhao et al.~\cite{zhao2020active} introduce an active learning framework aimed at reducing the labor-intensive label work. IRAST~\cite{liu2020semi} utilize a set of surrogate binary segmentation tasks to exploit the underlying constraints from unlabeled data. Meng et al.~\cite{meng2021spatial} propose a spatial uncertainty-aware teacher-student framework to alleviate uncertainty from labels. Lin et al. \cite{lin2022semi} propose an agency-guided counting approach that establishes correlations between foreground and background features in all images. \HUI{Recently, the work~\cite{lin2023optimal} proposes the OT-M algorithm to generate hard pseudo-labels (point maps) rather than soft ones (density maps) for the task, providing stronger supervision. Furthermore, MTCP~\cite{zhu2023multi} employs three multi-task branches, including density regression as the primary task, and binary segmentation and confidence prediction as auxiliary tasks, to perform credible pseudo-label learning, reducing noise in pseudo-labels for semi-supervised learning. In contrast, we consider semi-supervised crowd counting as a quantitative density-interval distribution matching problem and provide a self-supervised scheme via a consistency-constrained dual-branch structure.} 

Moreover, there are also relevant studies about \emph{weakly supervised counting}~\cite{yang2020weakly, lei2021towards, sindagi2019ha}, which primarily focus on learning from coarse annotations such as image-level labels or total counts. \HUI{Additionally, other works~\cite{han2020focus, liu2022leveraging, wang2019learning} concentrate on \emph{domain adaptation} to address the problem of data scarcity in crowd counting.}

\subsubsection{Vision Transformer.}
Vision Transformer (ViT)~\cite{dosovitskiy2020image} introduces the Transformer networks~\cite{vaswani2017attention} to image recognition.
\XP{Transformers further advances various tasks, such as object detection~\cite{carion2020end, zhu2020deformable, zheng2020end, sun2021rethinking}, instance or semantic segmentation~\cite{zheng2021rethinking, wang2021end, strudel2021segmenter, cheng2021per}, and object tracking~\cite{chen2021transformer, wang2021transformer, sun2020transtrack}.}
Lately, 
\XP{the works~\cite{lin2022boosting, wei2021scene,liang2021transcrowd}  use} the transformer encoder with self-attention to refine the image feature for crowd counting, \XP{whilst our method leverages the decoder with cross-attention to learn the density classification tokens.}
\begin{figure*}[t] 
\begin{center}
    \includegraphics[width=0.98\textwidth]{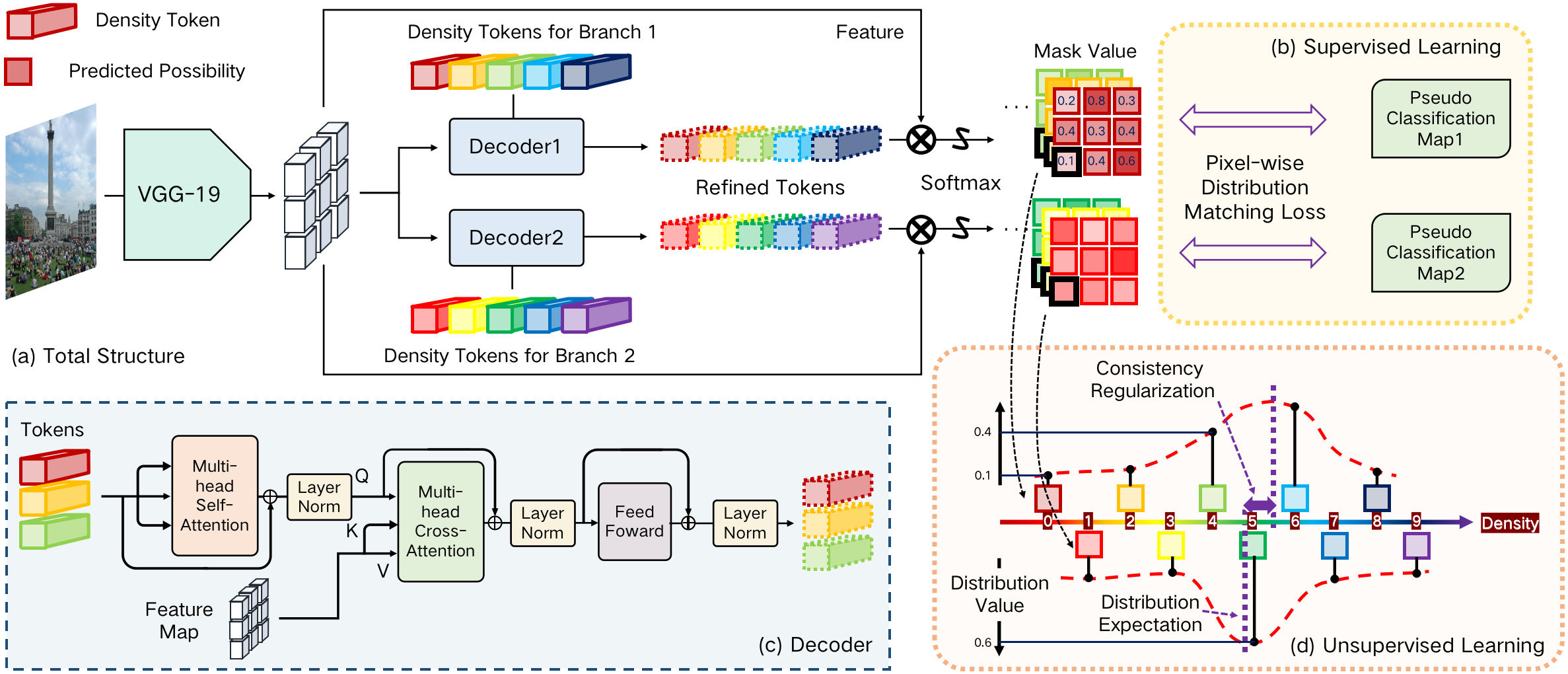}
\end{center}
\caption{The structure of \ourmodel. (a) The dual-branch structure with density tokens to predict interleaving density category. Different token colors represent the specified different density intervals. The softmax operation targets the category while each predicted distribution map represents the segmentation map learned by this specific token and the corresponding density category. (b) The structure of the decoder. (c) The inter-branch Expectation Consistency Regularization for self-supervised learning. \XP{(d)} The horizontal axis stands for the density values, with the attached squares for the discrete density intervals corresponding to the tokens. The vertical axis is the normalized distribution value of each category for that pixel.}
\label{fig:structure}
\end{figure*}

\section{Counting via Pixel-by-Pixel Probabilistic Distribution Modelling}
\label{sec:classification}

In this section, we first describe the setting of semi-supervised crowd counting and then explain the rationale of adopting the probability distribution to represent the crowd density.


Formally, we have a labeled dataset $\mathcal{X}$ consisting of images with point annotated ground truth and an unlabeled dataset $\mathcal{U}$ consisting of only crowd images. In  semi-supervised crowd counting, $\mathcal{U}$ \XP{usually} contains much more images than $\mathcal{X}$ for training a counting model, i.e., $\left | \mathcal{U} \right | \gg \left | \mathcal{X} \right |$.

Previous methods have utilized self-supervised criteria~\cite{liu2018leveraging, liu2019exploiting} or pseudo-label generation~\cite{sindagi2020learning, meng2021spatial} to exploit supervision signals under unlabeled data. These methods rely on the density prediction \XP{pipeline as the} most traditional supervised methods~\cite{zhang2016single, ma2019bayesian, lin2022boosting}. 
However, when only partial labels are available for training the model, the obtained density maps are likely to be noisy. It becomes increasingly challenging to predict a deterministic, accurate density value for each single pixel or small patches.
\XP{To solve this problem, we model the targeted density value of a pixel as a probability distribution, instead of a deterministic single value. The predicted density value $d$ is then given by the expectation as follows.}
\begin{equation}\label{eq:contdistribution}
    d = \int_0^{+\infty} p(x) x dx,
\end{equation}
where $x$ is the probable density value ranged in $[0, +\infty)$. The conventional prediction way, which can be represented as the Dirac delta, $p(x)=\delta (x-d)$,
is a special case of Eq.~\ref{eq:contdistribution}.
\XP{This approach is fragile when there is uncertainty and noise. Instead, by Eq.~\ref{eq:contdistribution}, we  revert to a general distribution function $p(x)$ without introducing any prior about the distribution such as Dirac.}

{To find a numerical \XP{form, to which deep learning models can be applied}, we further discretize Eq.~\ref{eq:contdistribution}:}
\begin{equation}\label{eq:distribution}
    d=\sum_{j=1}^{C}P\left ( x_j \right )x_j.
\end{equation}
\XP{The set $\left \{ x_1, x_2, \cdots, x_C \right \}$ are the discrete representations of the density intervals, which are obtained}
by quantizing the continuous density range $[0,+\infty)$ into $C$ mutually exclusive discretized intervals $[0,b_1),\  [b_1,b_2),\ ...,\ [b_{C-1},+\infty)$,
where $b_1,...,b_{C-1}$ are the ascending interval borders.
$P(x)$ is the discrete distribution function, which can be easily implemented through a softmax function and is consistent with the convolutional neural network. As a result, in this work, we transform the regression problem into a density interval classification problem, \emph{i.e.} from predicting an exact count to choosing a pre-defined density interval, in order to build more reliable prediction signals for semi-supervised counting.

\XP{On this basis, we propose our Pixel-by-Pixel Probability distribution modelling Network (\ourmodel) for semi-supervised learning.} \ourmodel{} is \XP{composed} of three modules to enhance the classification paradigm to semi-supervised crowd counting. First, we propose a Pixel-wise Distribution Matching (PDM) loss to meet the needs of effectively matching the distributions between the prediction and the label. After that, we introduce a transformer decoder with proposed density tokens to learn and preserve density information from different \XP{density intervals}. \XP{And finally, we design a dual-branch structure and propose a corresponding self-supervision mechanism} for semi-supervised learning.

\subsection{Pixel-wise Distribution Matching Loss}

In this section, we detail the \XP{proposed PDM loss and the corresponding supervision between predicted distribution and the ground-truth.}

To punish the difference between predicted distributions and ground truth, we first generate the training label $Y \in \{0,1\}^{N \times C}$ for the dual-branch from annotated points. We perform a 2-D Gaussian smoothing on these points, and then calculate the expected density value of each pixel. Each row in the label $\mathbf{y} \in \{0,1\}^{C}$ is in the form of one hot distribution and the category where the value equals $1$ represents the specific interval that the density of this certain pixel falls into.

We match the predicted distribution to the ground-truth distribution by minimizing the divergence between them. On this basis, we define the Pixel-wise Distribution Matching \XP{(PDM) loss based on} the Cumulative Distribution Function (CDF)~\cite{su2015distances, chun2000uncertainty} distance to act as the measuring function. 

Given $\mathbf{p}$ and $\mathbf{y}$ as \XP{the prediction and ground-truth label for a certain pixel respectively, and $\mathcal{G}(\mathbf{y},j) = \sum_{i=1}^j y_i$ as the cumulative distribution function, the loss can be calculated by}
\begin{equation}
\label{eq:pdm}
\begin{aligned}
    &\mathcal{L}_P =  \sum_{\mathbf{y},\mathbf{p}} (\sum_{j=1}^C (\mathcal{G}(\mathbf{y},j) - \mathcal{G}(\mathbf{p},j))^l)^{\frac{1}{l}},
\end{aligned}
\end{equation}
where $l$ is the level of Euclidean norm. The PDM loss measures the cumulative gap between the predicted distribution and the ground truth along the density dimension. It penalizes the distributions that are deviated. Note that the loss strictly equals to the Wasserstein loss under L1 norm ($l=1$)~\cite{kolouri2018sliced, de20211}. The one with L2 norm ($l=2$) is a satisfactory approximation and has stronger supervision on the parts with larger gaps between distributions, which exactly benefits the optimization of the loss more.

\textbf{The Rationality of PDM Loss.}
\XP{We provide an example to illustrate the advantages of our loss function. Suppose there are four intervals and we have an instance with the label of [0,1,0,0]. Given two predicted outputs A: [0.2,0.3,0.5,0] and B: [0.2,0.3,0,0.5], clearly A gets a more compact, single-mode output which shall be considered better than B.  However, the loss values of A and B are the same in terms of the Cross Entropy (0.36) and Mean Square Error (0.78), they can not be distinguished. In contrast, in terms of our PDM loss, the cumulative forms to calculate Eq.~\ref{eq:pdm} for A and B are [0.2,0.5,1.0,1.0] and [0.2,0.5,0.5,1.0] respectively and the corresponding loss values are 0.29 and 0.54. As a result, the two can be well differentiated.}

\textbf{\XP{Differences from DM-Count}.}
DM-Count~\cite{wang2020distribution} is an insightful optimal transport based counting approach to match the probability distributions of occurrence over the \emph{spatial} domain. In contrast, the proposed PDM loss matches the pixel-wise probability distributions over the \emph{density intervals}. \XP{Hence, the domains where \emph{optimal transport} performs by the two methods are distinctly different.}

\subsection{\XP{Transformer specialization}}

{Next, we introduce a set of density tokens to specialize the forwards of the transformer decoder with respect to the corresponding density intervals.}
{The density tokens are learnable embeddings with different density information, \XP{which} are fed to interact with the input extracted feature vectors to instruct the model prediction. Each token is endowed with unique semantic information and acts as an indicator of a density interval.} In other words, the \emph{density tokens} are prototypes corresponding to different density intervals. Specifically, we set $b_1$ {to} a small value and {treat} the token assigned to the first interval $\left [0,b_1 \right )$ as the background token. It is responsible for learning the features in areas without crowd in the image. {We denote $T \in \mathbb{R}^{C \times Z}$ as a matrix capsuling all $C$ tokens where $Z$ is the dimension of both the features and tokens}.

Then we use the transformer decoder break the limitation of local \XP{convolutional kernels}, correlating similar density information from various regions inside an image. 
The tokens are firstly processed by a multi-head self-attention module and a normalization layer. The relationships between tokens and the whole feature map are computed through cross attention:
\begin{equation}\label{eq:attn}
    \mathcal{C}(T, F) = \mathcal{S}(\frac{(TW^Q)(FW^K)^\mathsf{T}}{\sqrt{Z}})(FW^V).
\end{equation}
\XP{$F \in \mathbb{R}^{N \times Z}$ is the matrix of the input features, where $N$ is the pixel or patch number.} $\mathcal{S}$ is the softmax function, and $W^Q,W^K,W^V \in \mathbb{R}^{Z \times Z}$ are weight matrices for projections. \XP{Afterwards, we get the \emph{refined tokens} $\tilde{T}$, after processing further by a layer normalization and a feed-forward network, as illustrated in Figure~\ref{fig:structure} (c).}

Note that in Equation~\ref{eq:attn}, through the inner product of the two vectors, the cross attention learns which regions in the feature map that each category token should focus on. Inspired by this idea, \XP{in the forward pass, we leverage the density-interval-specialized token through a softmax activation to modulate the input patch features for predicting the final probabilities:}
\begin{equation}\label{eq:classification}
\begin{aligned}
     O = &\mathcal{S}(\tilde{T} \cdot F^\mathsf{T}),
\end{aligned}
\end{equation}
where the softmax operation is performed along the category dimension.
The predicted matrix $O \in \mathbb{R}^{C \times N}$ denotes the $C$ \XP{predicted distribution maps, each of which} represents the region distribution of the corresponding density interval in the whole image of $N$ patches. \XP{By Eq.~\ref{eq:classification}, we measure the similarity between the region features and refined density tokens, modulate the regional features and output the predicted density-interval distribution.}

\XP{As the idea of proposed density tokens is a natural extension in semantics of that of the \emph{query tokens} in the transformer, it can be optimized through the training pipeline of transformer using back propagation. Note that, only the original density tokens are restored, while the tokens refined adapted to the input regional features are not retained. As a result, the final density tokens are the hyper-parameters shared by all inputs in the reference stage.}

\begin{figure}[t]
\begin{center}
    \includegraphics[width=0.5\textwidth]{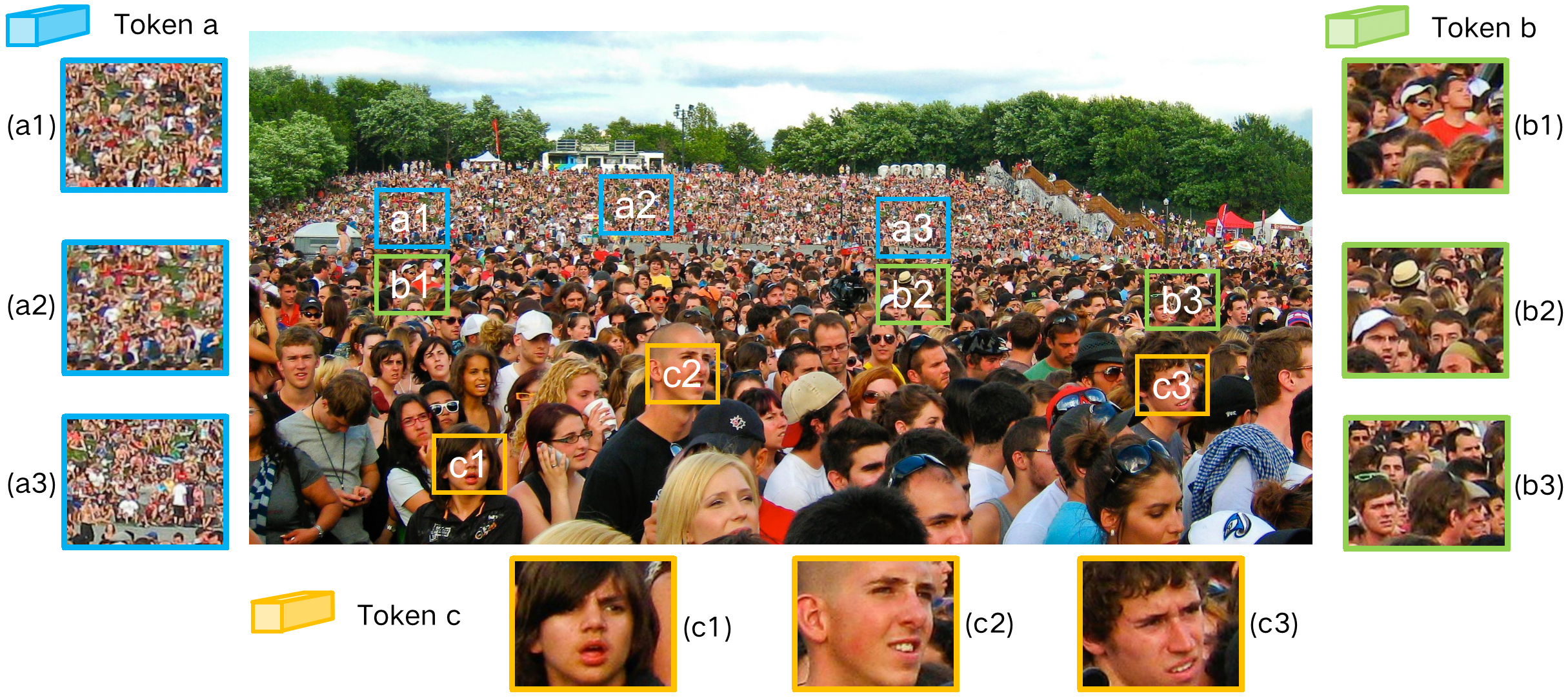}
\end{center}
\caption{Similar Regions of the same density levels \XP{exists within an image. We use a density token to specify a density interval and group} the regions of that level.}
\label{fig:rationale}
\end{figure}

\noindent \textbf{The Rationality of \XP{Tokens}} arises from the \XP{observation} that similar regions with same density intervals can be mined within an image. 
An example is shown in Figure~\ref{fig:rationale}, for a specific density interval like (a1), we can easily find other similar regions (a2/a3) all over the image. \XP{The density tokens (a/b/c) play a role of grouping different regions with the same density levels. During learning, the tokens are connected to the discrete representation of density probability distribution and finally with clear semantic associations. During inference, by using the tokens traversally, we specialize each forward of our decoder module with respect to a particular density interval distribution in turn.}

\noindent \textbf{Differences from randomly-\XP{initialized} queries.} Transformer decoder usually uses randomly initialized queries as the input. Here, we use DETR~\cite{carion2020end} as an example. In DETR, there is no clear distinction among the representative semantics of different queries. Thus in the training stage, association methods like Hungarian algorithm are required in every iteration to match queries with objects. Instead, we explicitly associate an exclusive density interval to each query throughout the model's lifetime. Thus we can generate tailored tokens with clear semantics.

\subsection{Inter-branch Expectation Consistency Based On Dual-Branch Interleaving Structure }

Although modelling the predicted density value as a discrete probabilistic distribution leads to more credible and less noisy prediction \XP{generally}, when the value falls near the boundary, \XP{noise in the output density values can easily lead to incorrect quantization, and further corrupt the classification results.}
Meanwhile, when converting the predicted interval category into density, the pre-defined discrete representation $x$ will inevitably have \XP{a quantization {gap}}.

To alleviate these limitations, we use a dual-branch structure with interleaving density representation, which consists of two parallel classification tasks with overlapping count intervals. The density which falls near the interval border of the first branch is more likely to be classified easily in another branch and meanwhile reducing the conversion gap without increasing the number of classification intervals. We study the influence of using different dual structures and show their details in the appendix.

\XP{The dual-branch structure with interleaving density representation has been used to address the inaccurate ground truth~\cite{wang2021uniformity}. To further accommodate it with semi-supervised counting, we make a step forward in this direction from two perspectives. Firstly, we associate the output of the network with the pixel-by-pixel probabilistic distribution and introduce a weighted, soft quantization level assignment mechanism.}
More specifically, during the inference stage, the work~\cite{wang2021uniformity} selects the category with maximum predicted value for each pixel or patch, and directly converts it to the corresponding representation value. Instead, we keep the distributions of predicted possibilities and leverage on the expectation to alleviate the conversion error. As a result, rather than simply averaging the predicted densities of two branches, we give each a certainty weight, which can be represented by the maximum classified possibility. When a branch predicts a large possibility for a certain category rather than similar values for multi categories, the branch has higher confidence about that prediction, thus we increase the proportion of it in the final prediction. \XP{Secondly, the interleaving two-branch structure provides a natural self-supervising mechanism that allows for imposing the consistency constraint between the two branches. On top of this constraint, we design an interleaving consistency regularization term which penalizes the deviation between the output expectations of the two-branches, to provide rich supervised signals in the absence of labels.}

Specifically, \XP{we denote by two $C$-dimensional vector $\mathbf{p}, \mathbf{q} \in \mathbb{R}^C$} the predicted classified possibilities of the dual branches for a certain pixel or patch. The vectors  satisfy that \XP{$\left \| \mathbf{p} \right \|_1 = \left \| \mathbf{q} \right \|_1 = 1$} and their elements are in the range of $[0,1]$ .
Thus the final density can be expressed by
\begin{equation}\label{eq:confidence}
\begin{aligned}
    \XP{d = \omega\ \mathbf{p} \cdot {\mathbf{v}}^\mathsf{T}_1 + (1-\omega)\ \mathbf{q} \cdot {\mathbf{v}}^\mathsf{T}_2},
\end{aligned}
\end{equation}
where the weight {$\omega = \left \| \mathbf{p}  \right \|_\infty  / (\left \| \mathbf{p}  \right \|_\infty + \left \| \mathbf{q}  \right \|_\infty )$ and $\left \| \cdot  \right \|_\infty$ is the  vector maximum norm}. \XP{We extend the proposed network to fit the dual-branch structure, where two different decoders are adopted and the density tokens are split into two interleaved sets, as shown in Figure~\ref{fig:structure}.}

On the basis, a self-supervised learning scheme is \XP{designed to leverage the unlabeled data for refining the model, where} the expectations of classified probability distribution on two branches tend to be consistent. \XP{We based on this constraint to construct the inter-branch Expectation Consistency Regularization (ECR) term. Moreover,} to prevent the regularization term from being negatively affected by the wrongly predicted probability distribution, we impose a selection mechanism to only consider the patches which are predicted with high certainty.

The mechanism is based on a dynamic \XP{pixel-wise} mask $\mathcal{E} \in \mathbb{R}^{N}$ which elements are in the range of $[0,1]$ to select or weigh the regions for supervision. Given $O_1, O_2$ as the predicted probability matrices by the two branches, the self-supervised ECR is defined as
\begin{equation}
\label{eq:consistency}
    \mathcal{L}_E = \Vert \mathcal{E} \circ \mathcal{R} \Vert^2_2,
\end{equation}
where \XP{ $\mathcal{R} = \mathbf{v}_1O_1 - \mathbf{v}_2O_2$ is a vector reflecting the inconsistency between the density expectations by the two branches and} $\circ$ is \XP{the} element-wise multiplication.

Similar with Eq.~\ref{eq:confidence}, we regard the maximum possibility $\left \| \mathbf{p}  \right \|_\infty$ in each distribution as the confidence. If the distribution is even, the confidence will be low, indicating that the model cannot predict a certain class for that patch with high certainty. In this case, we shall reduce its importance or exclude this patch in back-propagation dynamically. For efficient computation, we \XP{binarize $\mathcal{E} \in \{0,1\}^{N}$}. Only when the both confidences of two branches are \XP{sufficiently high, regularization on that pixel is activated.}
Given the confidence threshold $\xi \in [0,1)$ 
and the boolean function $\tau(cond)$
which outputs $1$ when the condition is true and $0$ otherwise, the supervision mask is defined as:
\begin{equation}
\mathcal{E} = \tau(\mathbf{o}_1 > \xi)\ \& \ \tau(\mathbf{o}_2 > \xi),
\end{equation}
where \XP{$\mathbf{o}_1$ and $\mathbf{o}_2$ are $N$-dimensional vectors takeing the maximum values of $O_1$ and $O_2$ along the  interval} dimension respectively. Finally, the overall training loss is the combination of density aware loss using in labeled data and consistency regularization with the parameter $\lambda$ using in unlabeled data.
\begin{equation}
    \mathcal{L} = \mathcal{L}_P + \lambda \mathcal{L}_E.
\end{equation}

\noindent \textbf{The Rationality of the Regularization.}
A common \XP{issue} in self-supervised consistency regularization is the confirmation bias~\cite{tarvainen2017mean}, which indicates that the mistakes of the model will \XP{probably be accumulated during semi-supervised learning. We utilize the regularization term to alleviate this bias from the following aspects.} First, we select the most reliable instances for self-supervision by using the mask in Eq.~\ref{eq:consistency}. Second, our network adopts two independent decoders and respective density tokens. As shown by~\cite{ke2019dual}, learning independent models helps to address the performance bottleneck caused by model coupling. Thus the proposed regularization term is plausible. We also provide a detailed study in the experiments.

\section{Experiments}
\subsection{Implementation Details}
We conduct extensive experiments on five crowd counting benchmarks to verify the effectiveness of proposed \ourmodel{}. The datasets are described as follows:

\noindent \textbf{UCF-QNRF}~\cite{idrees2018composition} The dataset contains congested crowd images, which are crawled from Flickr, Web Search, and Hajj footage. It includes 1,535 high-resolution images with 1.25 million annotated points. There are 1,201 and 334 images in the training and testing sets respectively. 

\noindent \textbf{JHU-Crowd++~\cite{sindagi2020jhu}} 
The dataset includes 4,372 images with 1.51 million annotated points. There are 2,272 images used for training, 500 images for validation, and the rest 1,600 images used for testing. The crowd images are collected from several sources on the Internet using different keywords and typically chosen under various conditions and geographical locations.

\noindent \textbf{ShanghaiTech A~\cite{zhang2016single}} The dataset contains 482 crowd images with 244,167 annotated points. The images are randomly chosen from the Internet where the number of annotations in an image ranges from 33 to 3,139. The training set has 300 images, and the testing set has the remaining 182 images. 

\noindent \textbf{ShanghaiTech B~\cite{zhang2016single}} The dataset contains 716 crowd images, which are taken in the crowded street of Shanghai. The number of annotations in an image ranges from 9 to 578. The training set has 316 images, and the testing set has the remaining 400 images.

\noindent \textbf{NWPU-CROWD~\cite{wang2020nwpu}} contains 5,109 images with 2.13 million annotated points. There are 3,109 images in the training set, 500 images in the validation set, and the remaining 1,500 images in the testing set. The dataset has a large density range from 0 to 20,033 and contains various illumination scenes.

The network structure and the training details are summarized as follows.

\noindent \textbf{Network Details.}
VGG-19, which is pre-trained on ImageNet, is adopted as our CNN backbone to extract features. We use Adam algorithm~\cite{kingma2014adam} to optimize the model with the learning rate $10^{-5}$. The number of decoder layers is set as $4$. 
We set $C=25$ and {follow~\cite{wang2021uniformity}} to calculate the reasonable density intervals.
For the loss parameters, we set $\lambda = 0.01$ and $\xi = 0.5$.

\noindent \textbf{Training Details.}
We adopt horizontal flipping and random scaling of [0.7, 1.3] for each training image. The random crop with a size of 512 × 512 is implemented, and as some images in ShanghaiTech A contain smaller resolution, the crop size for this dataset reduces to 256 × 256. We limit the shorter side of each image within 2048 pixels in all datasets. The experiments are held on one GPU of RTX3080.

\subsection{Comparisons to the State of the Arts}
We evaluate P$^3$Net on these datasets and compare it with state-of-the-art semi-supervised methods, as shown in Table~\ref{tab:semi-performance}. P$^3$Net consistently demonstrates outstanding counting accuracy under different semi-supervised settings and across various datasets. Notably, in the setting of 40\% labeled ratio, our method sets new performance records across all datasets, consistently surpassing DACount~\cite{lin2022semi} and OT-M~\cite{lin2023optimal}, the recent state-of-the-art approaches in semi-supervised crowd counting. The excellent results demonstrate the effectiveness of our method in semi-supervised crowd counting.

\def\arraystretch{1.1}
\renewcommand{\tabcolsep}{12 pt}{
\begin{table*}[t!]
\footnotesize
	\begin{center}
		\begin{tabular}{c|c|cc|cc|cc|cc}
			\toprule[1.5pt]
			\multicolumn{1}{c}{\multirow{2}*{Methods}} & \multicolumn{1}{c}{Labeled} &  \multicolumn{2}{c}{UCF-QNRF} &  \multicolumn{2}{c}{JHU++} &
			\multicolumn{2}{c}{ShanghaiTech A} &
			\multicolumn{2}{c}{ShanghaiTech B} \\
			
			\multicolumn{1}{c}{} & \multicolumn{1}{c}{Percentage} & \multicolumn{1}{c}{MAE} &
			\multicolumn{1}{c}{MSE} & \multicolumn{1}{c}{MAE} &
			\multicolumn{1}{c}{MSE} &  \multicolumn{1}{c}{MAE} &
			\multicolumn{1}{c}{MSE} &  \multicolumn{1}{c}{MAE} &
			\multicolumn{1}{c}{MSE}\\
			\hline
			\hline
			MT~\cite{tarvainen2017mean} & 5\%  & 172.4 & 284.9 & 101.5 & 363.5 & 104.7 & 156.9 & 19.3 & 33.2\\
			
			L2R~\cite{liu2018leveraging}  & 5\% & 160.1 & 272.3 & 101.4 & 338.8 & 103.0 & 155.4 & 20.3 & 27.6 \\
			GP~\cite{sindagi2020learning}  & 5\% & 160.0 & 275.0 & - & -  & 102.0 & 172.0 & 15.7 & 27.9 \\
            DACount~\cite{lin2022semi} & 5\%  & 120.2 & 209.3 & 82.2 & \textbf{294.9} & 85.4 & 134.5 & 12.6 & 22.8 \\ 
            \HUI{OT-M~\cite{lin2023optimal}} & 5\%  & \HUI{118.4} & \HUI{195.4} & \HUI{82.7} & \HUI{304.5} & \HUI{\textbf{83.7}} & \HUI{133.3} & \HUI{12.6} & \HUI{\textbf{21.5}} \\ 
			\ourmodel{} (Ours)  & 5\% & \textbf{115.3} & \textbf{195.2} & \textbf{80.8} & 306.1  & 85.5 & \textbf{131.0} & \textbf{12.0} & 22.0  \\
			\midrule
			MT~\cite{tarvainen2017mean}  & 10\% & 156.1 & 145.5 & 250.3 & 90.2 & 319.3  & 94.5 & 15.6 & 24.5 \\
		    L2R~\cite{liu2018leveraging} & 10\% & 148.9 & 249.8 & 87.5 & 315.3 & 90.3 & 153.5 & 15.6 & 24.4 \\
		    AL-AC~\cite{zhao2020active} & 10\% & - & - & - & -  & 87.9 & 139.5 & 13.9 & 26.2\\
		    IRAST~\cite{liu2020semi} & 10\% & - & - & - & -  & 86.9 & 148.9 & 14.7 & 22.9\\
		    IRAST+SPN~\cite{liu2020semi} & 10\%  & - & - & - & - & 83.9 & 140.1 & - & - \\
            \HUI{MTCP~\cite{zhu2023multi}} & 10\%  & - & - & - & - & \HUI{81.3} & \HUI{130.5} & \HUI{14.5} & \HUI{22.3} \\
            DACount~\cite{lin2022semi} & 10\%  & 109.0 & 187.2 & 75.9 & 282.3 & 74.9 & \textbf{115.5} & 11.1 & 19.1 \\ 
            \HUI{OT-M~\cite{lin2023optimal}} & 10\%  & \HUI{113.1} & \HUI{186.7} & \HUI{73.0} & \HUI{\textbf{280.6}} & \HUI{80.1} & \HUI{118.5} & \HUI{10.8} & \HUI{18.2} \\    
		    \ourmodel{} (Ours)  & 10\% & \textbf{103.4} & \textbf{179.0} & \textbf{71.8} & 294.4 & \textbf{72.1} & 116.4 & \textbf{10.1} & \textbf{18.2}  \\
		    \midrule
			MT~\cite{tarvainen2017mean}  & 40\% & 147.2 & 249.6 & 121.5 & 388.9 & 88.2 & 151.1 & 15.9 & 25.7 \\
		    L2R~\cite{liu2018leveraging} & 40\% & 145.1 & 256.1 & 123.6 & 376.1 & 86.5 & 148.2 & 16.8 & 25.1 \\
		    GP~\cite{sindagi2020learning}  & 40\% & 136.0 & - & - & - & 89.0 & - & - & - \\
		    IRAST~\cite{liu2020semi} & 40\% & 138.9  & - & - & - & - & - & - & - \\
		    SUA~\cite{meng2021spatial} & 40\% & 130.3 & 226.3 & 80.7 & 290.8 & 68.5 & 121.9 & 14.1 & 20.6 \\
            DACount~\cite{lin2022semi} & 40\%  & 91.1 & 153.4 & 65.1 & 260.0 & 67.5 & 110.7 & 9.6 & 14.6 \\ 
            \HUI{OT-M~\cite{lin2023optimal}} & 40\%  & \HUI{100.6} & \HUI{167.6} & \HUI{72.1} & \HUI{272.0} & \HUI{70.7} & \HUI{114.5} & \HUI{8.1} & \HUI{13.1} \\          
		    \ourmodel{} (Ours)  & 40\% & \textbf{90.0} & \textbf{155.4} & \textbf{58.9} & \textbf{251.9}  & \textbf{63.0} & \textbf{100.9} & \textbf{7.1} & \textbf{12.0} \\
			\bottomrule[1.5pt]
		\end{tabular}
	\end{center}
\caption{Comparisons with the state of the arts semi-supervised counting methods on \XP{four datasets}. The best performance is shown in \textbf{bold}. The results of other methods under the $40\%$ labeled setting are referred to \cite{meng2021spatial} and all other results are from the original papers.}
\label{tab:semi-performance}
\end{table*}}

\subsection{Semi-supervised Counting Performance on NWPU}\label{sec:nwpu}

\renewcommand{\tabcolsep}{8 pt}{
\begin{table*}[thbp!]
\small
	\begin{center}
		\begin{tabular}{c|cc|cc|cc}
			\toprule[1.5pt]
			\multicolumn{1}{c|}{\multirow{2}*{Labeled Ratio}} &  \multicolumn{2}{|c|}{$5\%$} &  \multicolumn{2}{|c|}{$10\%$} &
			\multicolumn{2}{|c}{$40\%$}  \\
			
			\multicolumn{1}{c|}{} &  \multicolumn{1}{|c}{MAE} &
			\multicolumn{1}{c|}{MSE} & \multicolumn{1}{|c}{MAE} &
			\multicolumn{1}{c|}{MSE} &  \multicolumn{1}{|c}{MAE} &
			\multicolumn{1}{c}{MSE} \\
			\midrule
			MT~\cite{tarvainen2017mean} & 184.0 & 648.0 & 144.1 & 508.6 & 129.8 & 515.0 \\
			L2R~\cite{liu2018leveraging}  & 159.2 & 650.3 & 138.3 & 550.2 & 125.0 & 501.9 \\
		    SUA~\cite{meng2021spatial} & - & - & - & - & 111.7 & 443.2  \\
		    \ourmodel{} (Ours)  & \textbf{116.7} & \textbf{598.8}  & \textbf{88.2} & \textbf{515.9} & \textbf{76.3} & \textbf{422.8} \\
			\bottomrule[1.5pt]
		\end{tabular}
	\end{center}
\caption{Comparisons with the state of the arts semi-supervised counting methods on NWPU. The experimental settings are referred to the work~\cite{meng2021spatial}.}
\label{tab:nwpu}
\end{table*}}

The semi-supervised setting on NWPU follows the work~\cite{meng2021spatial}. We keep the validation images to evaluate our model’s performance. In the training set, $10\%$ images are randomly selected as the validation set. For the setting of labeled ratios of $5\%$, $10\%$ and $40\%$, the corresponding proportion of images in the training set will be selected as labeled data and the rest images will be regarded as unlabeled data.

We compare our method with recent state-of-the-art semi-supervised methods, including mean teacher (MT)~\cite{tarvainen2017mean}, Learning to Rank (L2R)~\cite{liu2018leveraging} and SUA~\cite{meng2021spatial}. The qualitative result is shown in Table~\ref{tab:nwpu}. It can be observed that \ourmodel{} outperforms other methods by an obvious counting accuracy improvement on all three settings of ratios of labeled data.

\subsection{The impact of PDM and ECR loss.}
We conduct experiments to study the impact of two proposed loss functions. Specifically, $\mathcal{L}_P$ represents the PDM loss without ECR, and the combination of $\mathcal{L}_P$ and $\mathcal{L}_E$ forms the proposed \ourmodel{}. The comparison result is shown in Table~\ref{tab:unlabeled}. With the help of unlabeled data and the corresponding ECR, \ourmodel{} improves the counting accuracy of `supervisions from only labeled data' over $7.8$ and $12.2$ in terms of MAE and MSE respectively. The experimental results validate that through the self-supervision of ECR from unlabeled data, the prediction capability and accuracy of the model are enhanced. The improvement is the sense of semi-supervised learning.


\def\arraystretch{1.1}
\renewcommand{\tabcolsep}{10 pt}{
\begin{table*}[t]
\footnotesize
\begin{center}
\begin{tabular}{c|ccc|ccc}
  \toprule[1pt]
  Labeled Percentage & Loss & MAE & MSE & Loss & MAE & MSE\\
  \midrule
  $5\%$ & $\mathcal{L}_P$ & 129.5 & 212.8  & $\mathcal{L}_P + \lambda \mathcal{L}_E$ & 115.3 & 195.2 \\
  $10\%$ & $\mathcal{L}_P$ & 117.4 & 211.8 & $\mathcal{L}_P + \lambda \mathcal{L}_E$ & 103.4 & 179.0 \\
  $40\%$ & $\mathcal{L}_P$ & 97.8 & 167.6 & $\mathcal{L}_P + \lambda \mathcal{L}_E$ & 90.0 & 155.4 \\
  \midrule
  $100\%$ & $\mathcal{L}_P$ & 78.5 & 135.8 & - & - & - \\
  \toprule[1pt]
\end{tabular}
\caption{The impact of ECR loss. Experiments are conducted on UCF-QNRF. With the help of ECR to exploit supervisions from unlabeled data, we get a further improvement on counting accuracy.}
\label{tab:unlabeled}
\end{center}
\end{table*}}

 \subsection{The influence of norm level in PDM loss.}
We conduct experiments to study the influence of two different norm levels in PDM loss. Specifically, when $l=1$, the loss is strictly equals to the 1-d Wasserstein loss. We observed that the results obtained by l2 (MAE:115.3 and MSE: 195.2) are far better than l1 (MAE:127.0 and MSE: 223.9). We analyze that this comes from the fact that l2 will have stronger supervision on the parts with larger gaps between distributions, which exactly benefits the optimization of the loss more.

\subsection{Comparisons to the cross-domain methods}\label{sec:domain}

\renewcommand{\tabcolsep}{8 pt}{
\begin{table*}[thbp!]
\small
	\begin{center}
		\begin{tabular}{c|cc|cc|cc}
			\toprule[1.5pt]
			\multicolumn{1}{c|}{\multirow{2}*{Labeled Ratio}} &  \multicolumn{2}{|c|}{$5\%$} &  \multicolumn{2}{|c|}{$10\%$} &
			\multicolumn{2}{|c}{$40\%$}  \\
			
			\multicolumn{1}{c|}{} &  \multicolumn{1}{|c}{MAE} &
			\multicolumn{1}{c|}{MSE} & \multicolumn{1}{|c}{MAE} &
			\multicolumn{1}{c|}{MSE} &  \multicolumn{1}{|c}{MAE} &
			\multicolumn{1}{c}{MSE} \\
			\midrule
			SEFD~\cite{han2020focus} & 134.2 & 178.7 & 110.4 & 147.9 & 93.5 & 128.2 \\
            SSCD~\cite{liu2022leveraging} & 96.7 & 141.3 & 86.4 & 126.4 & 79.0 & 113.8 \\
			
		    \ourmodel{} (Ours)  & \textbf{85.5} & \textbf{131.0}  & \textbf{72.1} & \textbf{116.4} & \textbf{63.0} & \textbf{100.9} \\
			\bottomrule[1.5pt]
		\end{tabular}
	\end{center}
\caption{Comparisons with the cross-domain counting methods on SHA. \ourmodel{} consistently outperforms cross-domain methods by a substantial margin across all labeled ratios.}
\label{tab:domain}
\end{table*}}

\HUI{To further compare and validate the superiority of our approach, we reproduce two recent state-of-the-art cross-domain crowd counting methods, \textit{i.e.}, SSCD~\cite{liu2022leveraging} and SEFD~\cite{han2020focus} and test their performance on semi-supervised crowd counting datasets. Specifically, SEFD provides the open-source code\footnote{https://github.com/weizheliu/Cross-Domain-Crowd-Counting} for the model, and we directly employ it in semi-supervised crowd counting tasks. For SSCD, we reproduced the model based on the implementation details~\cite{liu2022leveraging}. Additionally, we utilized a detection model trained on the CrowdHuman dataset~\cite{shao2018crowdhuman} to generate semantic labels, thereby completing the auxiliary task for SSCD, which is consistent with the details mentioned in the paper. We test their performance on SHA dataset and the results are presented in Table~\ref{tab:domain}. It is shown that our method consistently outperforms cross-domain methods by a significant margin across all labeled ratios.}

\HUI{We attribute this superiority to the fact that cross-domain methods assume a substantial number of labeled source domain data compared to a small number of unlabeled target domain data, whereas in the semi-supervised setting, the available labeled data is considerably less than the unlabeled data. Our approach employs the Pixel-by-pixel Density Distribution Modeling method, which effectively addresses noise and uncertainty within the limited labeled data. Consequently, \ourmodel{} maximizes the utility of labeled data, resulting in a substantial improvement of counting accuracy in semi-supervised tasks.}

\subsection{The impact of Pixel-wise Distribution Matching loss} 
We study the proposed PDM loss by comparing it with the Cross Entropy (CE) loss and MSE loss, {and more noteworthy}, the counting loss including the Bayesian loss~\cite{ma2019bayesian} and DM loss~\cite{wang2020distribution} which achieve best results in the fully-supervised domain. The experimental result is shown in Table~\ref{tab:dm_loss}, which is held on UCF-QNRF dataset with a labeled ratio of $5\%$. Our loss outperforms all four losses by large margins. The result suggests that the {awareness of the semantic information is helpful in matching the distribution between prediction and ground truth.}
\XP{Moreover and surprisingly}, the CE loss and MSE loss, which are more specialized for classification \XP{originally, surpass} the \XP{counting} loss \XP{in this case. The reason probably lies in} that when only a small number of ground-truth labels is available, regarding the single-value density as a probability distribution provides a better way for improving the robustness and accuracy of the counting model.

\def\arraystretch{1.1}
\begin{table}[t] %
\centering
\renewcommand{\tabcolsep}{12 pt}{
\footnotesize
\begin{tabular}{c|c|c|c}
  \toprule[1pt]
   & CE & MSE & PDM \\
  \midrule
  MAE & 125.4 & 132.8 & \textbf{115.3} \\
  MSE & 211.6 & 223.2 & \textbf{195.2} \\
  \midrule
  & BL & DM & PDM \\
  \midrule
  MAE & 136.5 & 133.4 & \textbf{115.3} \\
  MSE & 234.7 & 225.3 & \textbf{195.2} \\
  \toprule[1pt]
\end{tabular}
\caption{Comparisons of using different losses to get supervisions from ground-truth. Experiments are held on UCF-QNRF with $5\%$ labels. } \label{tab:dm_loss}}
\end{table}

\def\arraystretch{1.1}
\begin{table*}[thbp!]
\centering
\renewcommand{\tabcolsep}{12 pt}{
\footnotesize
\centering
\begin{tabular}{cc|cc|cc}
  \toprule[1pt]
  Each Branch & Dual Branch & 5\% MAE & 5\% MSE & 100\% MAE & 100\% MSE \\
  \midrule
  Maximum & Average & 134.5 & 240.6 & 85.8 & 142.7 \\
  \midrule
  Expectation & Average & 130.0 & 215.9 & 79.3 & 137.5 \\
  \midrule
  Expectation & Confidence & 129.5 & 212.8 & 78.5 & 135.8 \\
  \toprule[1pt]
\end{tabular}
\caption{The influence of probabilistic distribution modelling and the use of $\omega$.}\label{tab:direct}}
\end{table*}

\subsection{The influence of probabilistic distribution modelling}

Table~\ref{tab:direct} reports the influence of modelling each pixel by probability distribution. For clear comparison, we set two baselines. First, instead of taking the expectation in probability distribution, we try to directly select the category with the maximum predicted score for each pixel in each branch and converts it to the corresponding predefined representation value. Second, instead of adopting $\omega$ to consider the different confidences between the dual branches, we make a simple average without using Eq.~\ref{eq:confidence}. We held experiments on UCF-QNRF under two settings: (1) only 5\% labeled data without any unlabeled data; (2) 100\% labeled data. This is because taking the maximum for each branch contradicts ECR, which supervised the model by expectation. For a fair comparison, we only use PDM to study their difference on labeled data. Using a soft expectation value instead of a hard maximum value for each branch results in remarkable growth in counting accuracy. Meanwhile, considering the prediction confidence when fusing between dual branches will further help improve the accuracy. \ourmodel{} is a combination of using expectation and prediction confidence.

\subsection{The Effectiveness of the Interval Partition Strategy}\label{sec:partition}

\HUI{To ensure the effectiveness of our interval partitions, we have adopted the partition strategy proposed in the paper~\cite{wang2021uniformity} to obtain the appropriate intervals. The Uniform Error Partition (UEP) follows the principle of maximum entropy and has been shown to be effective in mitigating misclassification errors and addressing sample imbalance issues.}

\HUI{In order to validate whether this partition strategy remains effective when integrated with our approach, we conducted an ablation experiment. Specifically, we compared it against two baseline strategies: UniformLen and UniformNum. The first strategy involves dividing intervals with equal interval lengths while the second strategy divides intervals to ensure an equal number of samples in each interval. The experiments are held on the SHA dataset, using $5\%$ labeled data. The results are summarized in the Table~\ref{tab:partition}.}

\renewcommand{\tabcolsep}{10 pt}{
\begin{table}[thbp!]
\small
\begin{center}
\begin{tabular}{c|c|c}
  \toprule[1pt]
  Strategy & MAE & MSE \\
  \midrule
  + UniformLen & 87.1 & 155.9 \\
  + UniformNum & 86.8 & 154.1 \\
  + UEP & 85.5 & 131.0 \\
  \toprule[1pt]
\end{tabular}
\caption{The influence of the interval partition strategy.} 
\label{tab:partition}
\end{center}
\end{table}}

\HUI{Comparison of the results reveals that when integrated with the UEP strategy, our method achieved a MAE of 85.5. This marks an improvement of 1.6 and 1.3 over the two baseline strategies, respectively, highlighting the efficacy of incorporating the partition strategy in \cite{wang2021uniformity} to enhance our approach's performance.}

\HUI{Furthermore, it's worth noting that the counting accuracy variations produced by different strategies are relatively minor. This discovery underscores the robustness of our proposed method across various partition strategies, confirming its effectiveness in semi-supervised counting tasks.}

\subsection{The Influence of Parameter $\lambda$}\label{sec:lambda}

$\lambda$ controls the proportion of gradient contribution of the unlabeled loss. As it gets larger, the self-supervised signals generated by unlabeled images will have greater effects on the model update. We hold experiments to study the influence of $\lambda$, which result is shown in Table~\ref{tab:lambda}.

The experiments are conducted on UCF-QNRF with a labeled ratio of $5\%$. As the result is shown, when $\lambda$ is in the appropriate range, \emph{i.e.} from 0.005 to 0.05, the gap of final counting performance is small. However, when the value is not suitable, the unlabeled images will impose too much or conversely, too little supervision on the gradient update of the model. Then the accuracy drops sharply, which is similar to excluding unlabeled data. Therefore, we set the unlabeled parameter $\lambda=0.01$.

\renewcommand{\tabcolsep}{10 pt}{
\begin{table*}[thbp!]
\small
\begin{center}
\begin{tabular}{c|c|c|c|c|c|c|c}
  \toprule[1pt]
  $\lambda$ & 0 & 0.0005 & 0.001 & 0.005 & 0.01 & 0.05 & 0.1 \\
  \midrule
  MAE & 129.5 & 127.8 & 123.4 & 118.6 & 115.3 & 116.8 & 122.5 \\
  MSE & 212.8 & 209.7 & 207.9  & 199.2 & 195.2 & 195.8 & 206.0 \\
  \toprule[1pt]
\end{tabular}
\caption{The influence of parameter $\lambda$ on UCF-QNRF (labeled ratio $5\%$).} 

\label{tab:lambda}
\end{center}
\end{table*}}

\subsection{The Performance against the Adverse Weather Conditions}\label{sec:weather}

\HUI{To assess the robustness of our method, we conducted experiments to evaluate its performance in the presence of adverse weather conditions in crowd images. Specifically, we utilized crowd images from the JHU-Crowd++ dataset~\cite{sindagi2020jhu}, which include weather labels and exhibit common weather-related degradations such as haze, snow, rain, etc. The results of these experiments are presented in  Table~\ref{tab:weather}.}

\HUI{We compared the performance of our \ourmodel{} with that of the recent DACount method~\cite{lin2022semi} method. The results demonstrate that across three different labeled ratios, our method consistently outperforms DACount. This clearly illustrates that our approach is capable of mitigating the negative influence caused by adverse weather conditions, consistently delivering strong performance and maintaining stability.}

\renewcommand{\tabcolsep}{8 pt}{
\begin{table*}[thbp!]
\small
	\begin{center}
		\begin{tabular}{c|cc|cc|cc}
			\toprule[1.5pt]
			\multicolumn{1}{c|}{\multirow{2}*{Labeled Ratio}} &  \multicolumn{2}{|c|}{$5\%$} &  \multicolumn{2}{|c|}{$10\%$} &
			\multicolumn{2}{|c}{$40\%$}  \\
			
			\multicolumn{1}{c|}{} &  \multicolumn{1}{|c}{MAE} &
			\multicolumn{1}{c|}{MSE} & \multicolumn{1}{|c}{MAE} &
			\multicolumn{1}{c|}{MSE} &  \multicolumn{1}{|c}{MAE} &
			\multicolumn{1}{c}{MSE} \\
			\midrule
			DACount~\cite{lin2022semi} & 158.6 & 712.1 & 141.2 & 665.9 & 113.4 & 596.7 \\
			\ourmodel{}  & 155.0 & 708.2 & 139.9 & 664.1 & 108.7 & 582.3 \\
			\bottomrule[1.5pt]
		\end{tabular}
	\end{center}
\caption{Performance in the presence of adverse weather conditions in crowd images. The experiments are held on the JHU dataset.}
\label{tab:weather}
\end{table*}}

\subsection{The Influence of Different Crowd Distributions in Labeled and Unlabeled Images}\label{sec:distributions}

\HUI{In this section, we conduct experiments to assess how crowd distributions in labeled and unlabeled images influence the performance of our method. Our investigation includes comparing results from balanced crowd distributions to two extreme scenarios: sparse unlabeled images and crowded unlabeled images. We sort the images in the training dataset based on the ground truth crowd counts, arranging them from images with the fewest number of people (sparse) to those with the highest number of people (crowded). Subsequently, we simulated the two extreme scenarios. First, we selected the top $5\%$ of images with the highest crowd counts as labeled data, while the remaining images with fewer number of people were used as unlabeled data, and the scenario is named sparse unlabeled scenario. Conversely, we selected the bottom $5\%$ of images with the lowest crowd counts as labeled data, while the majority of images with higher crowd densities were used as unlabeled data, which is named crowded unlabeled scenario. The experiments are held on the SHA dataset and the result is shown in Table~\ref{tab:distributions}.}

\renewcommand{\tabcolsep}{10 pt}{
\begin{table}[thbp!]
\small
\begin{center}
\begin{tabular}{c|c|c|c}
  \toprule[1pt]
  Methods & Distribution & MAE & MSE \\
  \midrule
  \ourmodel{} & Balanced & 85.5 & 131.0 \\
  \ourmodel{} & Sparse Unlabeled & 97.1 & 162.6 \\
  \ourmodel{} & Crowded Unlabeled & 128.5 & 235.5 \\
  \toprule[1pt]
\end{tabular}
\caption{The influence of different crowd distributions in labeled and unlabeled images. Experiments are held on SHA ($5\%$ labeled ratio).} 
\label{tab:distributions}
\end{center}
\end{table}}

\HUI{In the sparse unlabeled scenario, the MAE increased by 11.6, rising from 85.5 to 97.1. However, our approach of sparse unlabeled scenario still outperformed traditional models such as MT~\cite{tarvainen2017mean} (104.7) and L2R~\cite{liu2018leveraging} (103.0). In contrast, under the crowded unlabeled scenario, the MAE increased by 43, indicating a substantial decline in accuracy. This decline can be attributed to the model's lack of exposure to crowded labels during training. Additionally, the accuracy in counting dense crowds plays a pivotal role in the overall precision, contributing to the model's diminished performance.}

\HUI{These results underscore that different distributions in labeled and unlabeled data can have significant impacts on semi-supervised counting models. In our standard semi-supervised experiments, we employ a random data partition approach, which includes a balanced selection of sparse and crowded data. It proves to be a robust choice, ensuring the model is well-prepared to handle a wide range of crowd density scenarios typically encountered in real-world applications.}

\subsection{The Effect of Varying the Threshold in ECR Loss}\label{sec:xi}

The threshold $\xi$ controls the consistency regularization of the unlabeled loss. When $\xi=0$, all pixels in the image will be supervised to be consistent between the dual branches. As it gets larger, it will only effect on pixels with higher prediction confidences. We hold experiments to study the influence of $\xi$, which result is shown in Table~\ref{tab:xi}.

The experiments are conducted on UCF-QNRF with a labeled ratio of $5\%$. As the result is shown, the counting accuracy will vary with the change of the ECR threshold, and the appropriate threshold will significantly help the counting accuracy.

\renewcommand{\tabcolsep}{20 pt}{
\begin{table}[thbp!]
\small
\begin{center}
\begin{tabular}{c|c|c}
  \toprule[1pt]
  $\xi$ & MAE & MSE \\
  \midrule
  0	& 119.8	& 199.4 \\
  0.5 & 115.3 & 195.2 \\
  0.75 & 117.7 & 197.4 \\
  \toprule[1pt]
\end{tabular}
\caption{The influence of parameter $\xi$ on UCF-QNRF (labeled ratio $5\%$).} 

\label{tab:xi}
\end{center}
\end{table}}

\subsection{Performance under Fully-Supervised Setting}\label{sec:fully}
We hold experiments under the fully-supervised setting to study the effectiveness of proposed density tokens and Pixel-wise probabilistic Distribution (PDM) loss. \XP{It is worth mentioning that in the fully-supervised setting}, our model does not need the self-supervised inter-branch Expectation Consistency Regularization (ECR), \XP{which are design for learning from unlabeled data.}

We compare our method with state-of-the-art fully-supervised methods on five datasets named UCF-QNRF, JHU-Crowd++, ShanghaiTech A, ShanghaiTech B and NWPU. \XP{The results are summarized in Table~\ref{tab:performance}. Though it is not particularly designed for fully-supervised crowd counting, our method performs fairly well. For instance, on the UCF-QNRF dataset, it performs the best in terms of MAE and the second in terms of MSE, when compared with the most commonly used and best accepted methods like P2PNet~\cite{song2021rethinking}, UEPNet~\cite{wang2021uniformity}, DM-Count~\cite{wang2020distribution}, BL~\cite{ma2019bayesian}, etc.} This consistent performance boost shows the effectiveness of the proposed architecture and the PDM loss. \XP{When considering the superior performance achieved by our method in semi-supervised crowd counting reported in the main test, it further implies that optimal semi-supervised counting is built on both the ability to learn from labeled data and unlabeled data.}

\def\arraystretch{1.1}
\renewcommand{\tabcolsep}{10 pt}{
\begin{table*}[thbp!]
\small
	\begin{center}
		\begin{tabular}{c|cc|cc|cc|cc|cc}
			\toprule[1.5pt]
			\multicolumn{1}{c}{Dataset} &  \multicolumn{2}{c}{UCF-QNRF} &  \multicolumn{2}{c}{JHU++} & \multicolumn{2}{c}{ShanghaiTech A} &
			\multicolumn{2}{c}{ShanghaiTech B} & \multicolumn{2}{c}{NWPU} \\
			
			\multicolumn{1}{c}{Method} & \multicolumn{1}{c}{MAE} &
			\multicolumn{1}{c}{MSE} & \multicolumn{1}{c}{MAE} &
			\multicolumn{1}{c}{MSE} &  \multicolumn{1}{c}{MAE} &
			\multicolumn{1}{c}{MSE} &  \multicolumn{1}{c}{MAE} &
			\multicolumn{1}{c}{MSE} & \multicolumn{1}{c}{MAE} &
			\multicolumn{1}{c}{MSE}\\
			
			\hline
			\hline
			MCNN~\cite{zhang2016single} & 277 & 426 & 188.9 & 483.4 & 110.2 & 173.2 & 26.4 & 41.3 & 232.5 & 714.6 \\
			CSRNet~\cite{li2018csrnet} & - & - & 85.9 & 309.2 & 68.2 & 115.0 & 10.6 & 16.0 & 121.3 & 387.8 \\
			SANet~\cite{cao2018scale} & - & - & 91.1 & 320.4 & 67.0 & 104.5 & 8.4 & 13.6 & 190.6 & 491.4 \\
			S-DCNet~\cite{xiong2019open} & 104.4 & 176.1 & - & - & 58.3 & 95.0 & 6.7 & 10.7 & - & - \\
			BL~\cite{ma2019bayesian} & 88.7 & 154.8 & 75.0 & 299.9 & 62.8 & 101.8 & 7.7 & 12.7 & 105.4 & 454.2 \\
		    DM-Count~\cite{wang2020distribution} & 85.6 & 148.3 & - & - & 59.7 & 95.7 & 7.4 & 11.8 & 88.4 & 388.6 \\
		    UOT~\cite{ma2021learning} & 83.3 & 142.3 & 60.5 & 252.7 & 58.1 & 95.9 & 6.5 & 10.2 & 87.8 & 387.5\\
			S3~\cite{lin2021direct} & \uline{80.6} & 139.8 & \uline{59.4} & \uline{244.0} & 57.0 & 96.0 & 6.3 & 10.6 & 83.5 & \uline{346.9} \\
			P2PNet~\cite{song2021rethinking} & 85.3 & 154.5 & - & - & \textbf{52.7} & \textbf{85.1} & \uline{6.3} & \textbf{9.9} & \uline{77.4} & 362.0 \\
			UEPNet~\cite{wang2021uniformity} & 81.1 & \textbf{131.7} & - & - & \uline{54.6} & 91.2 & 6.4 & 10.9 & - & - \\
			\textbf{\XP{Our method}} & \textbf{78.5} & \uline{134.2} & \textbf{55.8} & \textbf{237.6} & 56.6 & \uline{89.9} & \textbf{6.2} & \uline{10.2} & \textbf{74.3} & \textbf{327.3} \\
			\bottomrule[1.5pt]
		\end{tabular}
	\end{center}
	
\caption{Comparisons with the state of the arts on UCF-QNRF, JHU-Crowd++, ShanghaiTech A, ShanghaiTech B and NWPU under fully-supervised setting. The best performance is shown in \textbf{bold} and the second best is shown in \uline{underlined}. Our model is very competitive with SOTA supervised methods.}
\label{tab:performance}
\end{table*}}

\subsection{Running Cost Evaluation}\label{sec:running}
We evaluate the running cost of our method, which comparison result is reported in Table~\ref{tab:cost}. The result of floating-point operations (FLOPs) is computed on one $384 \times 384$ input image and the result of inference time is tested on $1024 \times 1024$ images. We compare \ourmodel{} with BL~\cite{ma2019bayesian} model which serves as a basic counting network, and VGG19+Trans where Trans stands for the vanilla transformer encoder with self-attention.

The growth of model parameters comes from the dual-branch decoder structure. However, without the need for a regression head which is always composed of a convolutional network used in BL and VGG19+Trans, the FLOPs of \ourmodel{} achieve the minimum. Meanwhile, with respect to feature length $N$ and the token number $C << N$, the self attention in encoder takes $\mathcal{O}(N^2)$ time and space while the cross attention in decoder takes $\mathcal{O}(C^2+CN) \approx \mathcal{O}(CN)$. From the results compared with VGG19+Trans, \ourmodel{} uses fewer FLOPs and inference time, justifying the low computational cost of the decoder.

\def\arraystretch{1.1}
\renewcommand{\tabcolsep}{12 pt}{
\begin{table*}[thbp!]
\small
\begin{center}
\begin{tabular}{c|cccc}
\toprule[1pt]
 & BL~\cite{ma2019bayesian} & VGG19+Trans & \textbf{\ourmodel{}}\\
\midrule
Model Size (M) & 21.5 & 29.9 & 36.8\\
FLOPs (G) & 60.8 & 65.6 & 57.8\\
Inference Time (s / 100 images) & 9.8 & 11.4 & 10.2\\
\toprule[1pt]
\end{tabular}
\end{center}
\caption{Comparison of the Model Size, FLOPs and Inference Time. Trans stands for the vanilla encoder. The growth of model parameters comes from the dual-branch decoder structure. But without the need for a regression head which is always composed of a convolutional network, \XP{the number of floating-point operations executed (FLOPs) of \ourmodel{} is the minimum}, and even \XP{lower than} the \XP{the non-transformer-based} BL~\cite{ma2019bayesian}.}
\label{tab:cost}
\end{table*}}

\subsection{Random Initialization}\label{sec:initialization}

'Randomly initialized query' means that when each image is input to the model, the initial value of the query will be defined as 0 or a random number. And the position encodings for a query may be added. DETR~\cite{carion2020end} adopts this solution in order to avoid overfitting a query to a specific object in detection. Conversely, we want to preserve the semantics of the density categories. We hold the ablation experiment to study the 'randomly initialized query'. However, under the semi-supervised setting, it is not only difficult to converge, but also leads to a clear drop in accuracy. The MAE/MSE under the setting of 5\% on UCF-QNRF increased to 137.2/245.7 from 115.3/195.2 respectively. We analyze this phenomenon and find that the counting task relies heavily on the density represented by each query. We will add it to the supplementary material.

\subsection{Visualizations}\label{sec:visualization}
We provide visual comparisons between our model and the previous state-of-the-art semi-supervised counting model SUA~\cite{meng2021spatial} in Figure~\ref{fig:comp_unlabel}. We visualize the predicted densities on unlabeled training images of ShanghaiTech A of both models. The first row presents input images. The second row presents predicted density maps by SUA model while the third row presents predicted density maps by our \ourmodel{}. For SUA model in unlabeled data, there will be serious false alarms in the background. In contrast, our density token guided model can perform more stable and thus produce density maps with better accuracy. We also provide visualizations of attention maps in the appendix.

\renewcommand{\tabcolsep}{8 pt}{
\begin{figure*}[t!]
	\begin{center}
		\begin{tabular}{cccc}
			
			\includegraphics[height=0.14\linewidth]{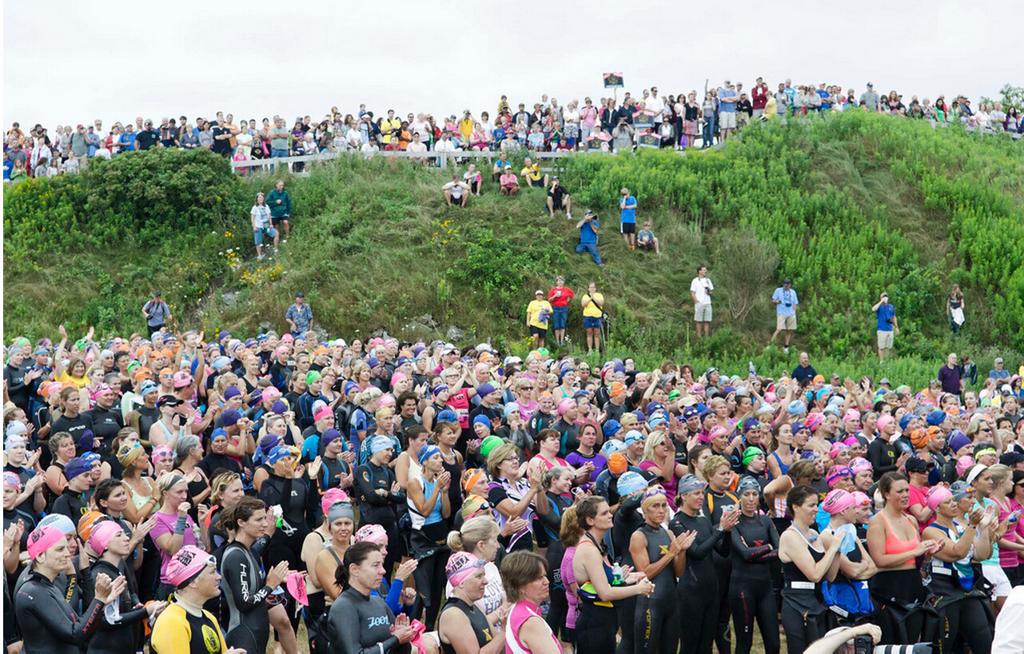}  &
			\includegraphics[height=0.14\linewidth]{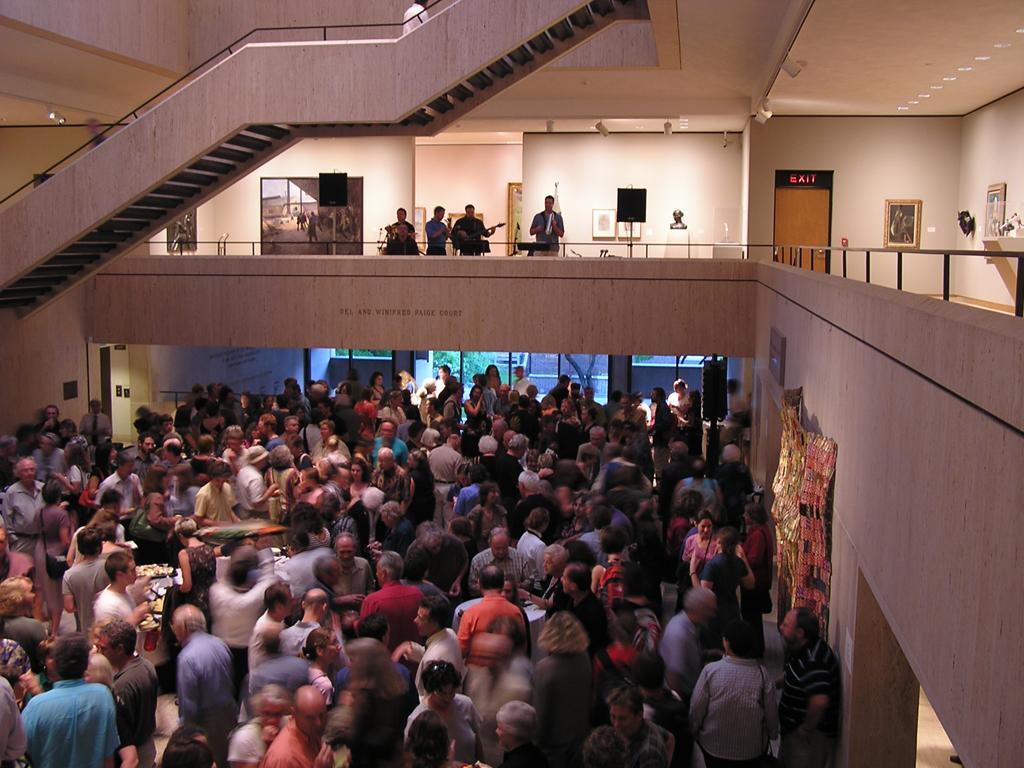} &
			\includegraphics[height=0.14\linewidth]{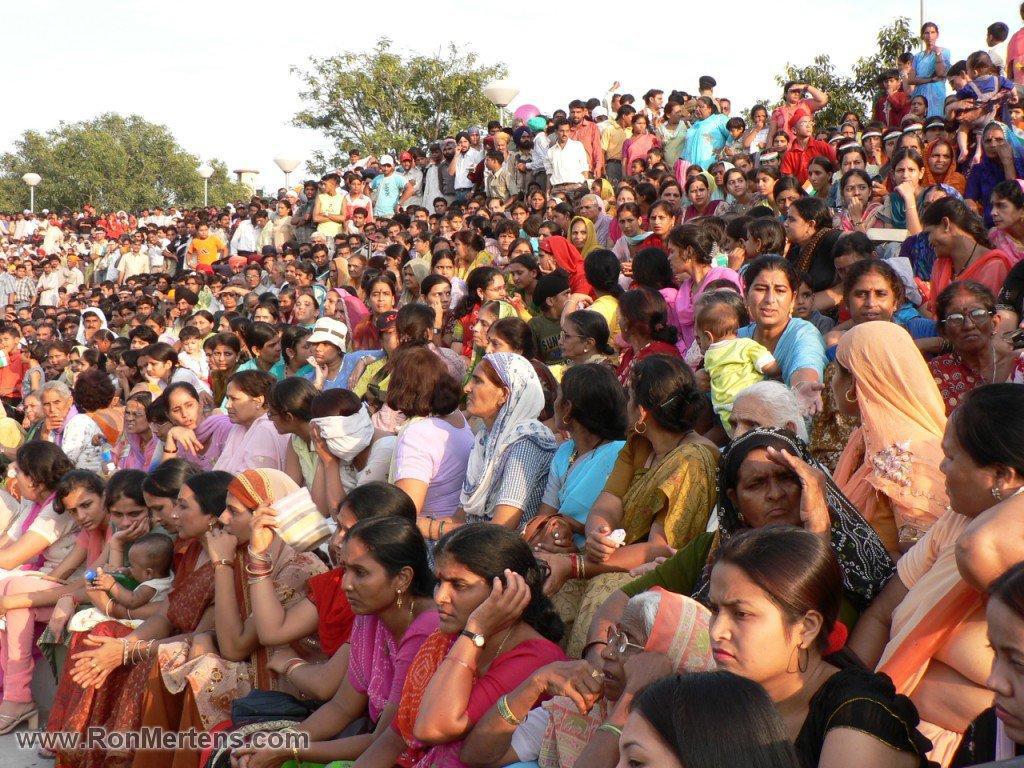}  &
			\includegraphics[height=0.14\linewidth]{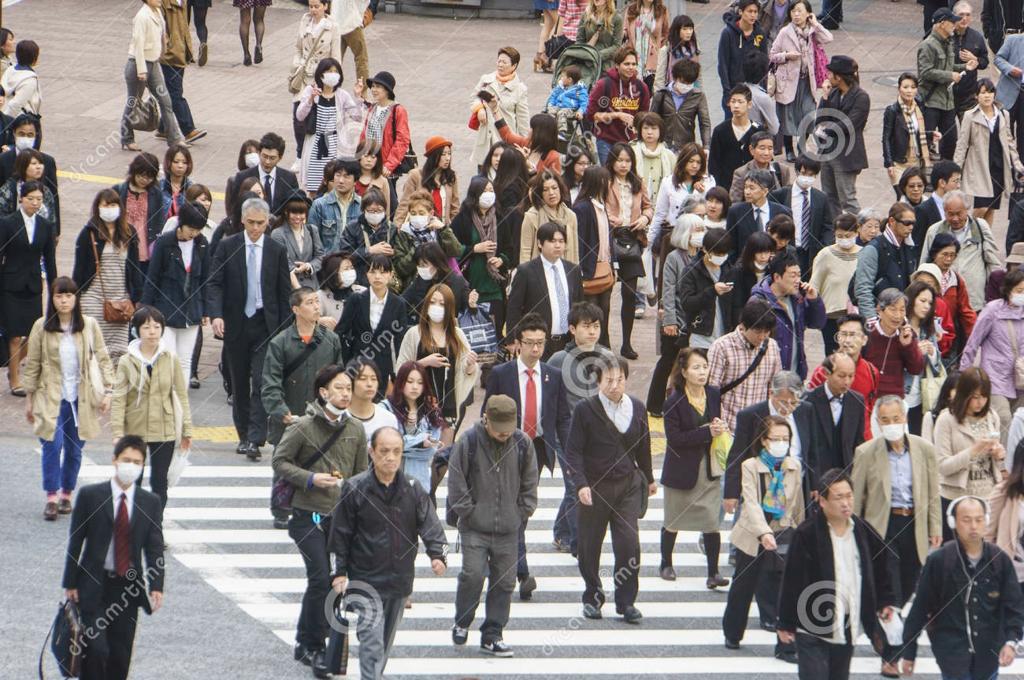} \\
			
			\footnotesize{GT: 371} & \footnotesize{GT: 184} & \footnotesize{GT: 357} & \footnotesize{GT: 102} \\
			
			\includegraphics[height=0.14\linewidth]{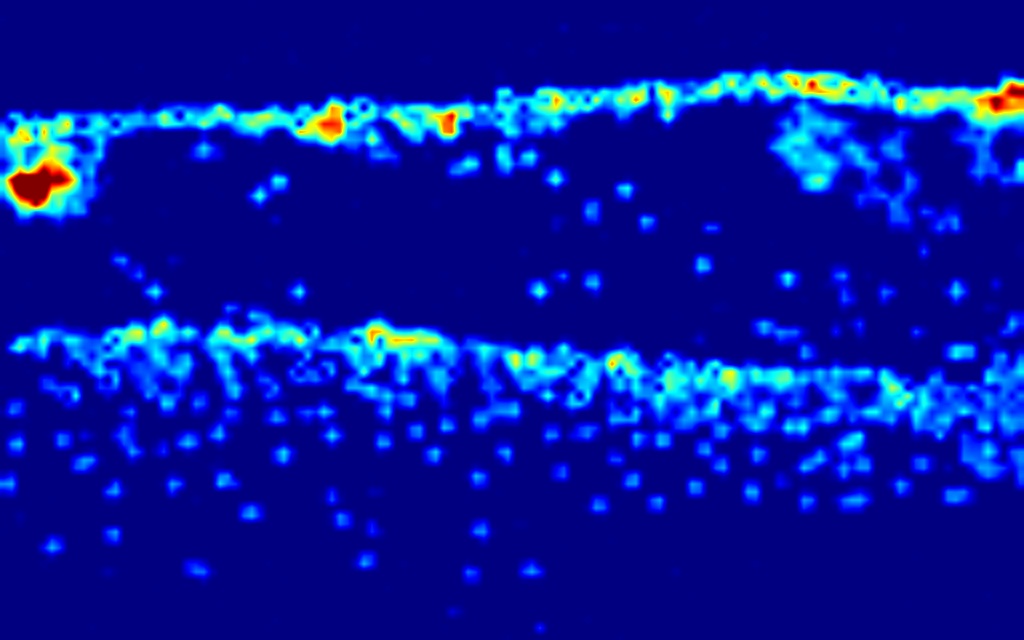}  &
			\includegraphics[height=0.14\linewidth]{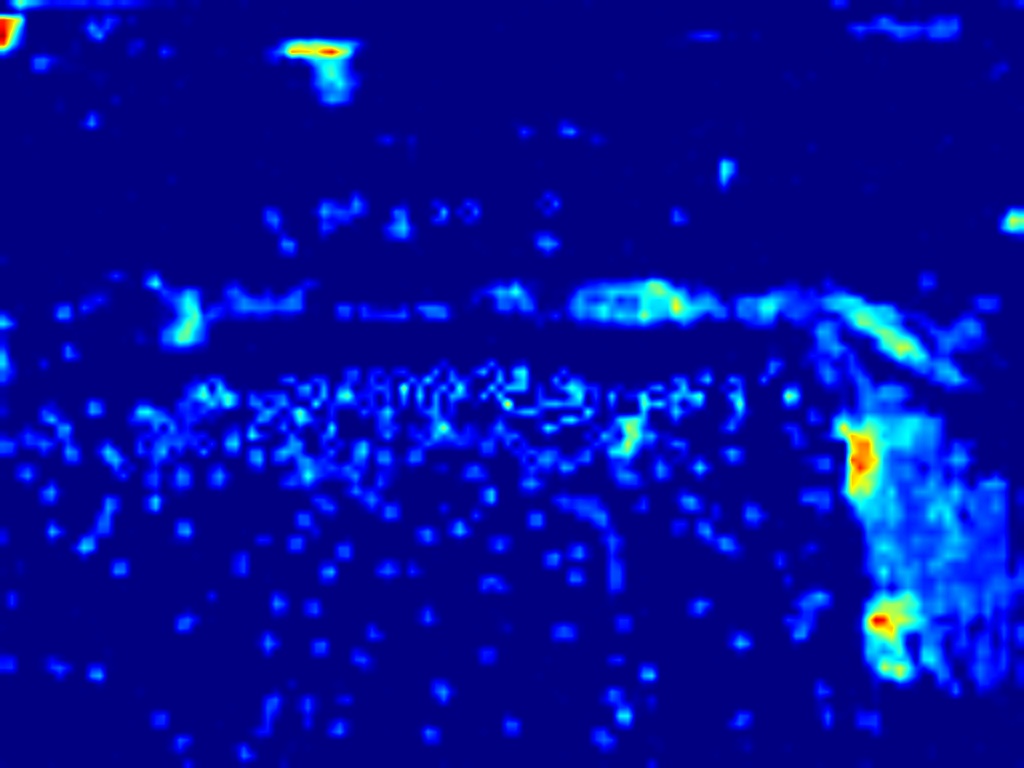} &
			\includegraphics[height=0.14\linewidth]{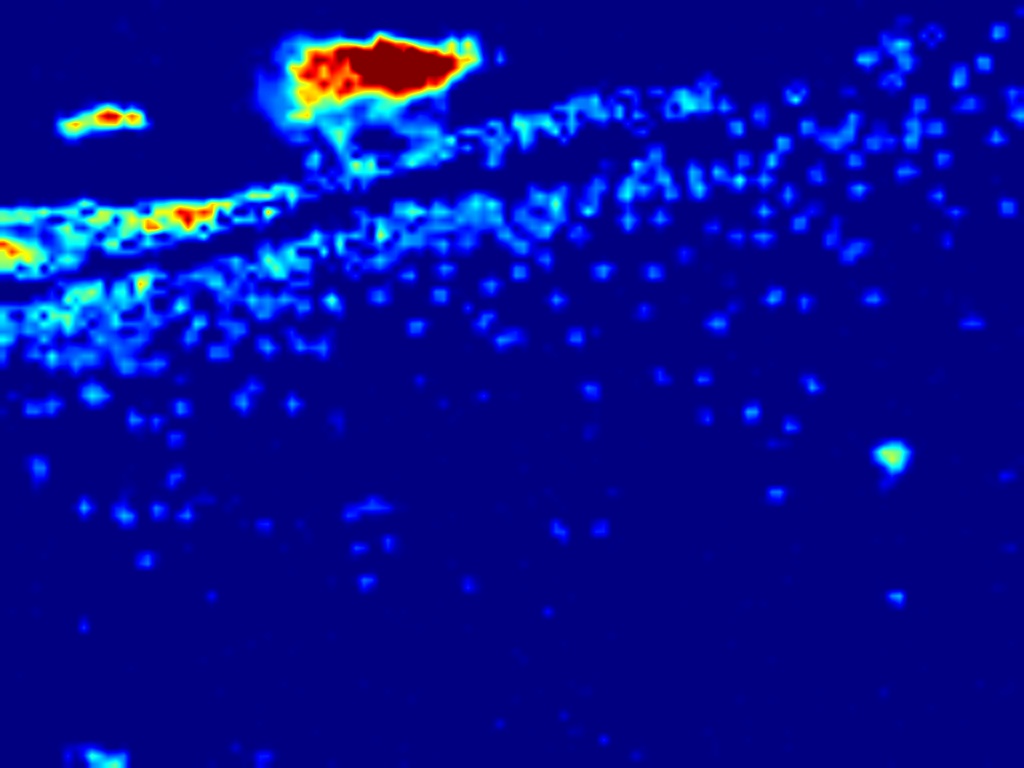}  &
			\includegraphics[height=0.14\linewidth]{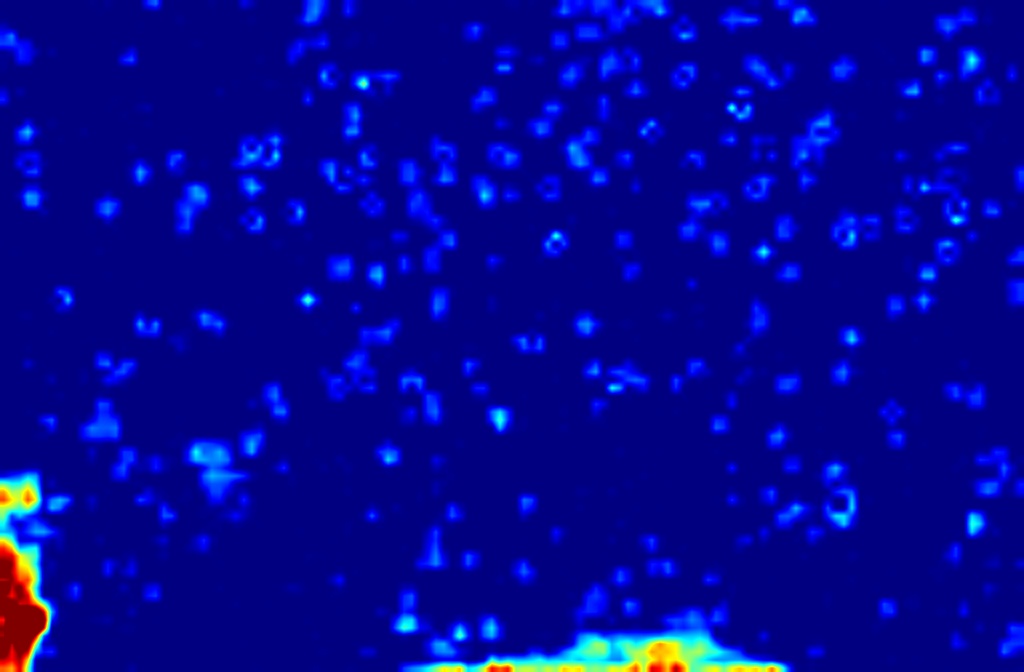} \\
			
			\footnotesize{SUA: 650} & \footnotesize{SUA: 548} & \footnotesize{SUA: 574} & \footnotesize{SUA: 392} \\
			\includegraphics[height=0.14\linewidth]{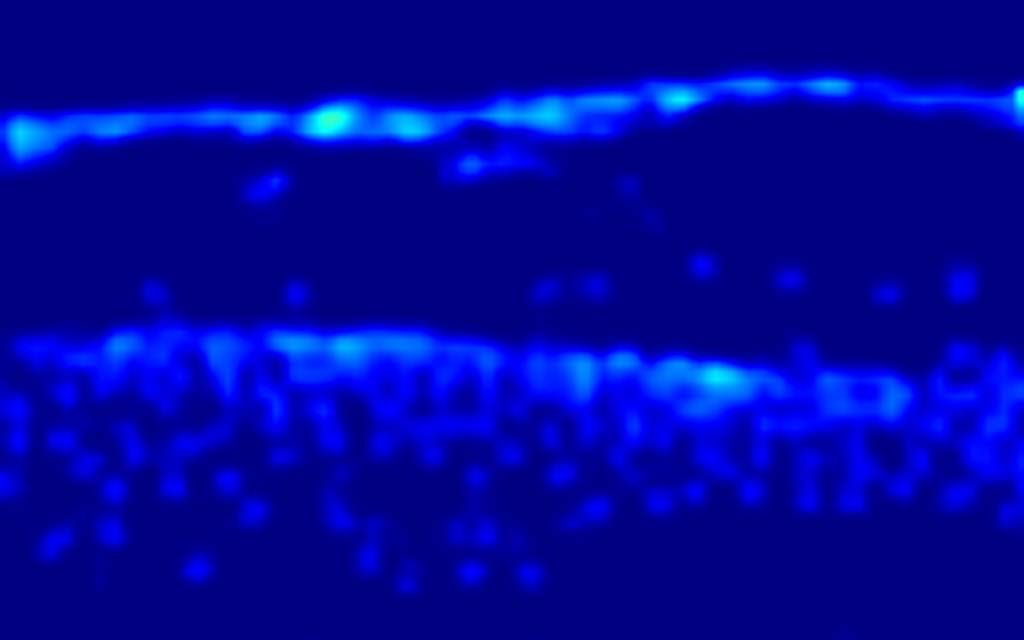}  &
			\includegraphics[height=0.14\linewidth]{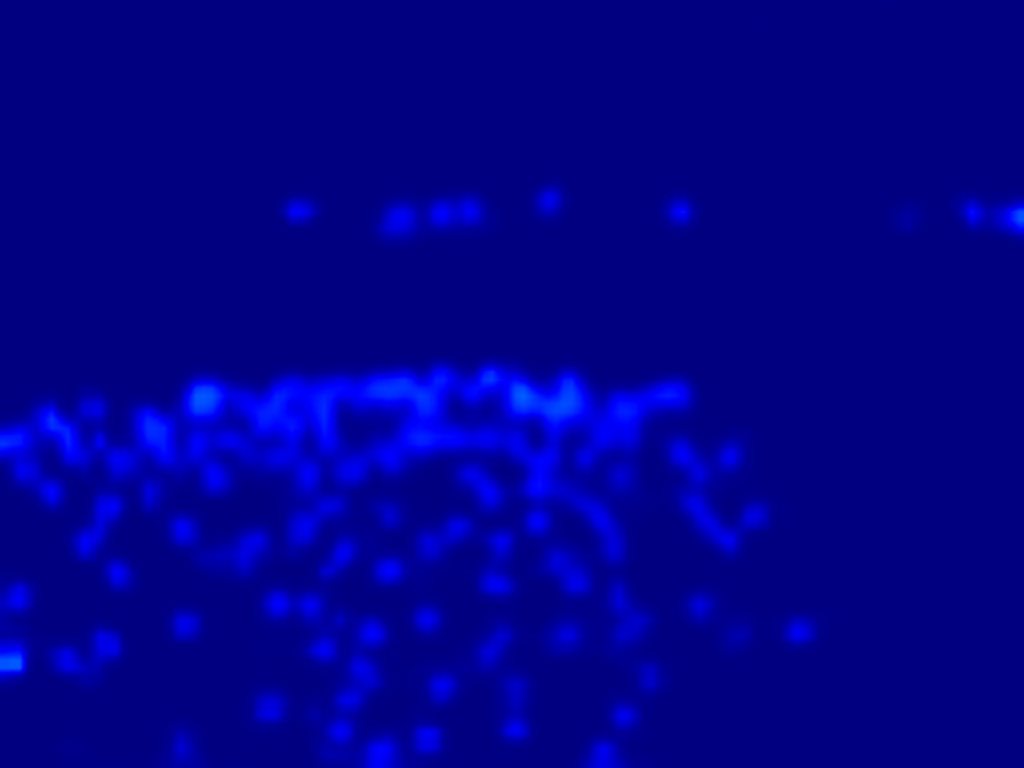} &
			\includegraphics[height=0.14\linewidth]{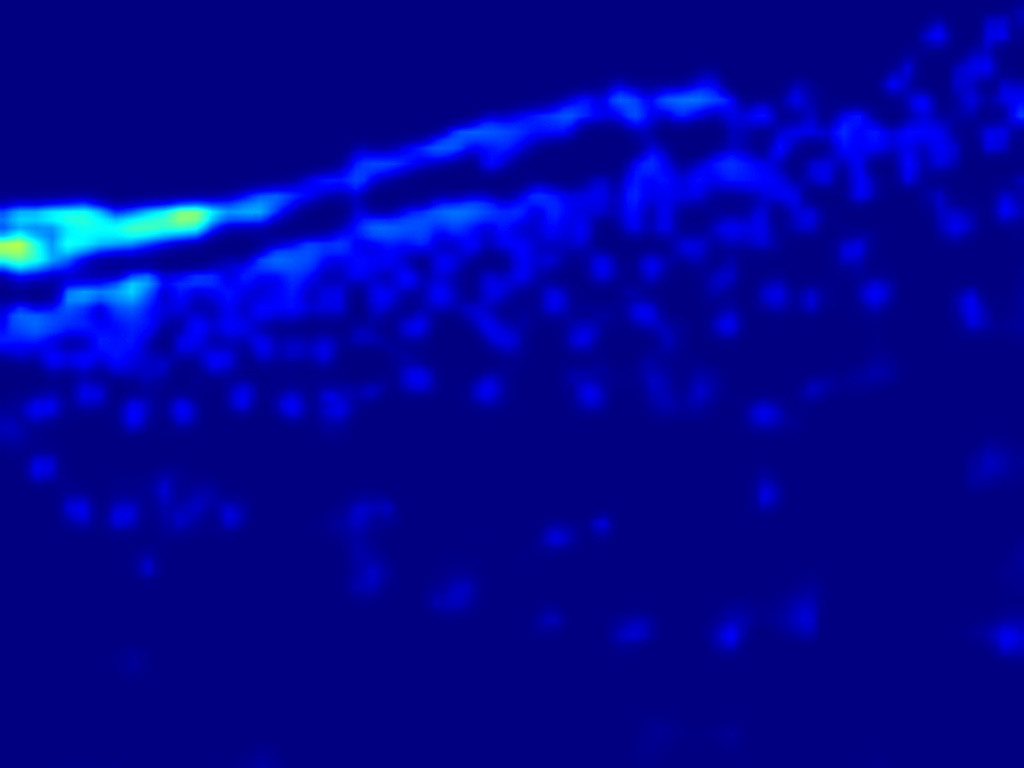}  &
			\includegraphics[height=0.14\linewidth]{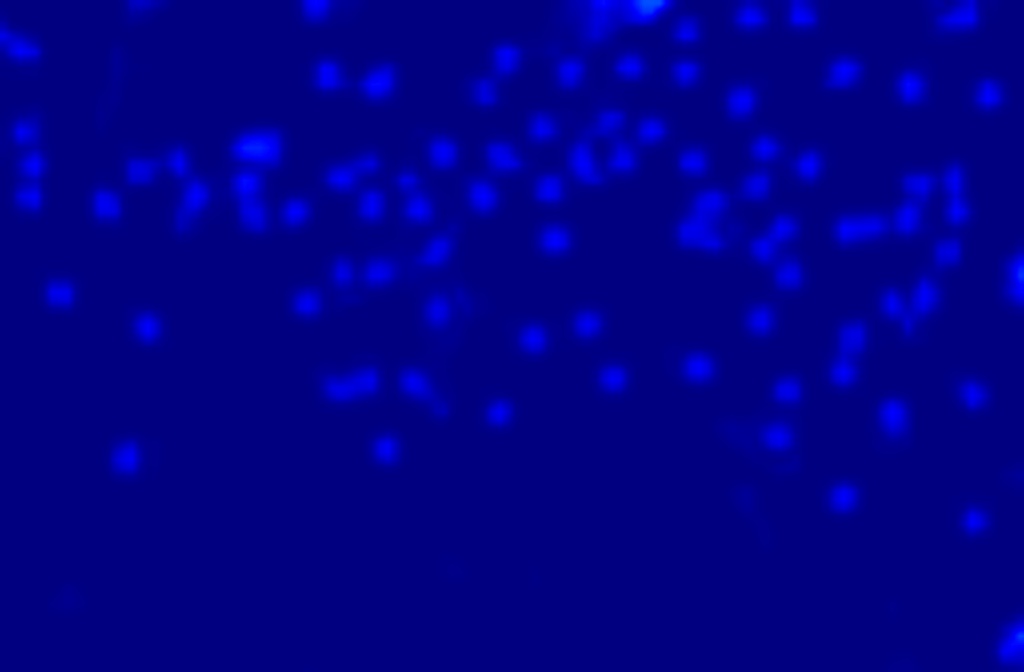} \\
			
			\footnotesize{\ourmodel{}: 379} & \footnotesize{\ourmodel{}: 175} & \footnotesize{\ourmodel{}: 331} & \footnotesize{\ourmodel{}: 121} \\

		\end{tabular}
		\caption{Visualizations of predicted densities on unlabeled training images of ShanghaiTech A. The first row: input images. The second row: predicted density maps by SUA model. The third {row}: predicted density maps by our \ourmodel{}. For SUA model in unlabeled data, serious false alarms in the background are observed, as shown in the second row. In contrast, our density token guided model can perform more stability and thus produce density maps with better accuracy in the third row.}
		\label{fig:comp_unlabel}
	\end{center}
\end{figure*}}

\subsection{The setting of density interval partitions and label generation}\label{sec:setting}

The interval partitions are preset and remain constant during training and inference stages. We follow the paper~\cite{wang2021uniformity} to obtain the appropriate intervals. 

The partitions for the first branch are

[0, 0.0019, 0.0081, 0.0165, 0.0272, 0.0404, 0.056, 0.076, 0.099, 0.126, 0.159, 0.199, 0.246, 0.303, 0.371, 0.454, 0.556, 0.684, 0.848, 1.06, 1.36, 1.8, 2.5, 3.9, 8.2].

The partitions for the second branch are

[0, 0.00087, 0.0046, 0.0119, 0.0214, 0.0333, 0.048, 0.065, 0.086, 0.112, 0.142, 0.178, 0.221, 0.272, 0.334, 0.409, 0.501, 0.615, 0.759, 0.945, 1.197, 1.55, 2.1, 3.0, 4.5, 8.5].

As the original ground truth for the counting task is in the form of discrete points. To generate labels for PDM loss, we first take the most popular density map generation solutions as the paper~\cite{zhang2016single}, smoothes each ground-truth point by a 2D Gaussian kernel. Then we calculate the total density in each patch and assign them into corresponding intervals, which are pre-defined by discretizing the whole density space.

\section{Discussion and Conclusion}

{We propose a dual-branch semi-supervised counting approach based on interleaved modelling of pixel-wise density probability distributions. The PDM loss matches the pixel-by-pixel density probability distribution to the ground truth. It shows good generalization capability, even when only a small amount of labeled data is available. Moreover, a set of tokens with clear semantic associations to the density intervals customizes the transformer decoder for the counting task. Furthermore, the inter-branch ECR term reconciles the expectations of two predicted distributions, which provides rich supervised signals for learning from unlabeled data. 
Our method compasses other methods by an average relative MAE reduction of over 22.0\%, 23.5\%, and 28.9\% with label ratios of 5\%, 10\%, and 40\% respectively. As a result, a new strong state of the art} for semi-supervised crowd counting is set up.

We also evaluate our approach under the fully-supervised setting. \XP{Our approach achieves 78.5 MAE on QNRF and thus works remarkably well under the fully supervised setting. This consistent performance boost implies that optimal semi-supervised counting is built on both the ability to learn from labeled data and unlabeled data. 
Compared with those methods focusing more on learning from unlabeled data, \ourmodel{} reaches a better balance of learning from both labeled and unlabeled data.} The limitation of our method is that ECR can alleviate, but cannot eliminate the bias in self-supervision. When the image background is too complex or the image is too crowded, it may lead to poor results. Nonetheless, this study still brings much inspiration for future studies about semi-supervised learning.


\ifCLASSOPTIONcompsoc
\section*{Acknowledgments}
\else
  \section*{Acknowledgment}
\fi

This work is funded by the National Natural Science Foundation of China (62076195, 62206271, 62376070, 12226004 and 61721002), the Fundamental Research Funds for the Central Universities (AUGA5710011522), the Guangdong Basic and Applied Basic Research Foundation (2020B1515130004), and the Shenzhen Key Technical Projects under Grant (JSGG20220831105801004, CJGJZD2022051714160501, and JCYJ20220818101406014).
\ifCLASSOPTIONcaptionsoff
  \newpage
\fi

\bibliographystyle{IEEEtran}
\bibliography{bib}

\begin{thebibliography}{10}
\providecommand{\url}[1]{#1}
\csname url@samestyle\endcsname
\providecommand{\newblock}{\relax}
\providecommand{\bibinfo}[2]{#2}
\providecommand{\BIBentrySTDinterwordspacing}{\spaceskip=0pt\relax}
\providecommand{\BIBentryALTinterwordstretchfactor}{4}
\providecommand{\BIBentryALTinterwordspacing}{\spaceskip=\fontdimen2\font plus
\BIBentryALTinterwordstretchfactor\fontdimen3\font minus \fontdimen4\font\relax}
\providecommand{\BIBforeignlanguage}[2]{{%
\expandafter\ifx\csname l@#1\endcsname\relax
\typeout{** WARNING: IEEEtran.bst: No hyphenation pattern has been}%
\typeout{** loaded for the language `#1'. Using the pattern for}%
\typeout{** the default language instead.}%
\else
\language=\csname l@#1\endcsname
\fi
#2}}
\providecommand{\BIBdecl}{\relax}
\BIBdecl

\bibitem{zhang2016single}
Y.~Zhang, D.~Zhou, S.~Chen, S.~Gao, and Y.~Ma, ``Single-image crowd counting via multi-column convolutional neural network,'' in \emph{CVPR}, 2016.

\bibitem{cao2018scale}
X.~Cao, Z.~Wang, Y.~Zhao, and F.~Su, ``Scale aggregation network for accurate and efficient crowd counting,'' in \emph{ECCV}, 2018.

\bibitem{ma2019bayesian}
Z.~Ma, X.~Wei, X.~Hong, and Y.~Gong, ``Bayesian loss for crowd count estimation with point supervision,'' in \emph{ICCV}, 2019.

\bibitem{liu2018leveraging}
X.~Liu, J.~Van De~Weijer, and A.~D. Bagdanov, ``Leveraging unlabeled data for crowd counting by learning to rank,'' in \emph{CVPR}, 2018.

\bibitem{liu2019exploiting}
------, ``Exploiting unlabeled data in cnns by self-supervised learning to rank,'' \emph{IEEE TPAMI}, 2019.

\bibitem{sindagi2020learning}
V.~A. Sindagi, R.~Yasarla, D.~S. Babu, R.~V. Babu, and V.~M. Patel, ``Learning to count in the crowd from limited labeled data,'' in \emph{ECCV}, 2020.

\bibitem{meng2021spatial}
Y.~Meng, H.~Zhang, Y.~Zhao, X.~Yang, X.~Qian, X.~Huang, and Y.~Zheng, ``Spatial uncertainty-aware semi-supervised crowd counting,'' in \emph{ICCV}, 2021.

\bibitem{wan2020modeling}
J.~Wan and A.~Chan, ``Modeling noisy annotations for crowd counting,'' \emph{NIPS}, 2020.

\bibitem{bai2020adaptive}
S.~Bai, Z.~He, Y.~Qiao, H.~Hu, W.~Wu, and J.~Yan, ``Adaptive dilated network with self-correction supervision for counting,'' in \emph{CVPR}, 2020.

\bibitem{idrees2018composition}
H.~Idrees, M.~Tayyab, K.~Athrey, D.~Zhang, S.~Al-Maadeed, N.~Rajpoot, and M.~Shah, ``Composition loss for counting, density map estimation and localization in dense crowds,'' in \emph{ECCV}, 2018.

\bibitem{sindagi2020jhu}
V.~Sindagi, R.~Yasarla, and V.~M. Patel, ``Jhu-crowd++: Large-scale crowd counting dataset and a benchmark method,'' \emph{PAMI}, 2020.

\bibitem{liu2019point}
Y.~Liu, M.~Shi, Q.~Zhao, and X.~Wang, ``Point in, box out: Beyond counting persons in crowds,'' in \emph{CVPR}, 2019.

\bibitem{liu2018decidenet}
J.~Liu, C.~Gao, D.~Meng, and A.~G. Hauptmann, ``Decidenet: Counting varying density crowds through attention guided detection and density estimation,'' in \emph{CVPR}, 2018.

\bibitem{lempitsky2010learning}
V.~Lempitsky and A.~Zisserman, ``Learning to count objects in images,'' \emph{NIPS}, 2010.

\bibitem{babu2017switching}
D.~Babu~Sam, S.~Surya, and R.~Venkatesh~Babu, ``Switching convolutional neural network for crowd counting,'' in \emph{CVPR}, 2017.

\bibitem{li2018csrnet}
Y.~Li, X.~Zhang, and D.~Chen, ``Csrnet: Dilated convolutional neural networks for understanding the highly congested scenes,'' in \emph{CVPR}, 2018.

\bibitem{zeng2017multi}
L.~Zeng, X.~Xu, B.~Cai, S.~Qiu, and T.~Zhang, ``Multi-scale convolutional neural networks for crowd counting,'' in \emph{ICIP}, 2017.

\bibitem{sindagi2019multi}
V.~A. Sindagi and V.~M. Patel, ``Multi-level bottom-top and top-bottom feature fusion for crowd counting,'' in \emph{ICCV}, 2019.

\bibitem{ma2020learning}
Z.~Ma, X.~Wei, X.~Hong, and Y.~Gong, ``Learning scales from points: A scale-aware probabilistic model for crowd counting,'' in \emph{ACM Multimedia}, 2020.

\bibitem{shi2019revisiting}
M.~Shi, Z.~Yang, C.~Xu, and Q.~Chen, ``Revisiting perspective information for efficient crowd counting,'' in \emph{CVPR}, 2019.

\bibitem{yan2019perspective}
Z.~Yan, Y.~Yuan, W.~Zuo, X.~Tan, Y.~Wang, S.~Wen, and E.~Ding, ``Perspective-guided convolution networks for crowd counting,'' in \emph{ICCV}, 2019.

\bibitem{wang2020distribution}
B.~Wang, H.~Liu, D.~Samaras, and M.~H. Nguyen, ``Distribution matching for crowd counting,'' \emph{NIPS}, 2020.

\bibitem{ma2021learning}
Z.~Ma, X.~Wei, X.~Hong, H.~Lin, Y.~Qiu, and Y.~Gong, ``Learning to count via unbalanced optimal transport,'' in \emph{AAAI}, 2021.

\bibitem{lin2021direct}
H.~Lin, X.~Hong, Z.~Ma, X.~Wei, Y.~Qiu, Y.~Wang, and Y.~Gong, ``Direct measure matching for crowd counting,'' \emph{IJCAI}, 2021.

\bibitem{xiong2019open}
H.~Xiong, H.~Lu, C.~Liu, L.~Liu, Z.~Cao, and C.~Shen, ``From open set to closed set: Counting objects by spatial divide-and-conquer,'' in \emph{ICCV}, 2019.

\bibitem{liu2019counting}
L.~Liu, H.~Lu, H.~Xiong, K.~Xian, Z.~Cao, and C.~Shen, ``Counting objects by blockwise classification,'' \emph{TCSVT}, 2019.

\bibitem{liu2020adaptive}
X.~Liu, J.~Yang, W.~Ding, T.~Wang, Z.~Wang, and J.~Xiong, ``Adaptive mixture regression network with local counting map for crowd counting,'' in \emph{ECCV}, 2020.

\bibitem{wang2021uniformity}
C.~Wang, Q.~Song, B.~Zhang, Y.~Wang, Y.~Tai, X.~Hu, C.~Wang, J.~Li, J.~Ma, and Y.~Wu, ``Uniformity in heterogeneity: Diving deep into count interval partition for crowd counting,'' in \emph{ICCV}, 2021.

\bibitem{zhao2020active}
Z.~Zhao, M.~Shi, X.~Zhao, and L.~Li, ``Active crowd counting with limited supervision,'' in \emph{ECCV}, 2020.

\bibitem{liu2020semi}
Y.~Liu, L.~Liu, P.~Wang, P.~Zhang, and Y.~Lei, ``Semi-supervised crowd counting via self-training on surrogate tasks,'' in \emph{ECCV}, 2020.

\bibitem{lin2022semi}
H.~Lin, Z.~Ma, X.~Hong, Y.~Wang, and Z.~Su, ``Semi-supervised crowd counting via density agency,'' in \emph{ACM MM}, 2022.

\bibitem{lin2023optimal}
W.~Lin and A.~B. Chan, ``Optimal transport minimization: Crowd localization on density maps for semi-supervised counting,'' in \emph{CVPR}, 2023, pp. 21\,663--21\,673.

\bibitem{zhu2023multi}
P.~Zhu, J.~Li, B.~Cao, and Q.~Hu, ``Multi-task credible pseudo-label learning for semi-supervised crowd counting,'' \emph{IEEE Transactions on Neural Networks and Learning Systems}, 2023.

\bibitem{yang2020weakly}
Y.~Yang, G.~Li, Z.~Wu, L.~Su, Q.~Huang, and N.~Sebe, ``Weakly-supervised crowd counting learns from sorting rather than locations,'' in \emph{ECCV}, 2020.

\bibitem{lei2021towards}
Y.~Lei, Y.~Liu, P.~Zhang, and L.~Liu, ``Towards using count-level weak supervision for crowd counting,'' \emph{Pattern Recognition}, 2021.

\bibitem{sindagi2019ha}
V.~A. Sindagi and V.~M. Patel, ``Ha-ccn: Hierarchical attention-based crowd counting network,'' \emph{IEEE Transactions on Image Processing}, 2019.

\bibitem{han2020focus}
T.~Han, J.~Gao, Y.~Yuan, and Q.~Wang, ``Focus on semantic consistency for cross-domain crowd understanding,'' in \emph{ICASSP 2020-2020 IEEE International Conference on Acoustics, Speech and Signal Processing (ICASSP)}.\hskip 1em plus 0.5em minus 0.4em\relax IEEE, 2020, pp. 1848--1852.

\bibitem{liu2022leveraging}
W.~Liu, N.~Durasov, and P.~Fua, ``Leveraging self-supervision for cross-domain crowd counting,'' in \emph{CVPR}, 2022, pp. 5341--5352.

\bibitem{wang2019learning}
Q.~Wang, J.~Gao, W.~Lin, and Y.~Yuan, ``Learning from synthetic data for crowd counting in the wild,'' in \emph{Proceedings of the IEEE/CVF conference on computer vision and pattern recognition}, 2019, pp. 8198--8207.

\bibitem{dosovitskiy2020image}
A.~Dosovitskiy, L.~Beyer, A.~Kolesnikov, D.~Weissenborn, X.~Zhai, T.~Unterthiner, M.~Dehghani, M.~Minderer, G.~Heigold, S.~Gelly \emph{et~al.}, ``An image is worth 16x16 words: Transformers for image recognition at scale,'' in \emph{ICLR}, 2020.

\bibitem{vaswani2017attention}
A.~Vaswani, N.~Shazeer, N.~Parmar, J.~Uszkoreit, L.~Jones, A.~N. Gomez, {\L}.~Kaiser, and I.~Polosukhin, ``Attention is all you need,'' in \emph{NIPS}, 2017.

\bibitem{carion2020end}
N.~Carion, F.~Massa, G.~Synnaeve, N.~Usunier, A.~Kirillov, and S.~Zagoruyko, ``End-to-end object detection with transformers,'' in \emph{ECCV}, 2020.

\bibitem{zhu2020deformable}
X.~Zhu, W.~Su, L.~Lu, B.~Li, X.~Wang, and J.~Dai, ``Deformable detr: Deformable transformers for end-to-end object detection,'' in \emph{ICLR}, 2020.

\bibitem{zheng2020end}
M.~Zheng, P.~Gao, X.~Wang, H.~Li, and H.~Dong, ``End-to-end object detection with adaptive clustering transformer,'' \emph{arXiv preprint}, 2020.

\bibitem{sun2021rethinking}
Z.~Sun, S.~Cao, Y.~Yang, and K.~M. Kitani, ``Rethinking transformer-based set prediction for object detection,'' in \emph{ICCV}, 2021.

\bibitem{zheng2021rethinking}
S.~Zheng, J.~Lu, H.~Zhao, X.~Zhu, Z.~Luo, Y.~Wang, Y.~Fu, J.~Feng, T.~Xiang, P.~H. Torr \emph{et~al.}, ``Rethinking semantic segmentation from a sequence-to-sequence perspective with transformers,'' in \emph{CVPR}, 2021.

\bibitem{wang2021end}
Y.~Wang, Z.~Xu, X.~Wang, C.~Shen, B.~Cheng, H.~Shen, and H.~Xia, ``End-to-end video instance segmentation with transformers,'' in \emph{CVPR}, 2021.

\bibitem{strudel2021segmenter}
R.~Strudel, R.~Garcia, I.~Laptev, and C.~Schmid, ``Segmenter: Transformer for semantic segmentation,'' \emph{arXiv preprint}, 2021.

\bibitem{cheng2021per}
B.~Cheng, A.~Schwing, and A.~Kirillov, ``Per-pixel classification is not all you need for semantic segmentation,'' \emph{NIPS}, 2021.

\bibitem{chen2021transformer}
X.~Chen, B.~Yan, J.~Zhu, D.~Wang, X.~Yang, and H.~Lu, ``Transformer tracking,'' in \emph{CVPR}, 2021.

\bibitem{wang2021transformer}
N.~Wang, W.~Zhou, J.~Wang, and H.~Li, ``Transformer meets tracker: Exploiting temporal context for robust visual tracking,'' in \emph{CVPR}, 2021.

\bibitem{sun2020transtrack}
P.~Sun, Y.~Jiang, R.~Zhang, E.~Xie, J.~Cao, X.~Hu, T.~Kong, Z.~Yuan, C.~Wang, and P.~Luo, ``Transtrack: Multiple-object tracking with transformer,'' \emph{arXiv preprint}, 2020.

\bibitem{lin2022boosting}
H.~Lin, Z.~Ma, R.~Ji, Y.~Wang, and X.~Hong, ``Boosting crowd counting via multifaceted attention,'' in \emph{CVPR}, 2022.

\bibitem{wei2021scene}
X.~Wei, Y.~Kang, J.~Yang, Y.~Qiu, D.~Shi, W.~Tan, and Y.~Gong, ``Scene-adaptive attention network for crowd counting,'' \emph{arXiv preprint}, 2021.

\bibitem{liang2021transcrowd}
D.~Liang, X.~Chen, W.~Xu, Y.~Zhou, and X.~Bai, ``Transcrowd: Weakly-supervised crowd counting with transformer,'' \emph{arXiv preprint}, 2021.

\bibitem{su2015distances}
H.~Su and H.~Zhang, ``Distances and kernels based on cumulative distribution functions,'' in \emph{Emerging Trends in Image Processing, Computer Vision and Pattern Recognition}.\hskip 1em plus 0.5em minus 0.4em\relax Elsevier, 2015, pp. 551--559.

\bibitem{chun2000uncertainty}
M.-H. Chun, S.-J. Han, and N.-I. Tak, ``An uncertainty importance measure using a distance metric for the change in a cumulative distribution function,'' \emph{Reliability Engineering \& System Safety}, vol.~70, no.~3, pp. 313--321, 2000.

\bibitem{kolouri2018sliced}
S.~Kolouri, P.~E. Pope, C.~E. Martin, and G.~K. Rohde, ``Sliced-wasserstein autoencoder: An embarrassingly simple generative model,'' \emph{arXiv preprint}, 2018.

\bibitem{de20211}
M.~De~Angelis and A.~Gray, ``Why the 1-wasserstein distance is the area between the two marginal cdfs,'' \emph{arXiv preprint arXiv:2111.03570}, 2021.

\bibitem{tarvainen2017mean}
A.~Tarvainen and H.~Valpola, ``Mean teachers are better role models: Weight-averaged consistency targets improve semi-supervised deep learning results,'' \emph{NIPS}, 2017.

\bibitem{ke2019dual}
Z.~Ke, D.~Wang, Q.~Yan, J.~Ren, and R.~W. Lau, ``Dual student: Breaking the limits of the teacher in semi-supervised learning,'' in \emph{ICCV}, 2019.

\bibitem{wang2020nwpu}
Q.~Wang, J.~Gao, W.~Lin, and X.~Li, ``Nwpu-crowd: A large-scale benchmark for crowd counting and localization,'' \emph{PAMI}, 2020.

\bibitem{kingma2014adam}
D.~P. Kingma and J.~Ba, ``Adam: A method for stochastic optimization,'' \emph{arXiv preprint}, 2014.

\bibitem{shao2018crowdhuman}
S.~Shao, Z.~Zhao, B.~Li, T.~Xiao, G.~Yu, X.~Zhang, and J.~Sun, ``Crowdhuman: A benchmark for detecting human in a crowd,'' \emph{arXiv preprint arXiv:1805.00123}, 2018.

\bibitem{song2021rethinking}
Q.~Song, C.~Wang, Z.~Jiang, Y.~Wang, Y.~Tai, C.~Wang, J.~Li, F.~Huang, and Y.~Wu, ``Rethinking counting and localization in crowds: A purely point-based framework,'' in \emph{ICCV}, 2021.

\end{thebibliography}

%





\clearpage

\section{Appendix}
In the appendix, we detailed alternative structures of dual-branch with tokens (Sec.~\ref{sec:inde}). These alternatives and corresponding experimental results demonstrate the independence and effectiveness of \ourmodel{} structure. Furthermore, we give the pseudo code of \ourmodel{} learning process (Sec.~\ref{sec:code}) and provide visualizations of attention maps (Sec.~\ref{sec:visualizationapp}).

\subsection{Detailed Alternative Structure of Dual-Branch with Tokens}\label{sec:inde}

To demonstrate the importance of \XP{the independence of model structures,} we introduce four alternative structures. Specifically, they have the same CNN backbone, Pixel-wise Distribution Matching loss and inter-branch Expectation Consistency Regularization as \ourmodel{}. A general comparison of these structures is shown in Table~\ref{tab:alternatives}. And the experimental results are shown in Table~\ref{tab:inde}. 

\def\arraystretch{1.1}
\renewcommand{\tabcolsep}{12 pt}{
\begin{table*}[thbp!]
\small
\begin{center}
\begin{tabular}{c|ccccc}
\toprule[1pt]
 & \ourmodel{} & SDDS & STDS & STSS & DTSS\\
\midrule
Two Independent Decoders & $\checkmark$ & & $\checkmark$ & $\checkmark$ & $\checkmark$\\
Two Series of Tokens & $\checkmark$ & $\checkmark$ & & & $\checkmark$\\
Interleaving Density Semantics & $\checkmark$ & $\checkmark$ & $\checkmark$ &  & \\
\toprule[1pt]
\end{tabular}
\end{center}
\caption{A comparison of structures for different alternatives.}
\label{tab:alternatives}
\end{table*}}

\begin{figure}[t]
\begin{center}
    \includegraphics[width=0.5\textwidth]{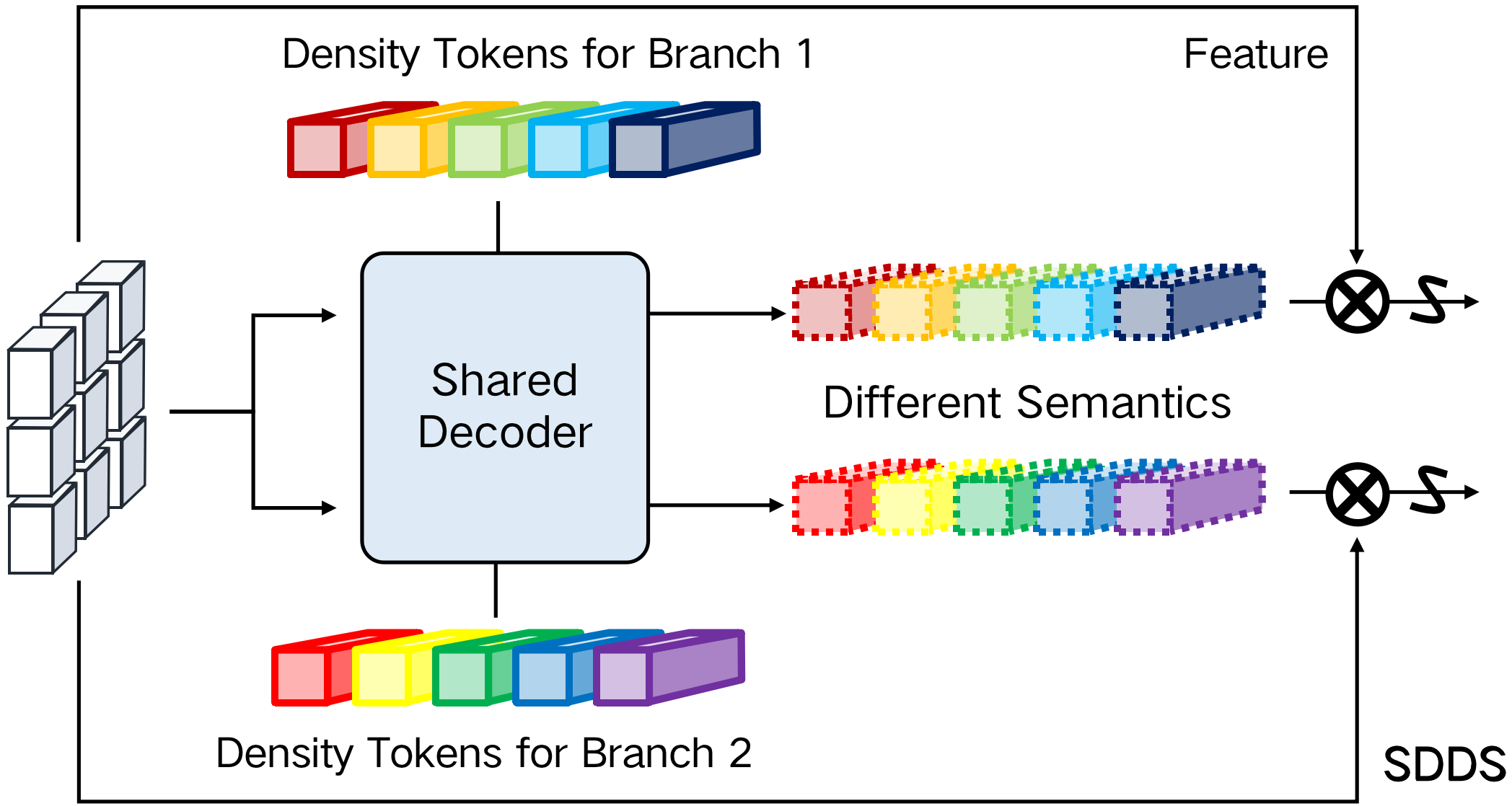}
\end{center}
\caption{SDDS uses a shared decoder with two independent sets of tokens for interleaving intervals.}
\label{fig:sdds}
\end{figure}

\begin{figure}[t]
\begin{center}
    \includegraphics[width=0.5\textwidth]{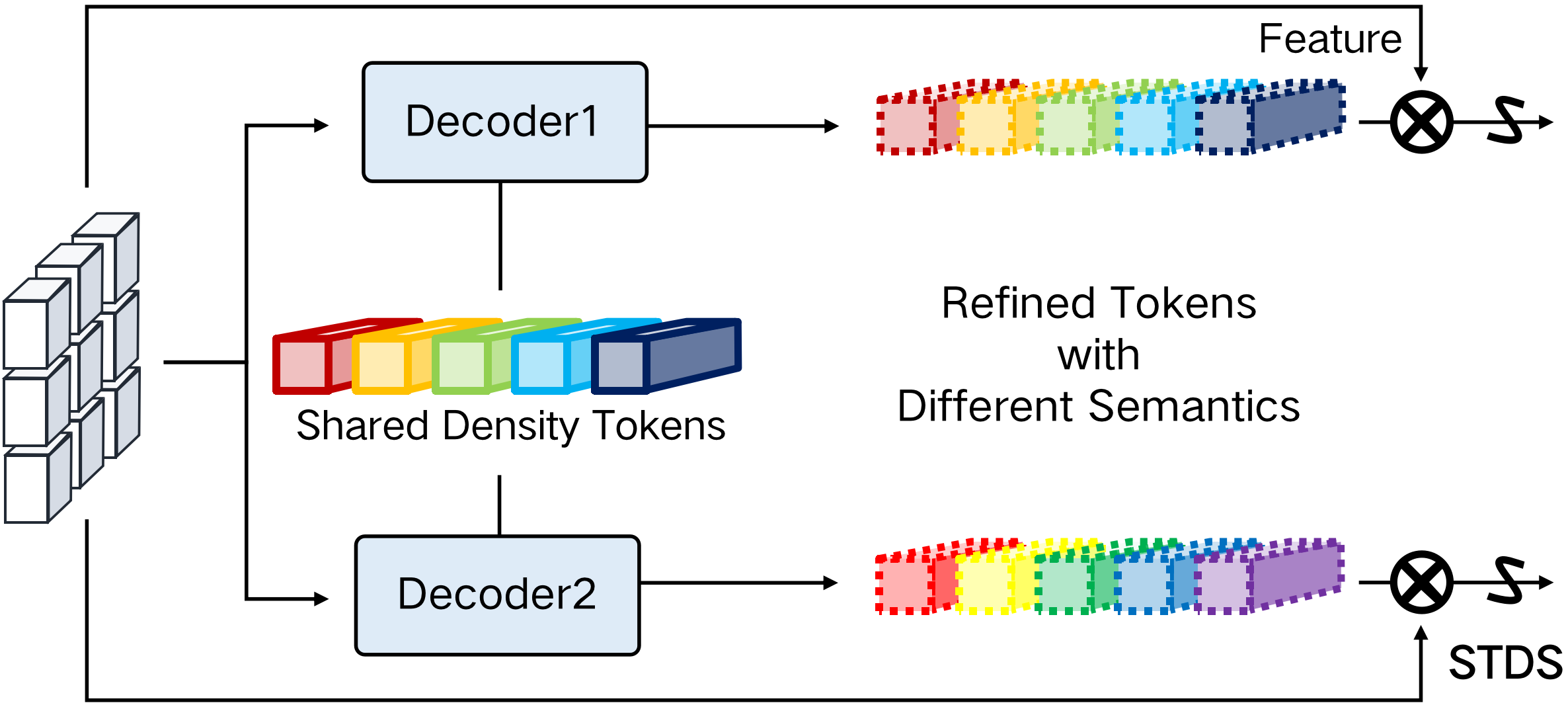}
\end{center}
\caption{STDS uses two independent decoders with a shared set of tokens for interleaving intervals.}
\label{fig:stds}
\end{figure}

\begin{figure}[t]
\begin{center}
    \includegraphics[width=0.5\textwidth]{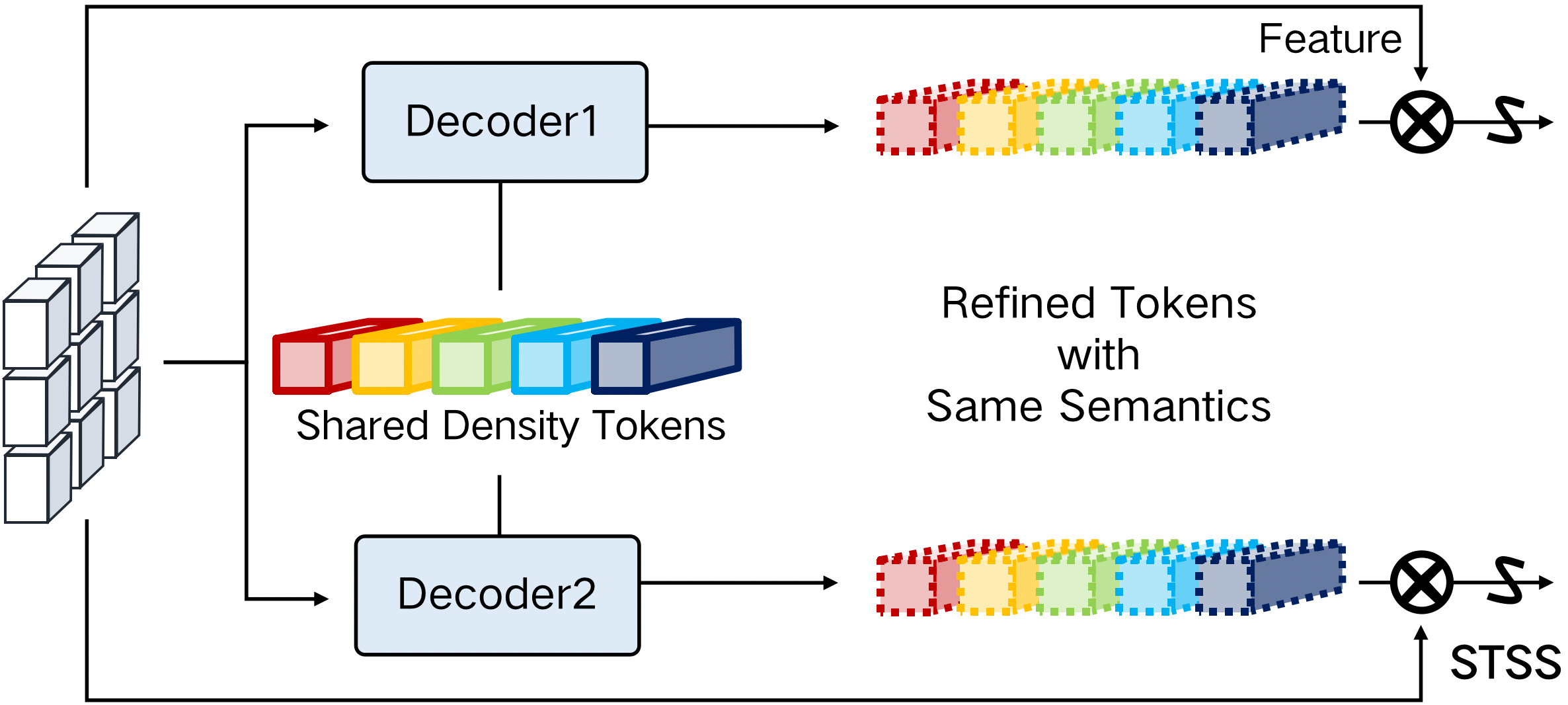}
\end{center}
\caption{STSS uses two independent decoders with a shared set of tokens for the same intervals.}
\label{fig:stss}
\end{figure}

\begin{figure}[t]
\begin{center}
    \includegraphics[width=0.5\textwidth]{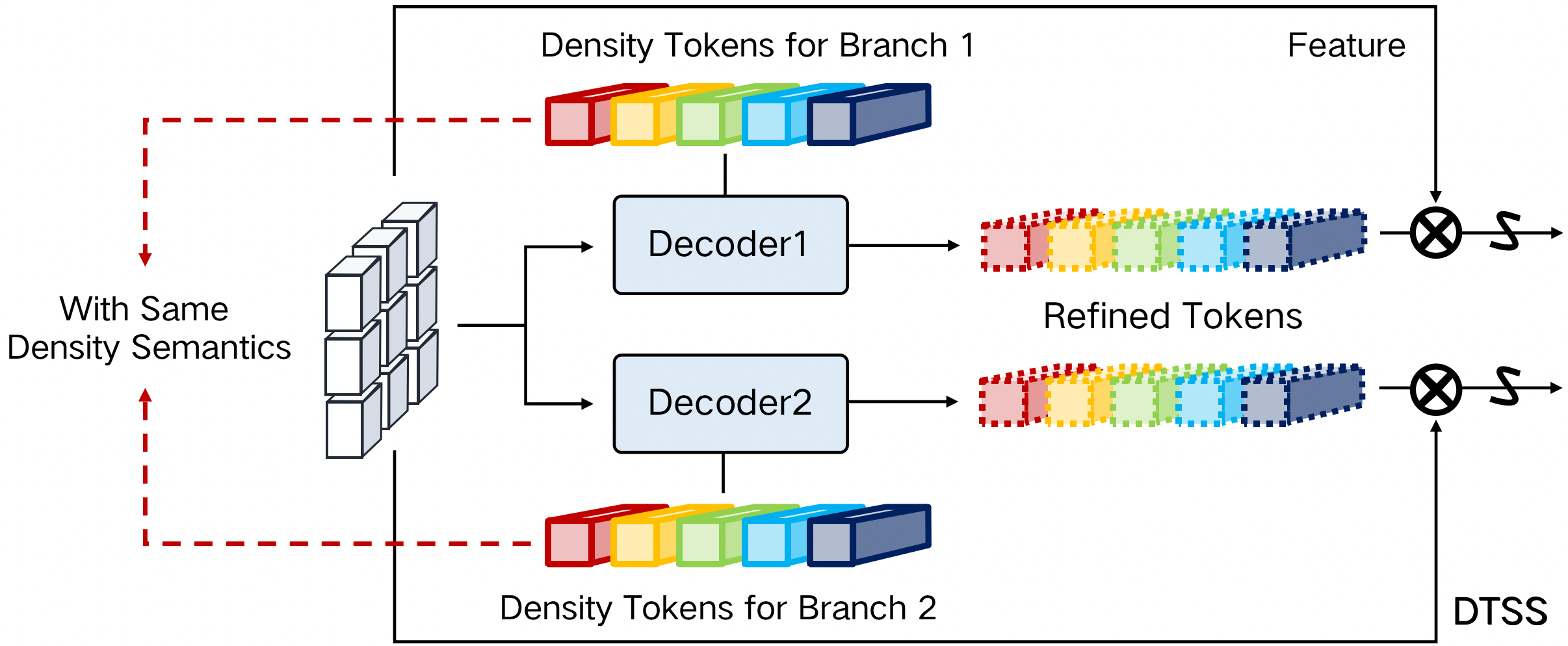}
\end{center}
\caption{DTSS uses two independent decoders with two independent sets of tokens for the same intervals.}
\label{fig:dtss}
\end{figure}

\begin{itemize}
    \item `Shared Decoder with Different Semantics (SDDS)' uses a common decoder to refine two interleaving and independent density tokens. Compared to \ourmodel{}, this alternative misses the independence of dual decoders for separate interaction between the feature map and the respective density tokens. The structure is shown in Figure~\ref{fig:sdds}. Since each token maintains exclusive semantic information, this structure still retains a certain degree of independence.
    
    \item `Shared Tokens with Different Semantics (STDS)' uses independent dual decoders and a common series of density tokens with different semantics. The structure is shown in Figure~\ref{fig:stds}. The alternative lacks the independence of different series of token features to represent interleaving density information. Instead, different semantics are endowed to the different decoders. Through the modulation and interaction of dual decoders, the refined tokens will fine-tune to be with interleaving density information. The drop in accuracy is attributed to the lack of semantic difference of initial density token prototypes in two interleaving branches.
    
    \item `Shared Tokens with Same Semantics (STSS)' uses independent dual decoders but a common series of density tokens with the same semantics. This structure, which is shown in Figure~\ref{fig:stss}, lacks the independence of different series of representative token features and the interleaving density support. However, since the initialization of parameters in dual decoders is different, the confirmation bias will be partially eliminated, so the consistency regularization can still play a slight role.
    
    \item `Different Tokens with Same Semantics (DTSS)' adopts the same independent model with \ourmodel{} but the dual density tokens represent the same density intervals. This alternative lacks the assumption that the semantics are exclusive and build on different supports, thus the consistency regularization is easier to satisfy. The structure is shown in Figure~\ref{fig:dtss} and its semi-supervised performance deteriorates compared to that of \ourmodel{}.
    
\end{itemize}

\def\arraystretch{1.1}
\renewcommand{\tabcolsep}{20 pt}{
\begin{table*}[thbp!]
\small
\begin{center}
\begin{tabular}{c|c|c|c|c|c}
  \toprule[1pt]
  & SDDS & STDS & STSS & DTSS & \ourmodel{} \\
  \midrule
  MAE & 123.3 & 121.4 & 123.7 & 123.1 & \textbf{115.3} \\
  MSE & 215.7 & 197.8 & 213.4 & 206.0 & \textbf{195.2} \\
  \toprule[1pt]
\end{tabular}
\caption{The impact of \XP{the settings of the} consistency regularization with different structures. Experiments are conducted on UCF-QNRF with a labeled ratio $5\%$.}
\label{tab:inde}
\end{center}
\end{table*}}

\subsection{Pseudo Code}\label{sec:code}
We provide a pseudo code for \ourmodel{} learning in Algorithm~\ref{algo:model}.

\begin{algorithm}
\caption{\ourmodel{} Learning}\label{algo:model}
\LinesNumbered 
\KwIn{Labeled dataset $\mathcal{X}$ and unlabeled dataset $\mathcal{U}$.}
\KwOut{The counting model $\theta$ with dual density tokens $T={T_1, T_2}$.}
Initialize $\mathcal{L} \leftarrow 0$\;
\For{{\rm epoch in} $\left[1, maxepoch\right]$}{
\For{{\rm each sample} $s \in \mathcal{X} \cup \mathcal{U}$}{
Get the predicted distribution matrices $O_1, O_2$ by corresponding $T_1, T_2$\;
\eIf{$s \in \mathcal{X}$}{
Generate the training labels $Y_1, Y_2$ by ground-truth\;
$\mathcal{L} \leftarrow \mathcal{L}_P$ by calculating the PDM loss based on Eq. 7\;}{
Generate the pixel-wise mask $\mathcal{E}$ based on Eq. 9\;
$\mathcal{L} \leftarrow \lambda \mathcal{L}_E$ by calculating the ECR loss based on Eq. 8\;
}
Update the counting model $R$ and density tokens $T$ minimizing $L$\;}}
\textbf{Return}\ {the trained counting model $\theta^\prime$ with $T^\prime$.}
\end{algorithm}

\subsection{Visualizations}\label{sec:visualizationapp}
We visualize the attention maps for each density token in dual branch to study their effects. Visualizations are shown in Figure~\ref{fig:viz} for a labeled image and Figure~\ref{fig:viz2} for an unlabeled image. 

The attention maps are generated by the same model, which is trained on UCF-QNRF with a labeled ratio of $5\%$. A and B stand for different branch and the numbers represent the different tokens. Tokens with higher numbers specify the density interval with higher density. The quantitative results show that the density tokens work well whether the predicted density was supervised by a clear groud-truth label or not.

\renewcommand{\tabcolsep}{5 pt}{
\begin{figure*}[t!]
	\begin{center}
		\begin{tabular}{cccccc}
			
			\includegraphics[width=0.14\linewidth]{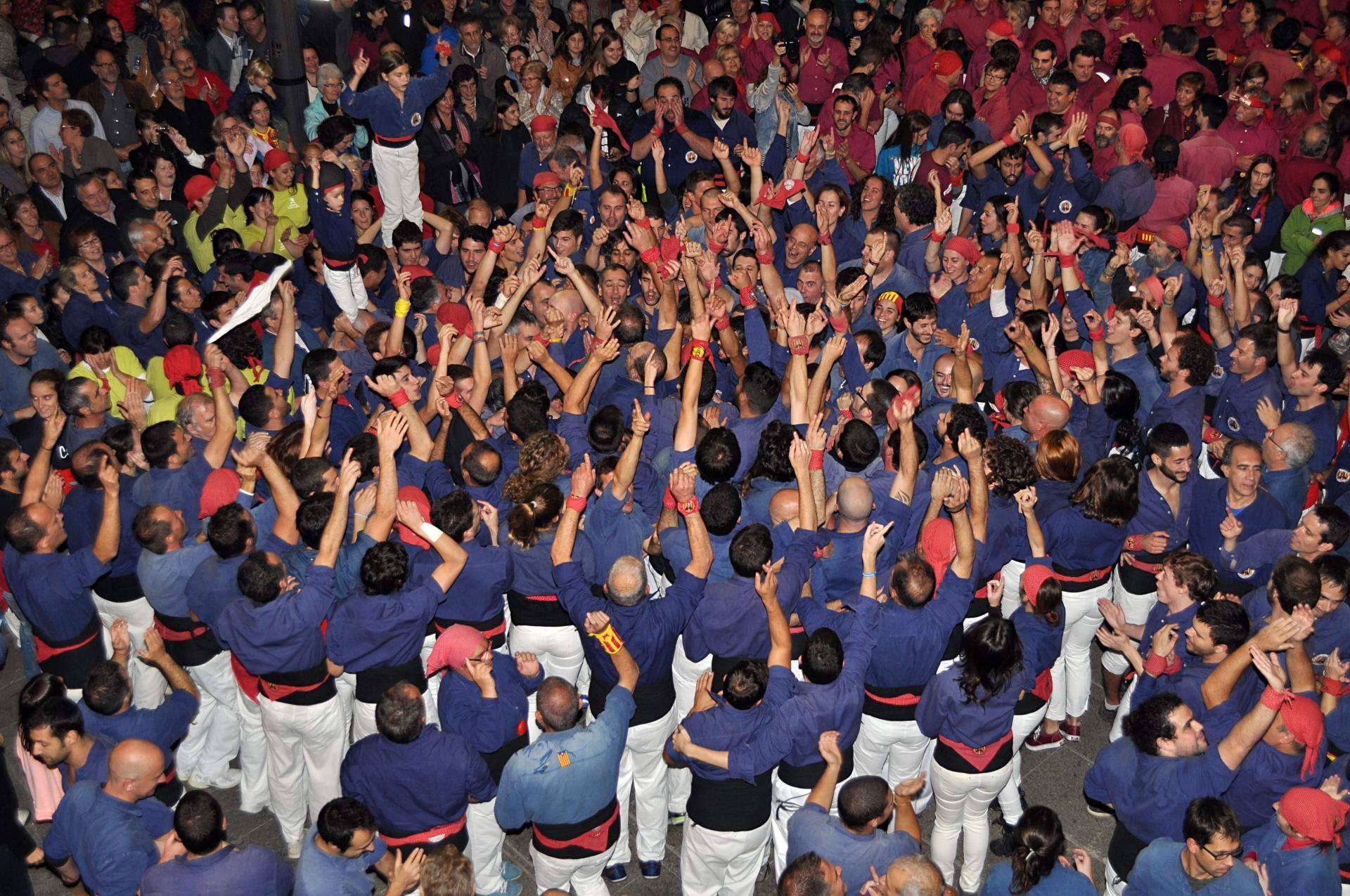}  &
			\includegraphics[width=0.14\linewidth]{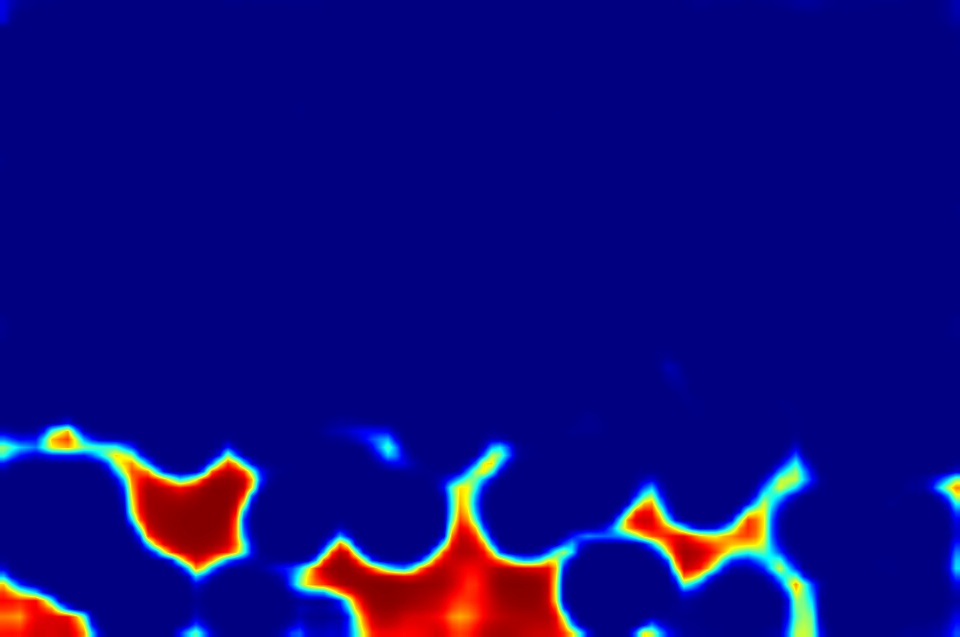}  &
			\includegraphics[width=0.14\linewidth]{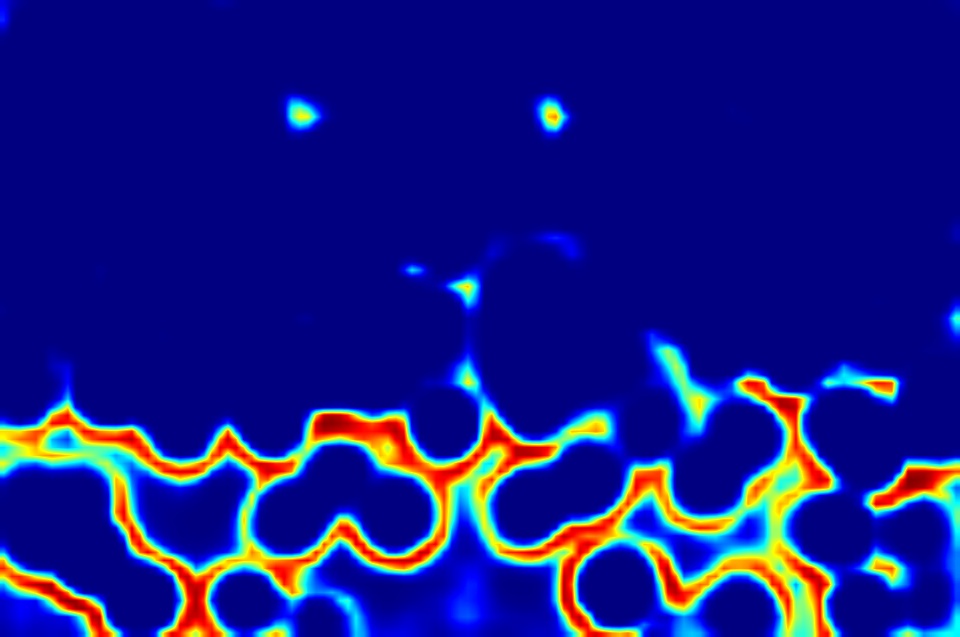} &
			\includegraphics[width=0.14\linewidth]{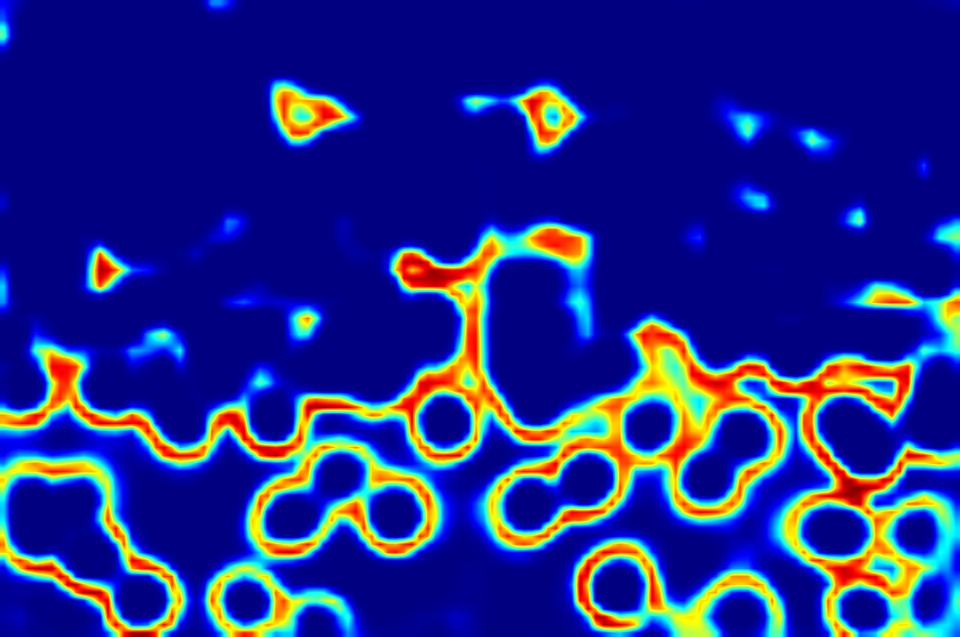}  &
			\includegraphics[width=0.14\linewidth]{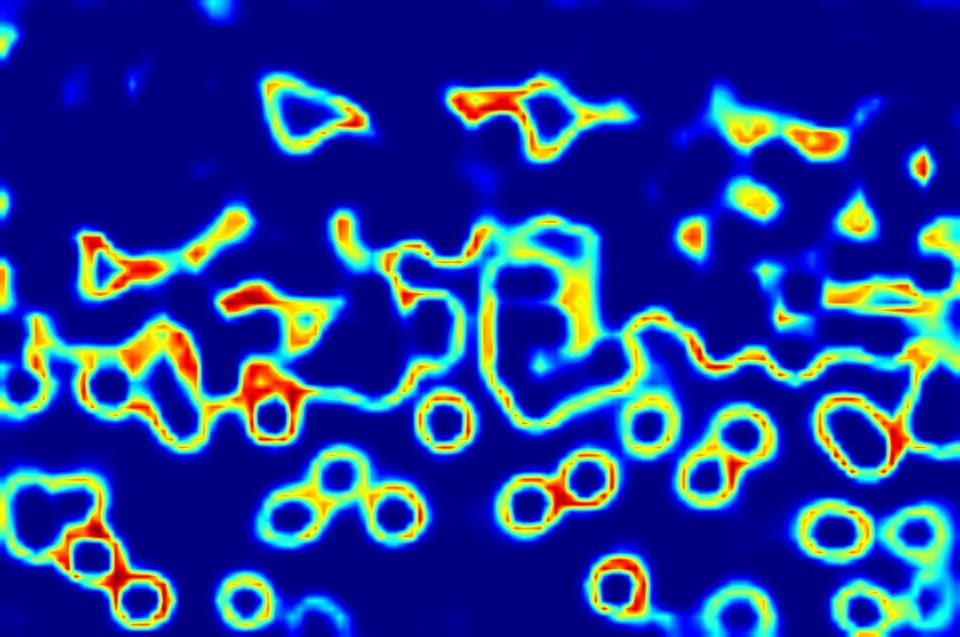} &
			\includegraphics[width=0.14\linewidth]{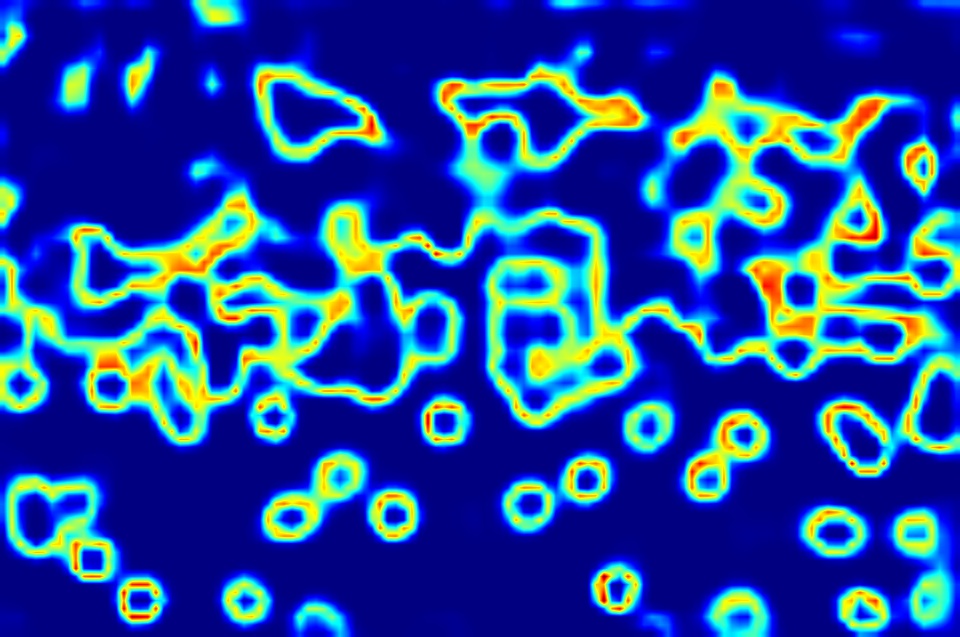} \\
			
			\footnotesize{Labeled Image} & \footnotesize{A1} & \footnotesize{A2} & \footnotesize{A3} & \footnotesize{A4} & \footnotesize{A5} \\
			
			\includegraphics[width=0.14\linewidth]{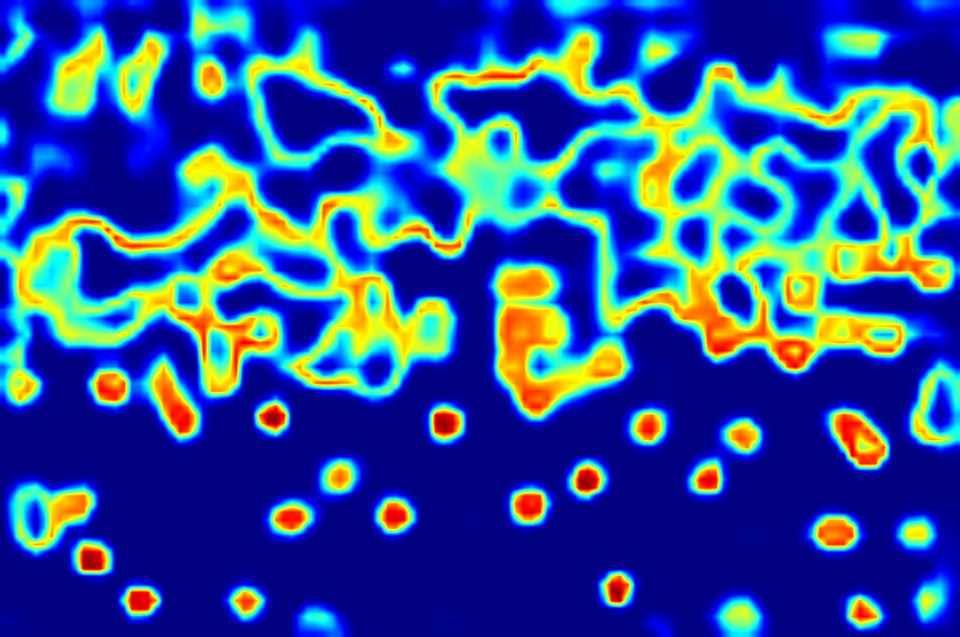}  &
			\includegraphics[width=0.14\linewidth]{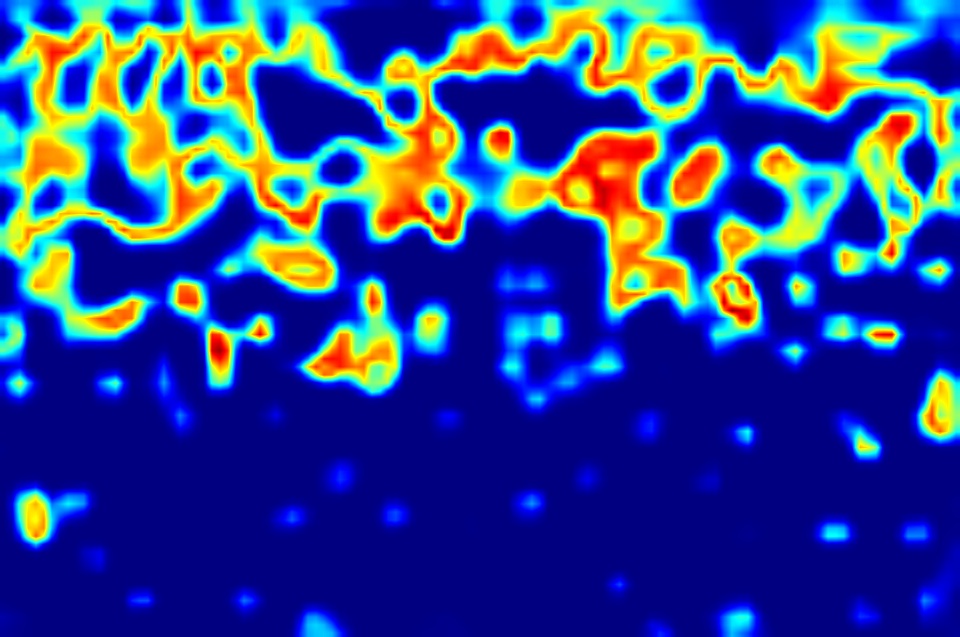}  &
			\includegraphics[width=0.14\linewidth]{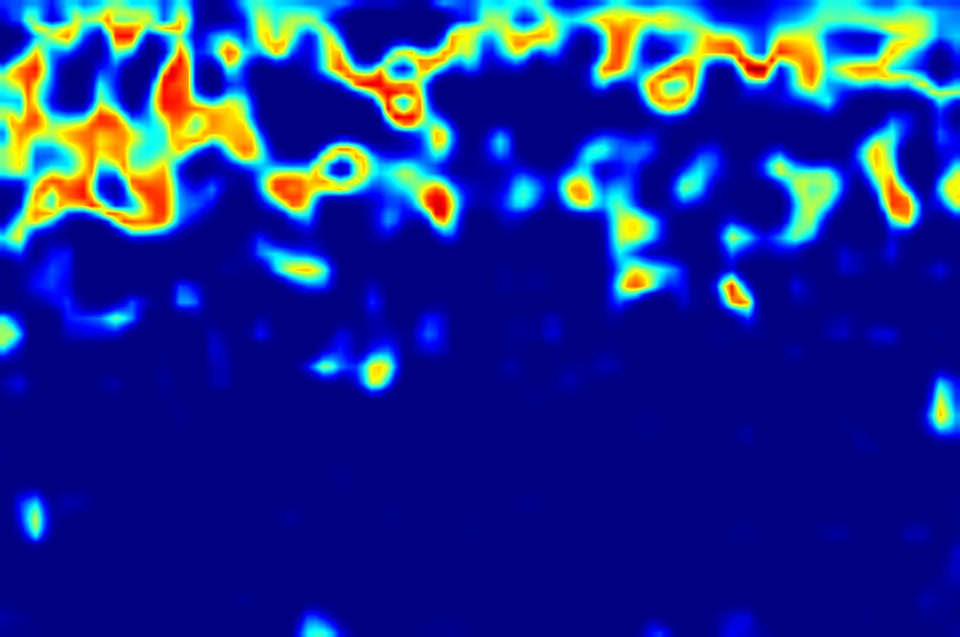} &
			\includegraphics[width=0.14\linewidth]{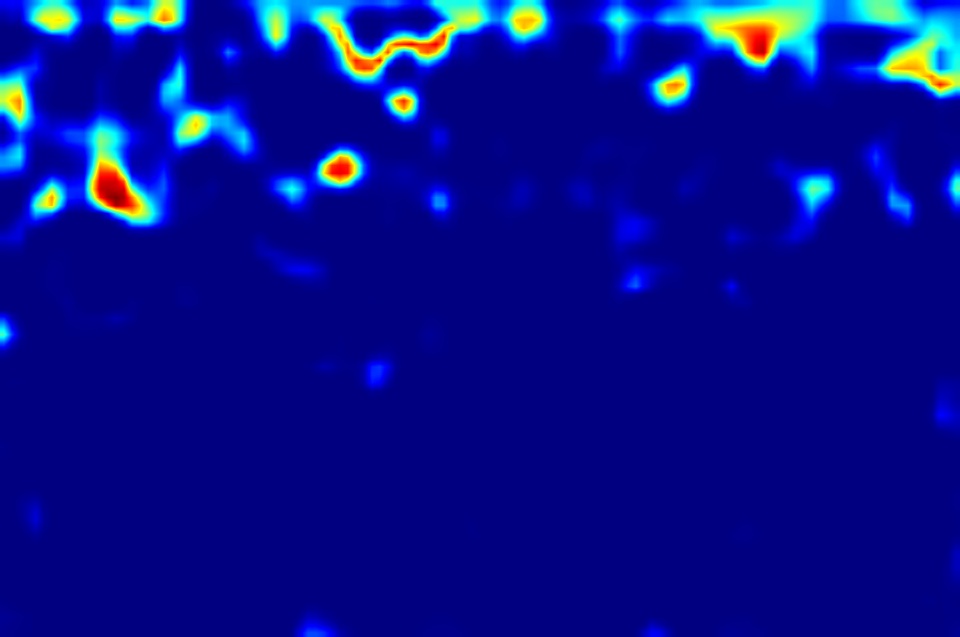}  &
			\includegraphics[width=0.14\linewidth]{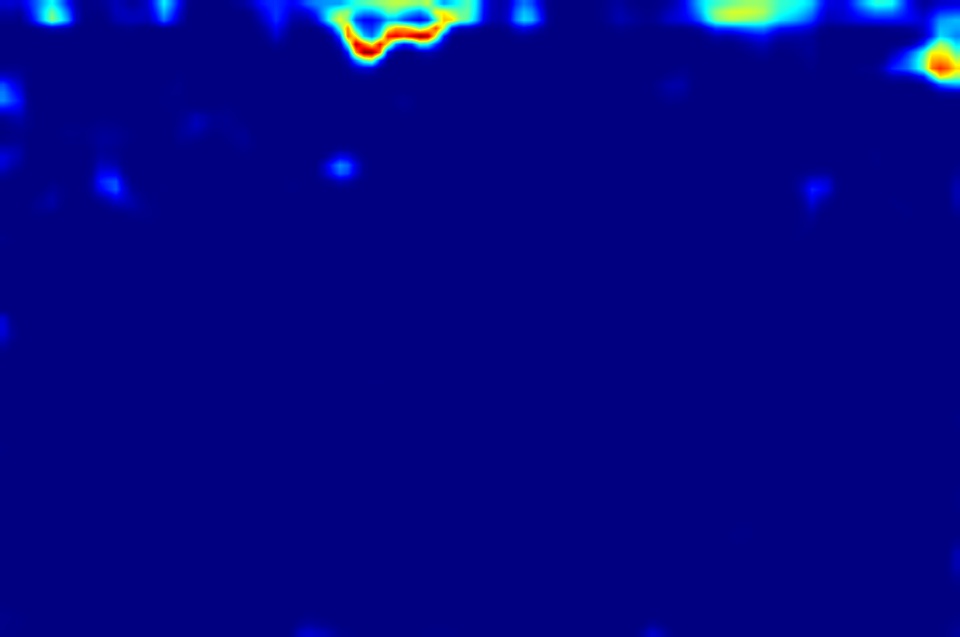} &
			\includegraphics[width=0.14\linewidth]{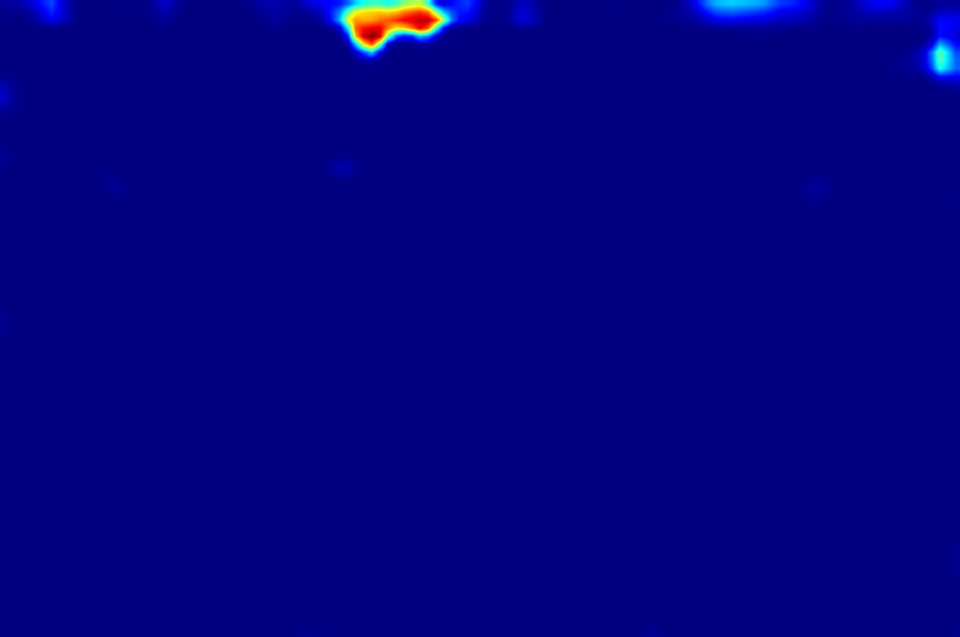} \\
			
			\footnotesize{A6} & \footnotesize{A7} & \footnotesize{A8} & \footnotesize{A9} & \footnotesize{A10} & \footnotesize{A11} \\
			
			\includegraphics[width=0.14\linewidth]{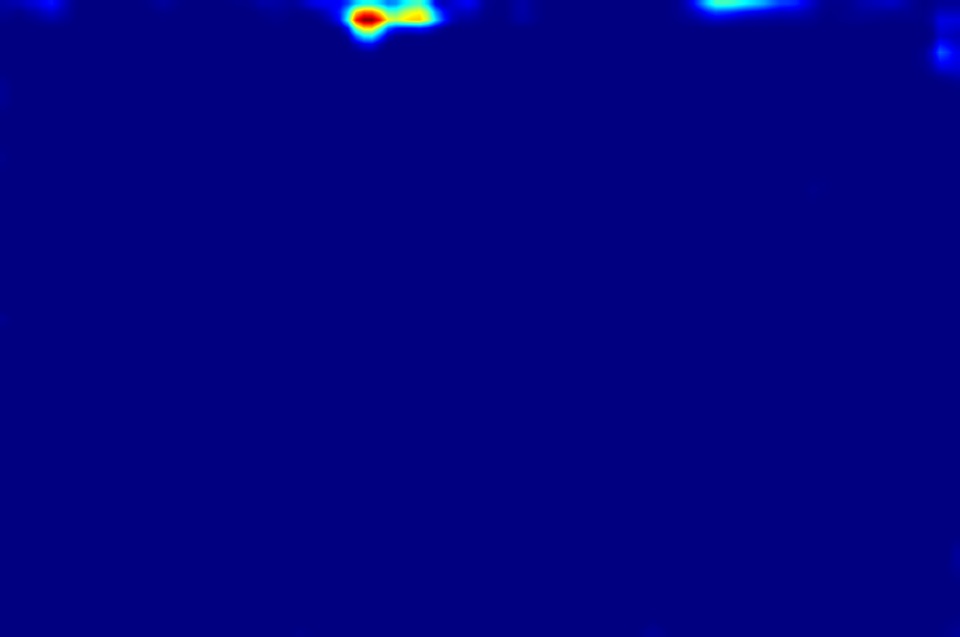}  &
			\includegraphics[width=0.14\linewidth]{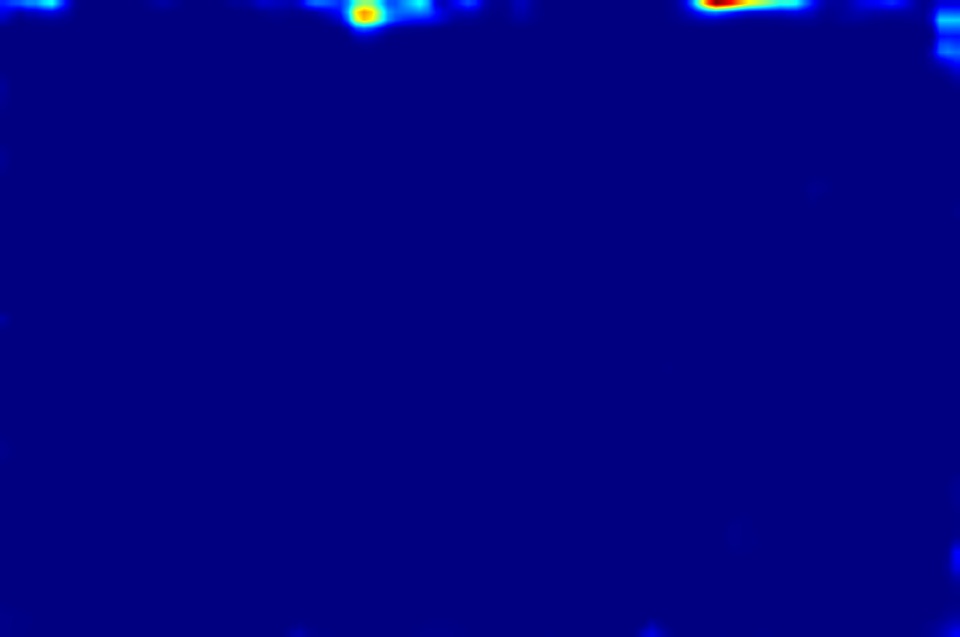}  &
			\includegraphics[width=0.14\linewidth]{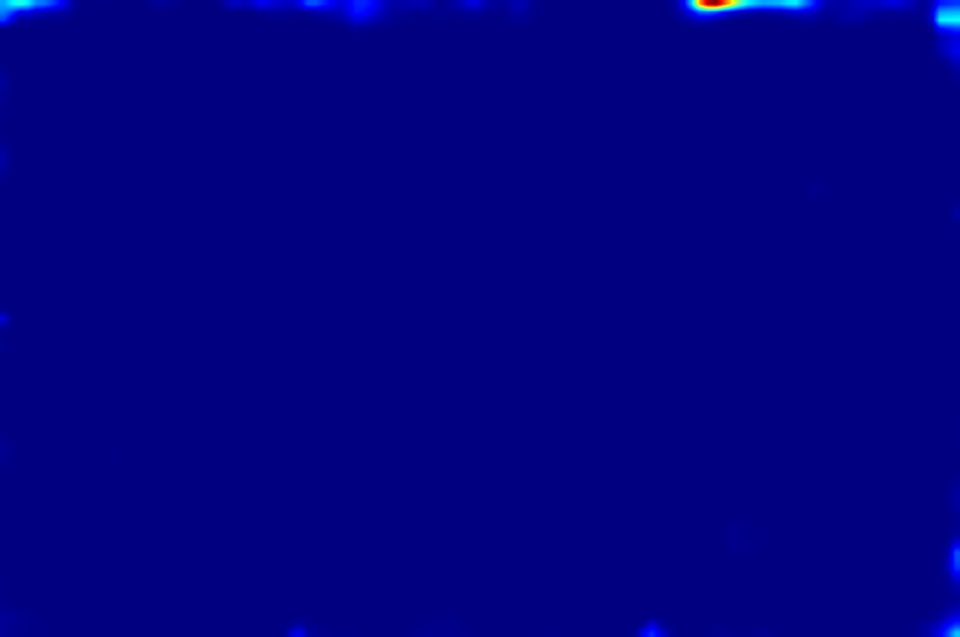} &
			\includegraphics[width=0.14\linewidth]{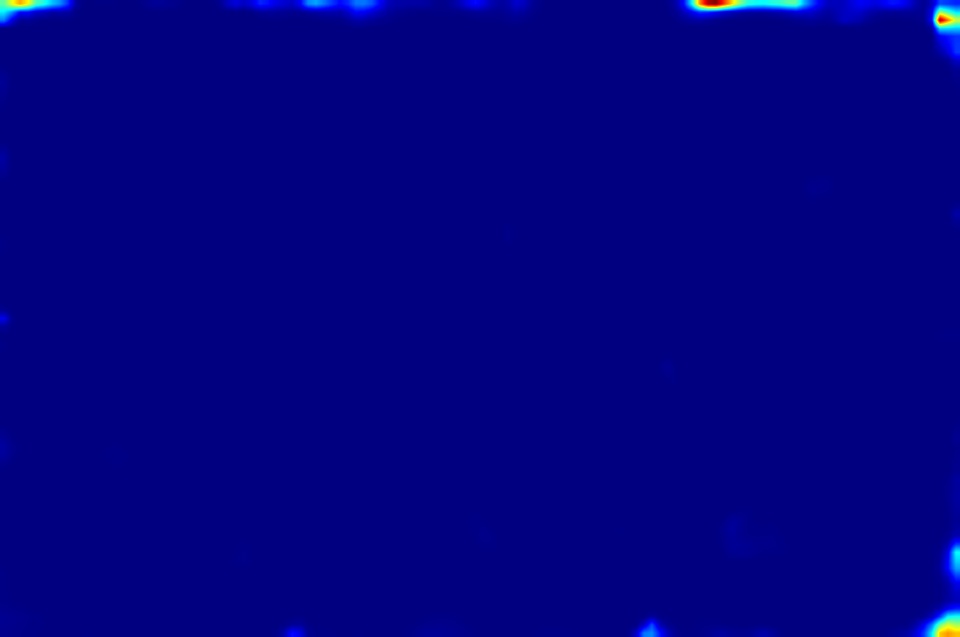}  &
			\includegraphics[width=0.14\linewidth]{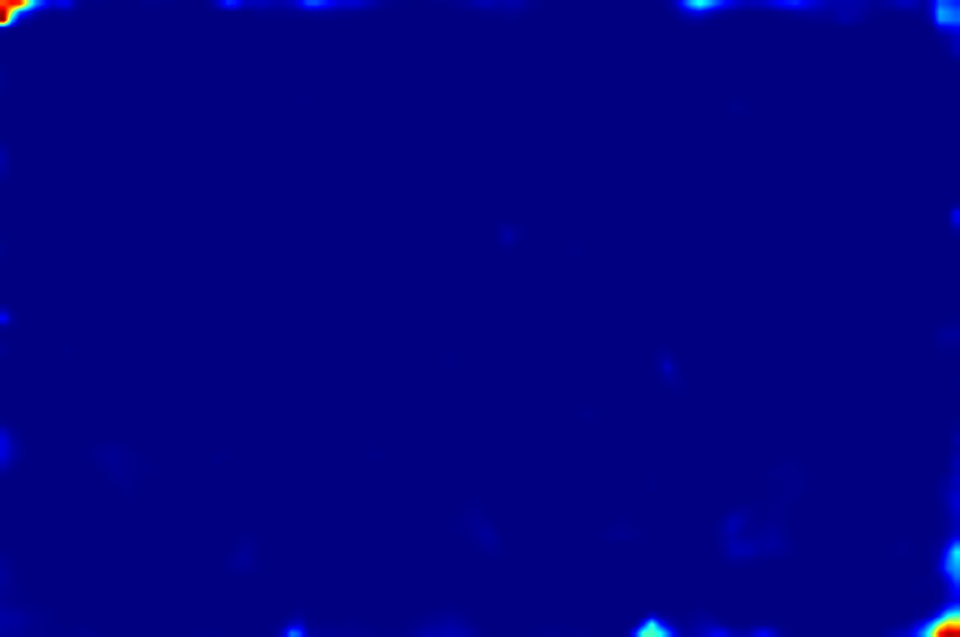} &
			\includegraphics[width=0.14\linewidth]{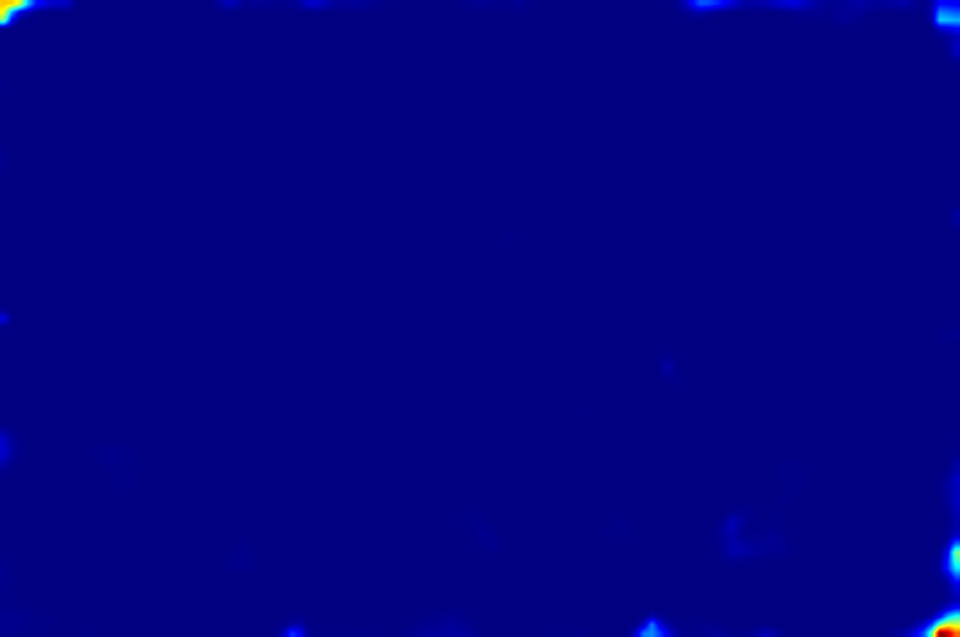} \\
			
			\footnotesize{A12} & \footnotesize{A13} & \footnotesize{A14} & \footnotesize{A15} & \footnotesize{A16} & \footnotesize{A17} \\
			
			\includegraphics[width=0.14\linewidth]{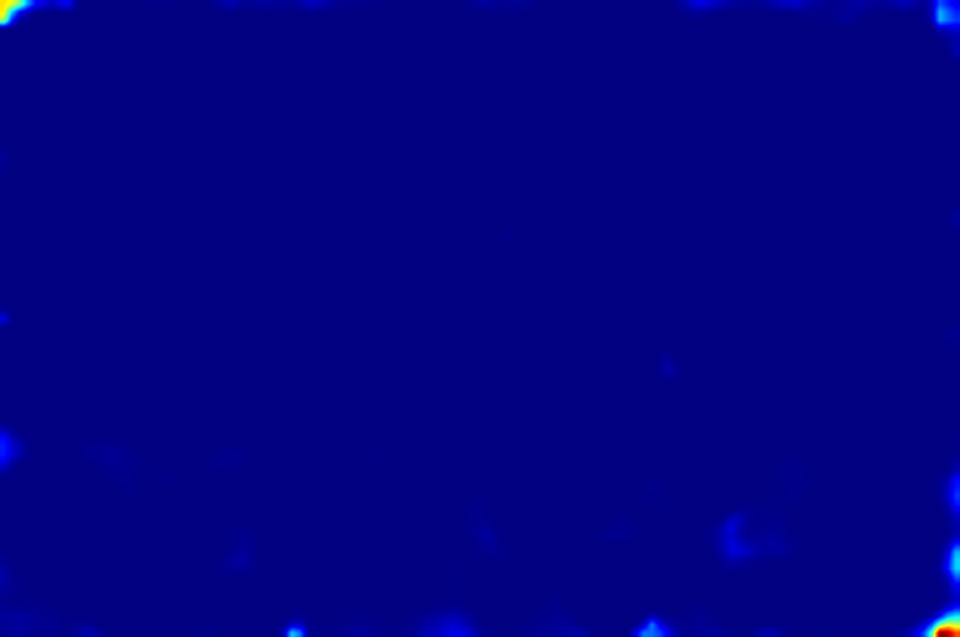}  &
			\includegraphics[width=0.14\linewidth]{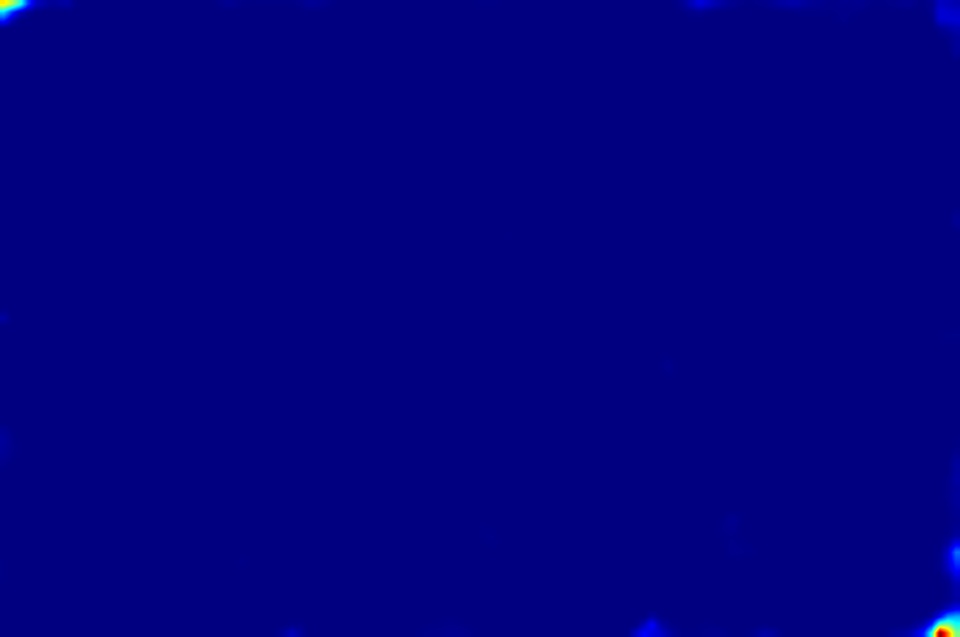}  &
			\includegraphics[width=0.14\linewidth]{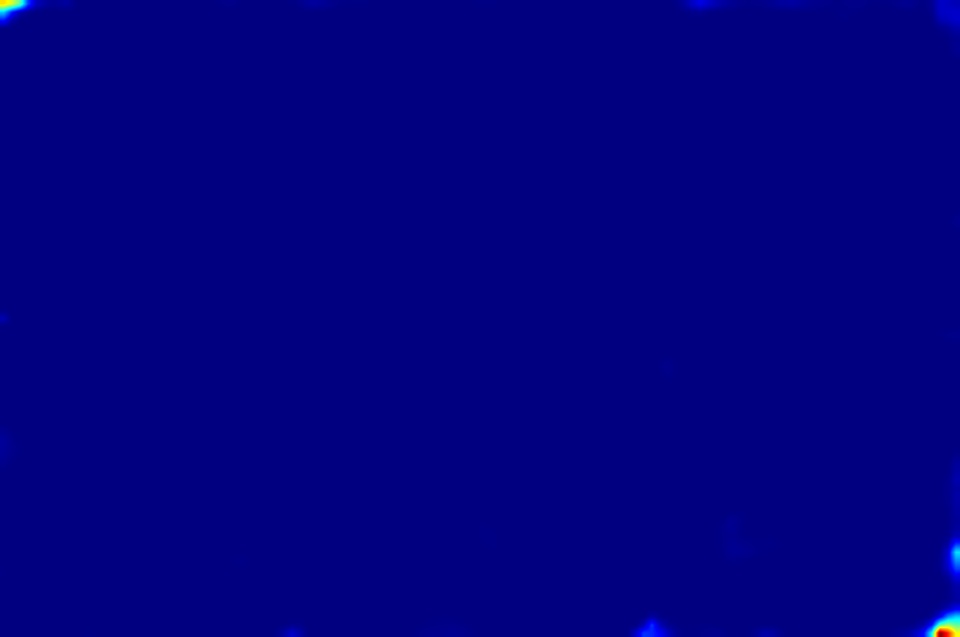} &
			\includegraphics[width=0.14\linewidth]{vis1_306_10.jpg}  &
			\includegraphics[width=0.14\linewidth]{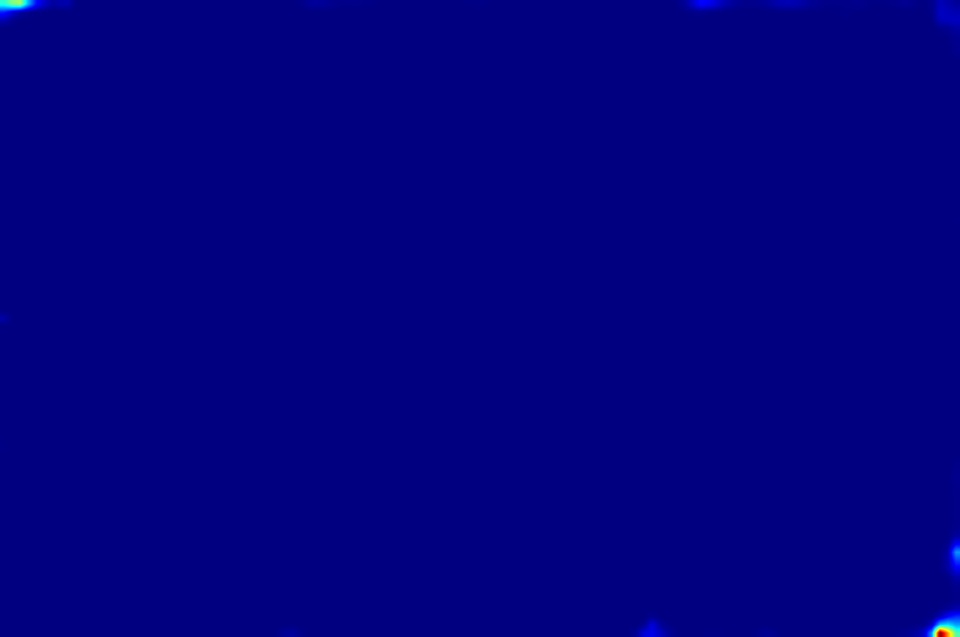} &
			\includegraphics[width=0.14\linewidth]{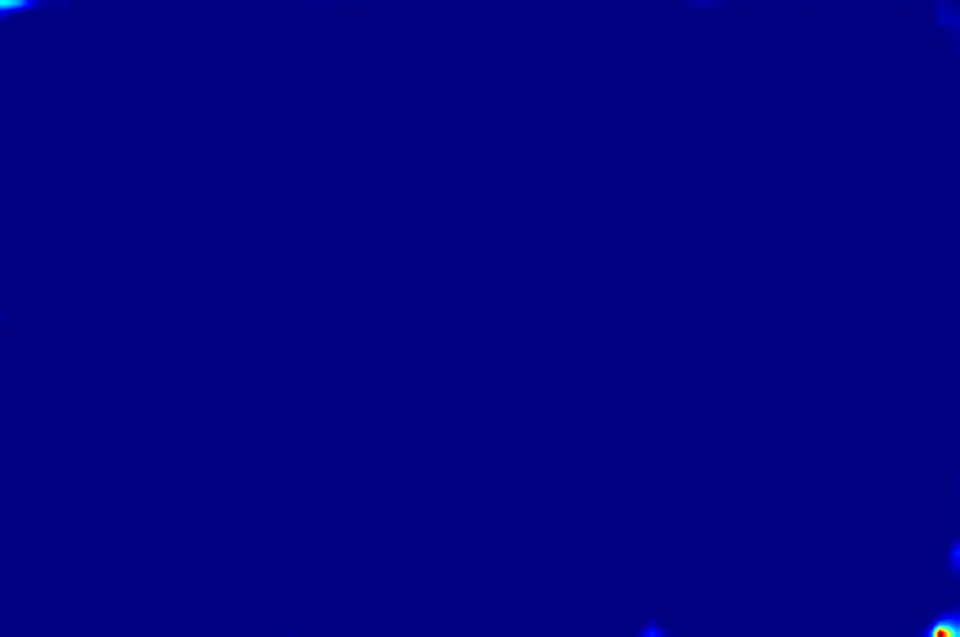} \\
			
			\footnotesize{A18} & \footnotesize{A19} & \footnotesize{A20} & \footnotesize{A21} & \footnotesize{A22} & \footnotesize{A23} \\
			
			\includegraphics[width=0.14\linewidth]{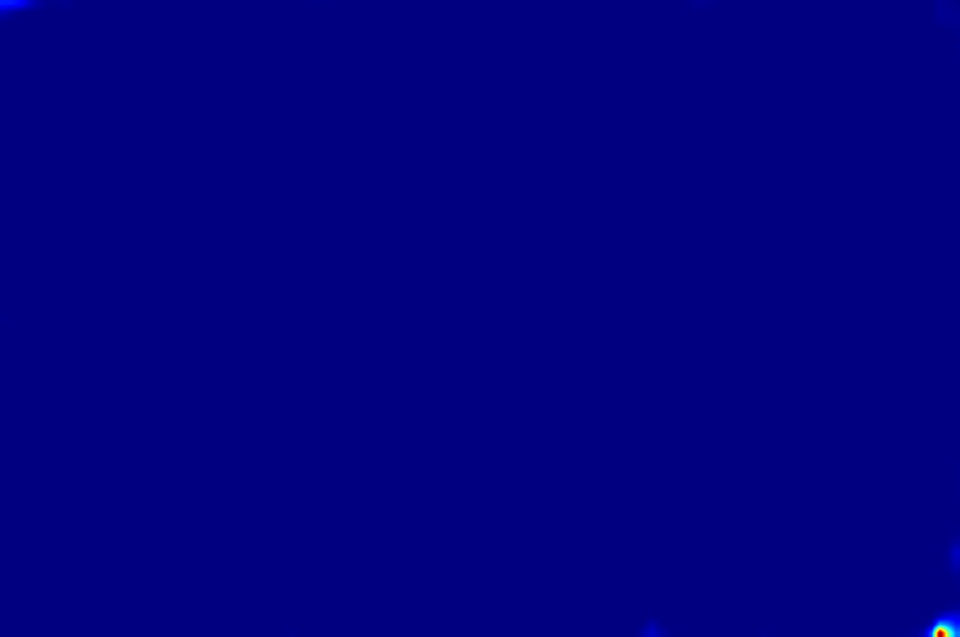}  &
			\includegraphics[width=0.14\linewidth]{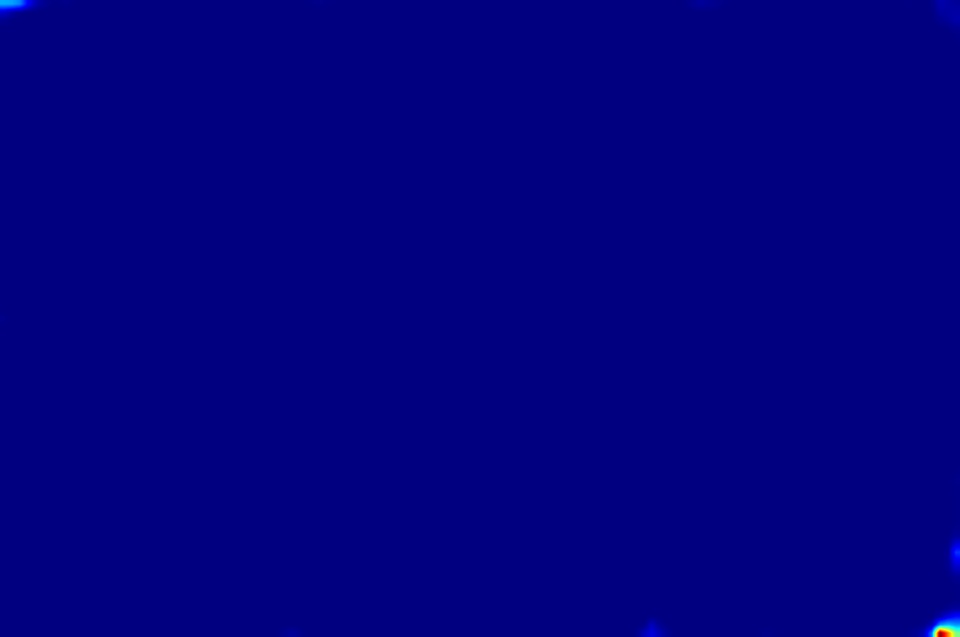}  &
			\includegraphics[width=0.14\linewidth]{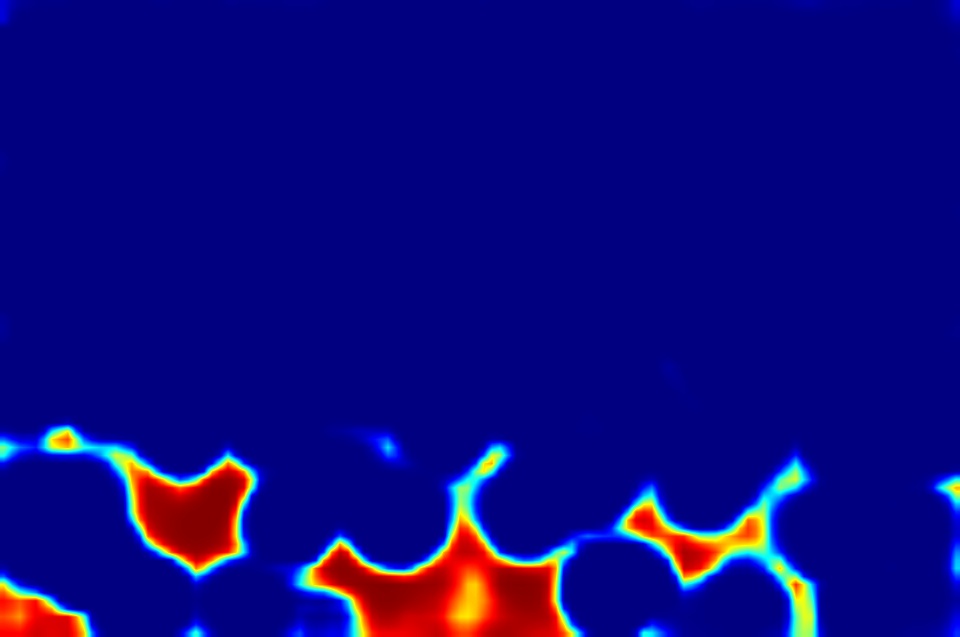} &
			\includegraphics[width=0.14\linewidth]{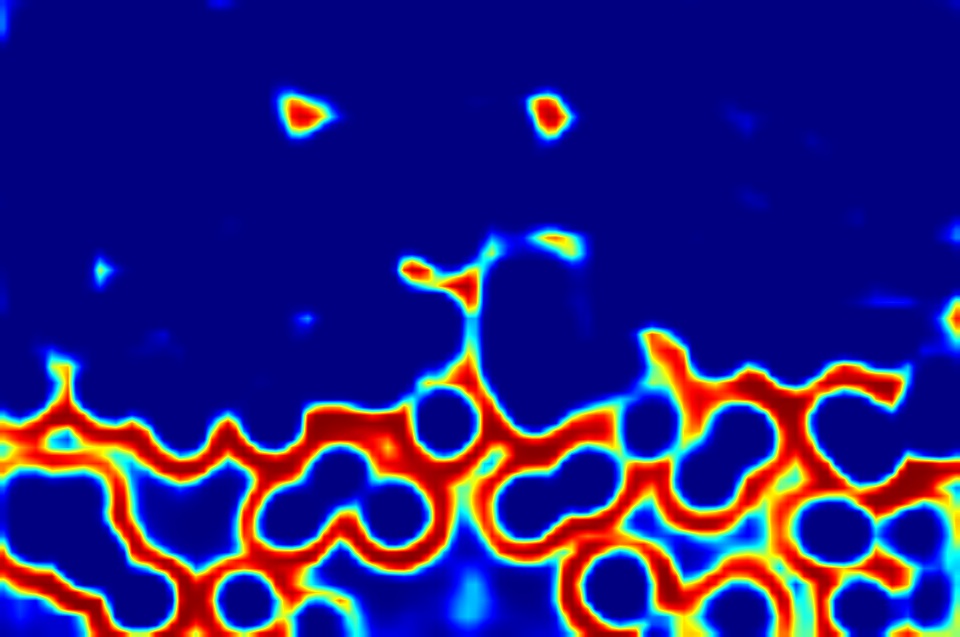}  &
			\includegraphics[width=0.14\linewidth]{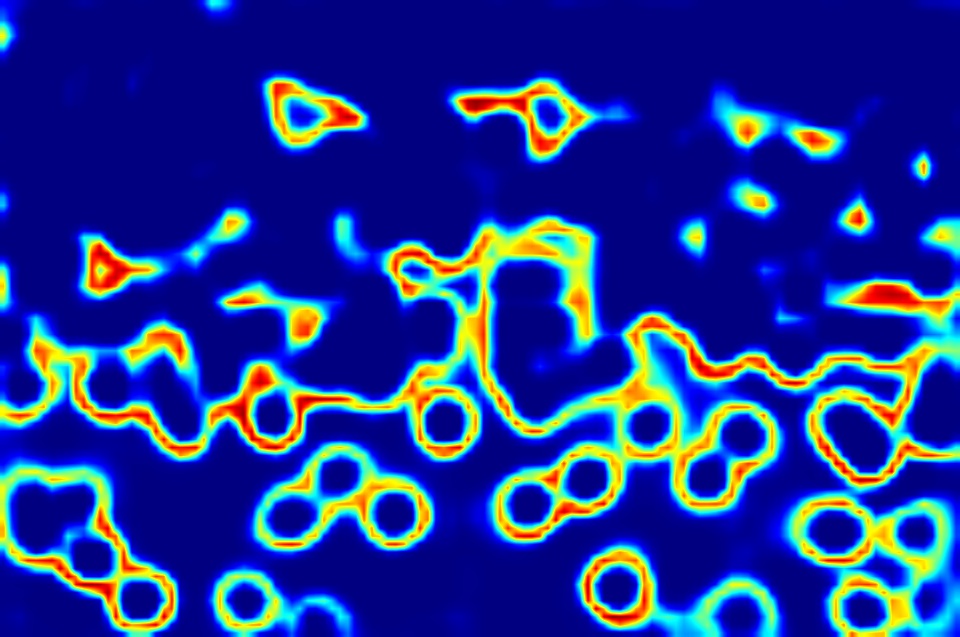} &
			\includegraphics[width=0.14\linewidth]{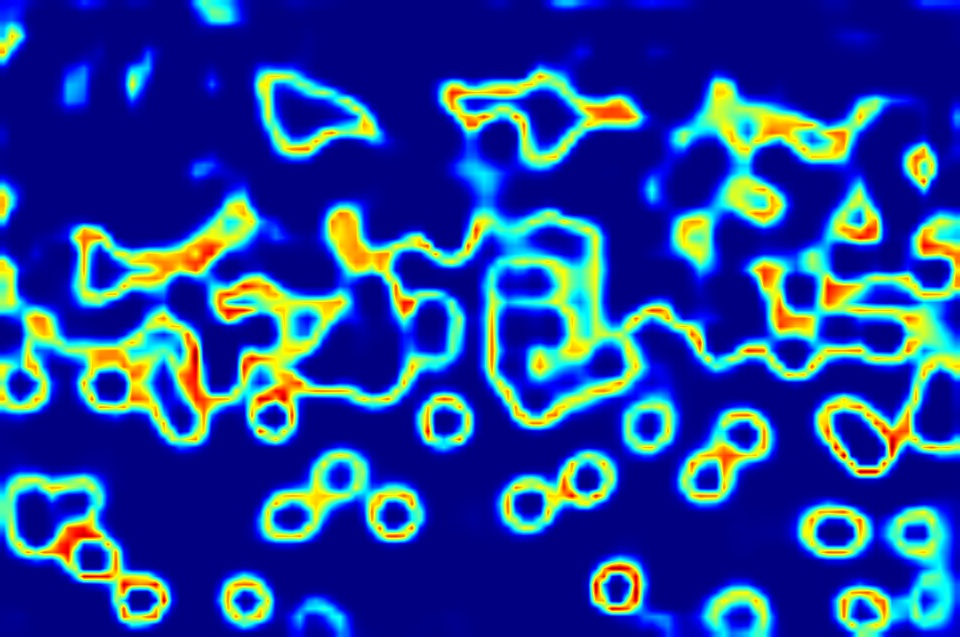} \\
			
			\footnotesize{A24} & \footnotesize{A25} & \footnotesize{B1} & \footnotesize{B2} & \footnotesize{B3} & \footnotesize{B4} \\
			
			\includegraphics[width=0.14\linewidth]{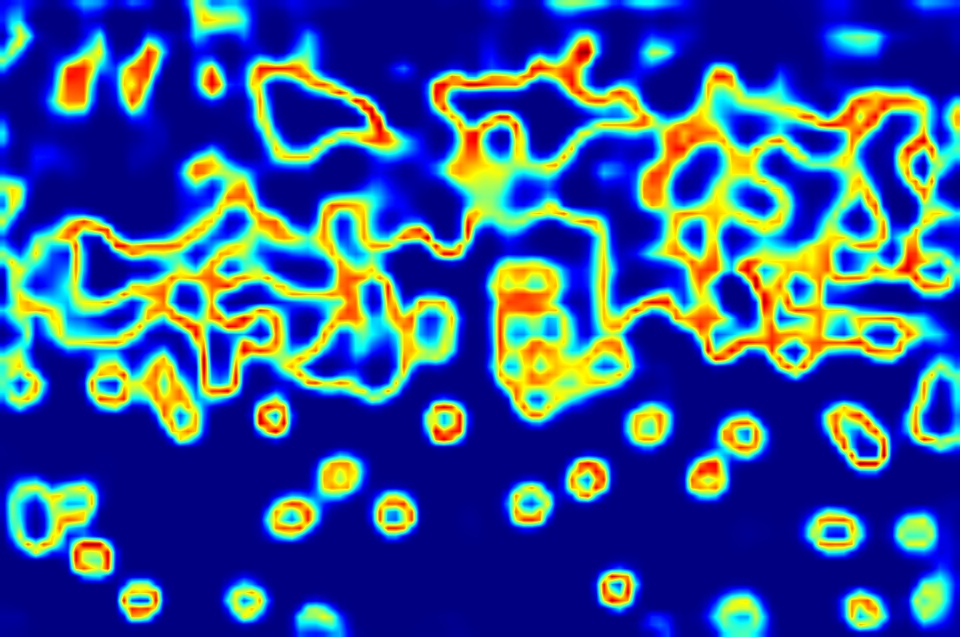}  &
			\includegraphics[width=0.14\linewidth]{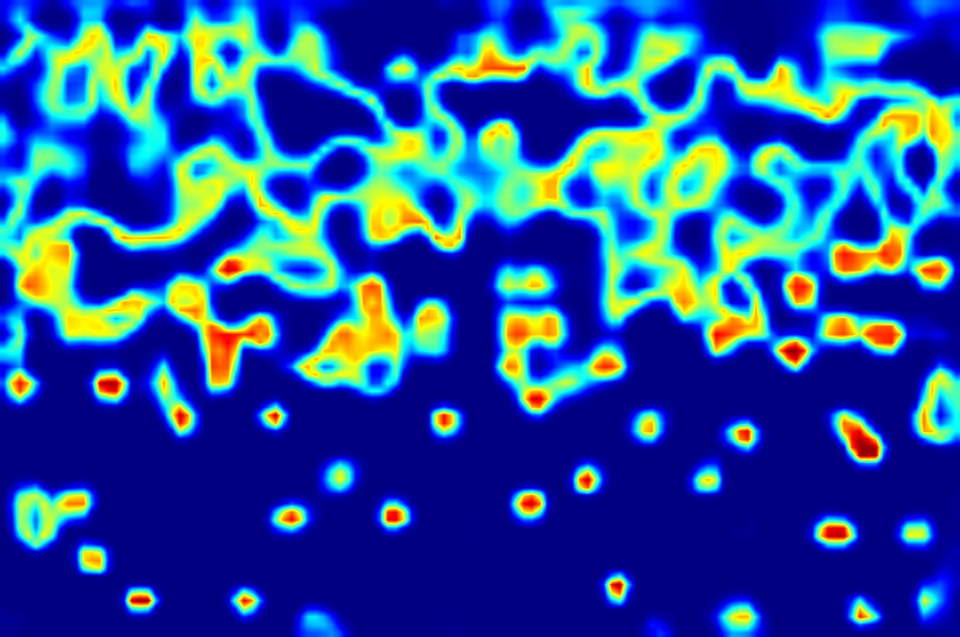}  &
			\includegraphics[width=0.14\linewidth]{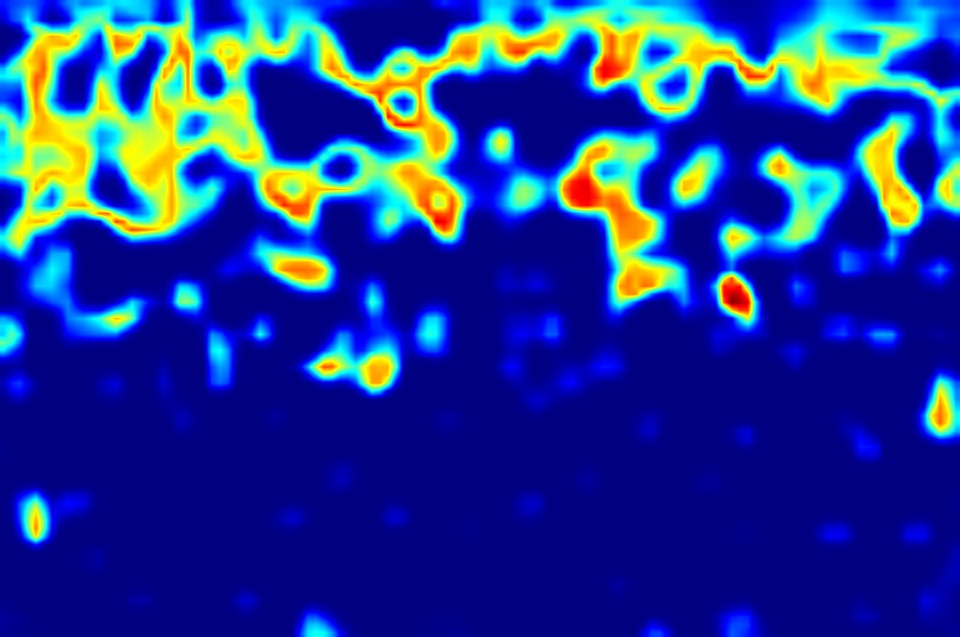} &
			\includegraphics[width=0.14\linewidth]{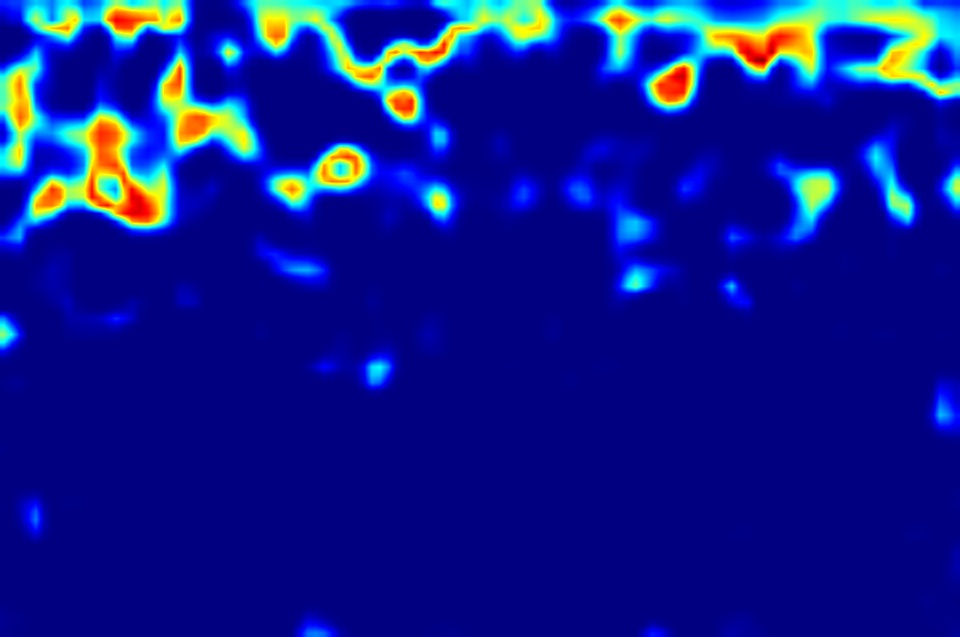}  &
			\includegraphics[width=0.14\linewidth]{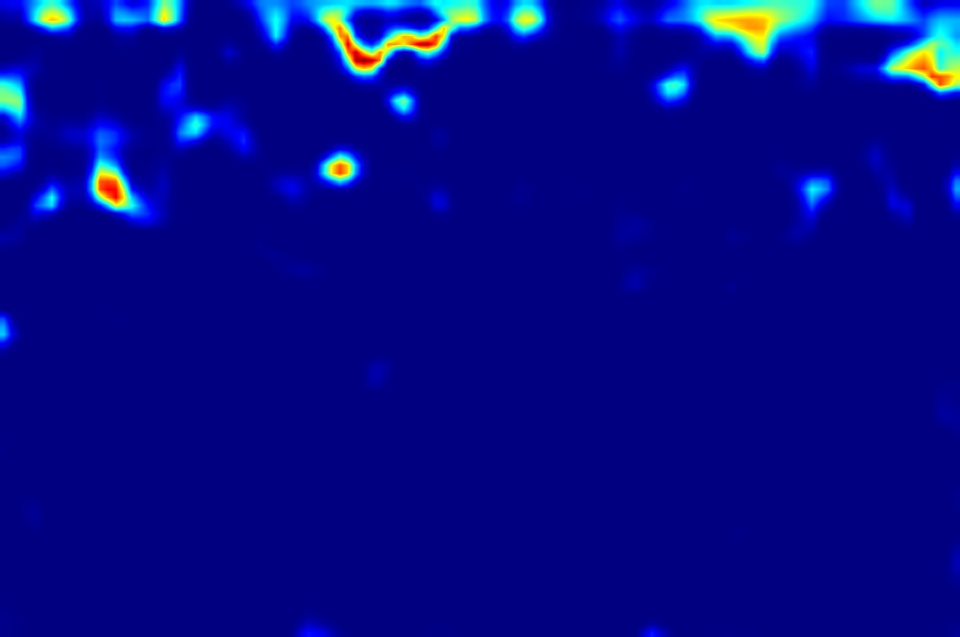} &
			\includegraphics[width=0.14\linewidth]{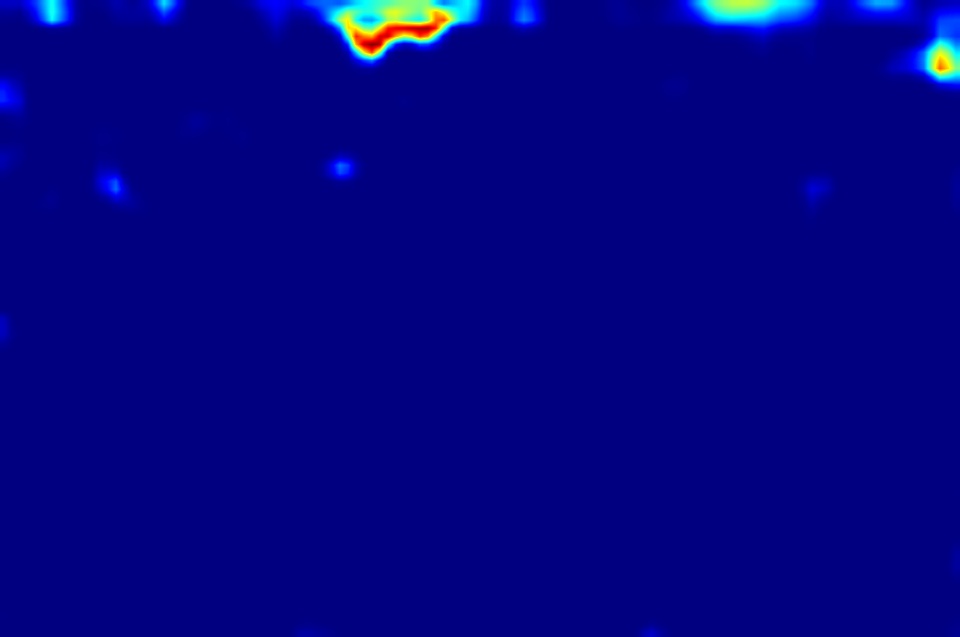} \\
			
			\footnotesize{B5} & \footnotesize{B6} & \footnotesize{B7} & \footnotesize{B8} & \footnotesize{B9} & \footnotesize{B10} \\
			
			\includegraphics[width=0.14\linewidth]{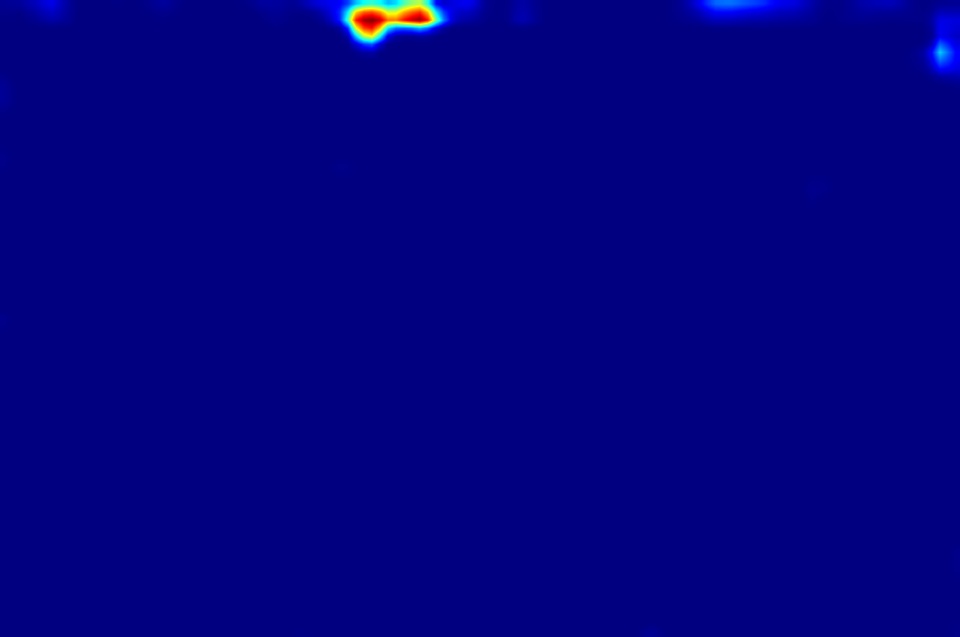}  &
			\includegraphics[width=0.14\linewidth]{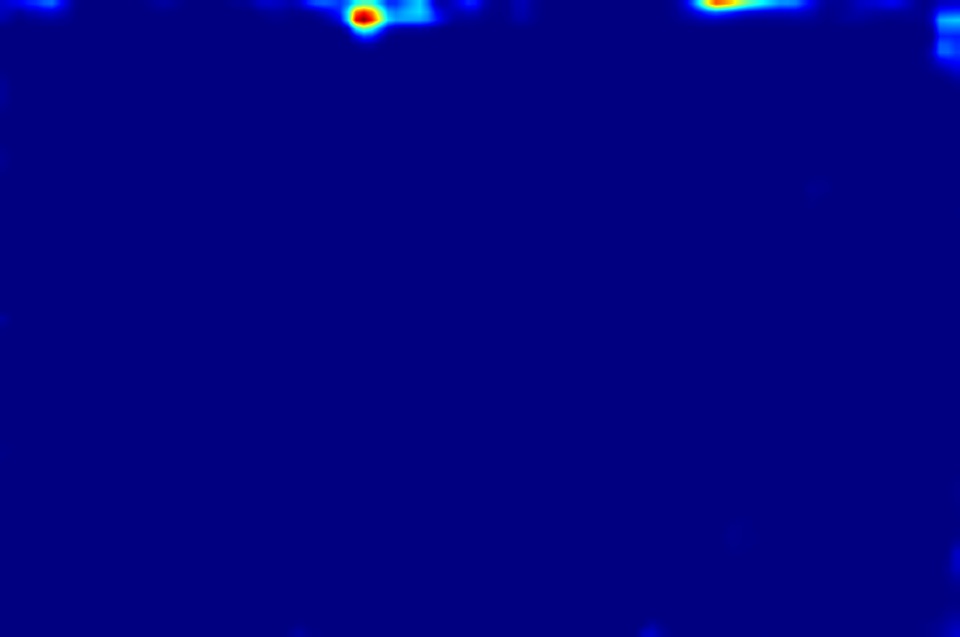}  &
			\includegraphics[width=0.14\linewidth]{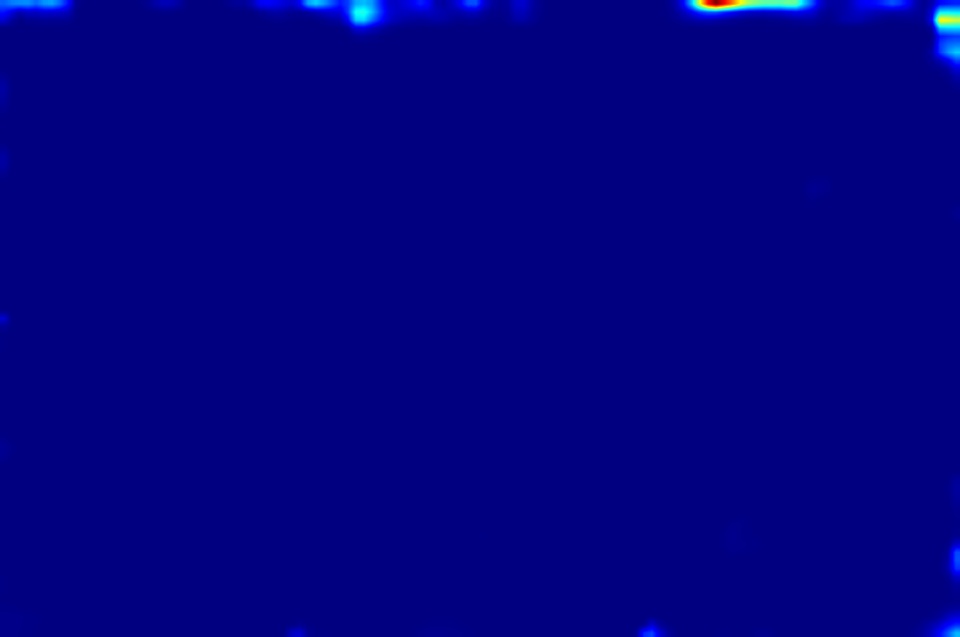} &
			\includegraphics[width=0.14\linewidth]{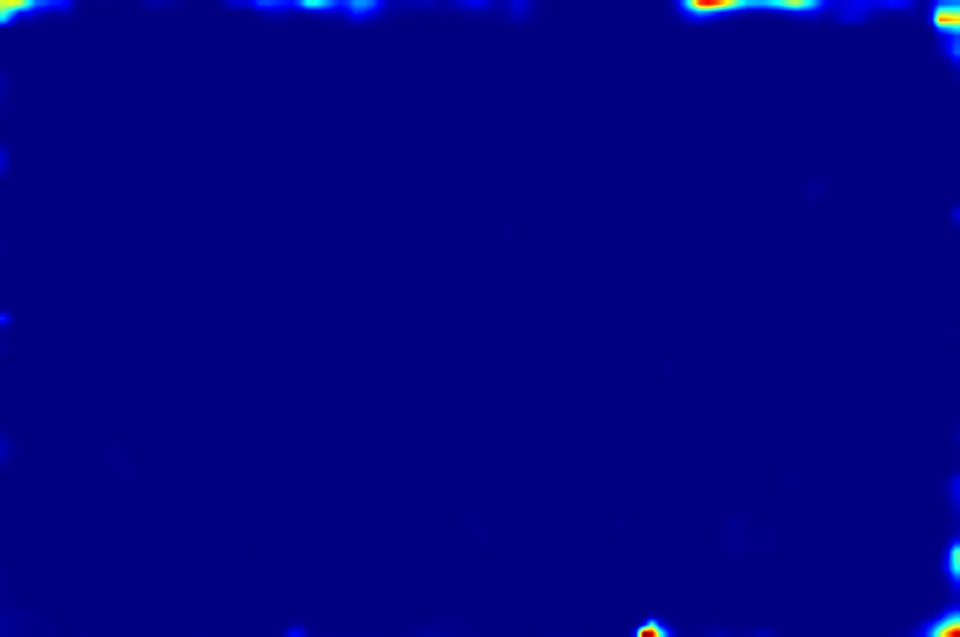}  &
			\includegraphics[width=0.14\linewidth]{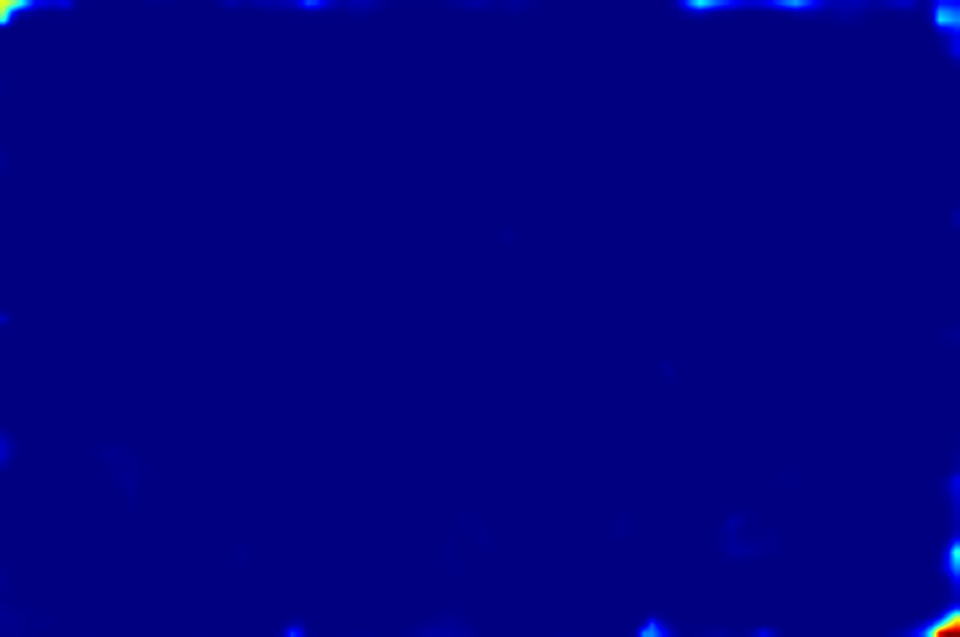} &
			\includegraphics[width=0.14\linewidth]{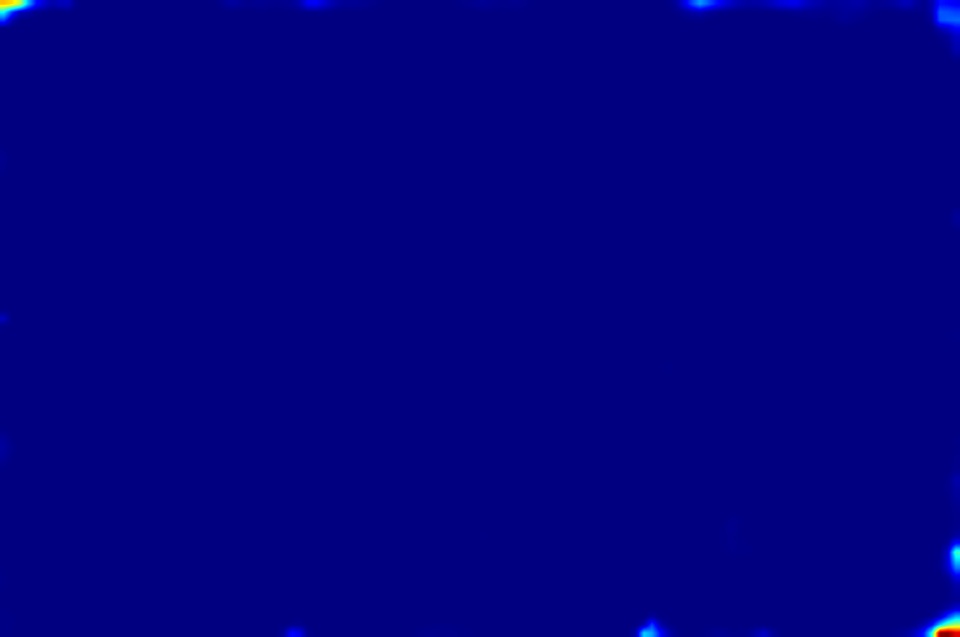} \\
			
			\footnotesize{B11} & \footnotesize{B12} & \footnotesize{B13} & \footnotesize{B14} & \footnotesize{B15} & \footnotesize{B16} \\
			
			\includegraphics[width=0.14\linewidth]{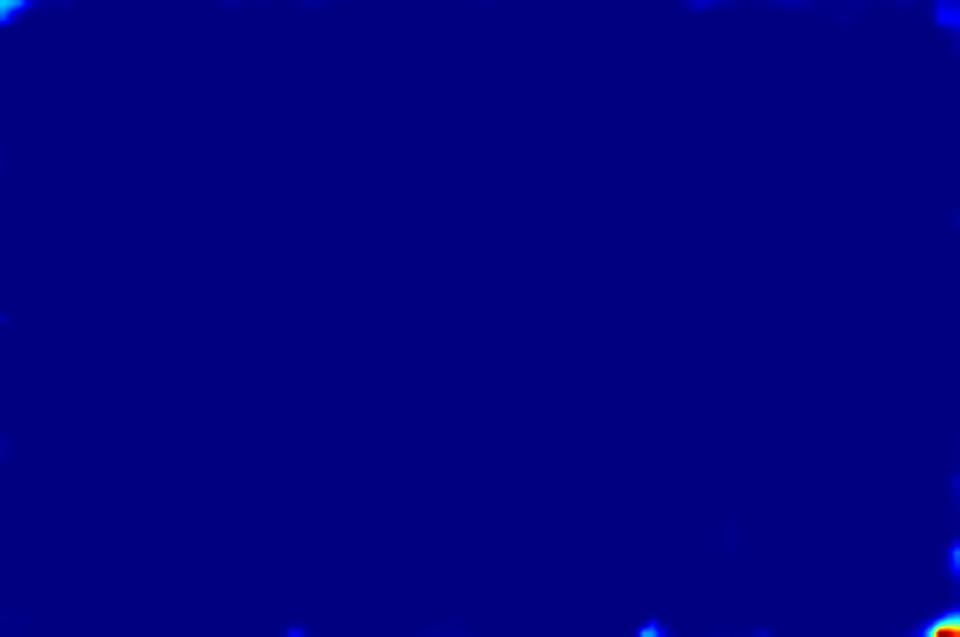}  &
			\includegraphics[width=0.14\linewidth]{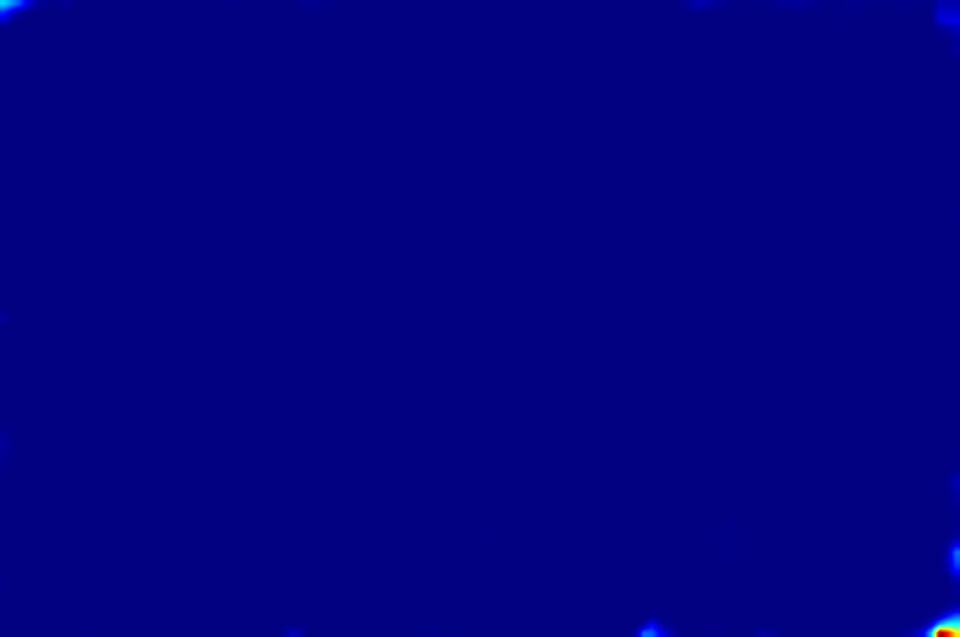}  &
			\includegraphics[width=0.14\linewidth]{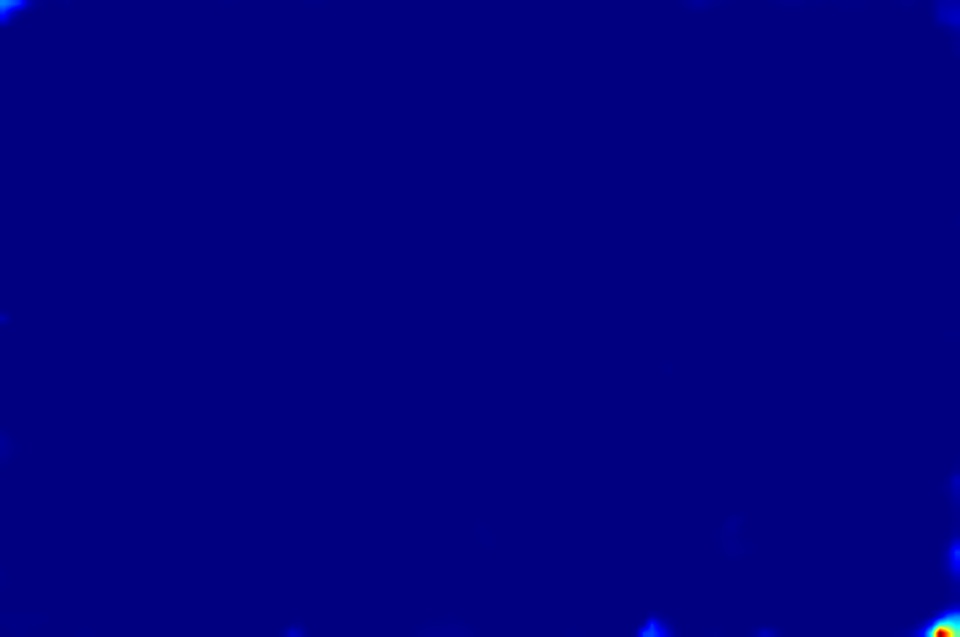} &
			\includegraphics[width=0.14\linewidth]{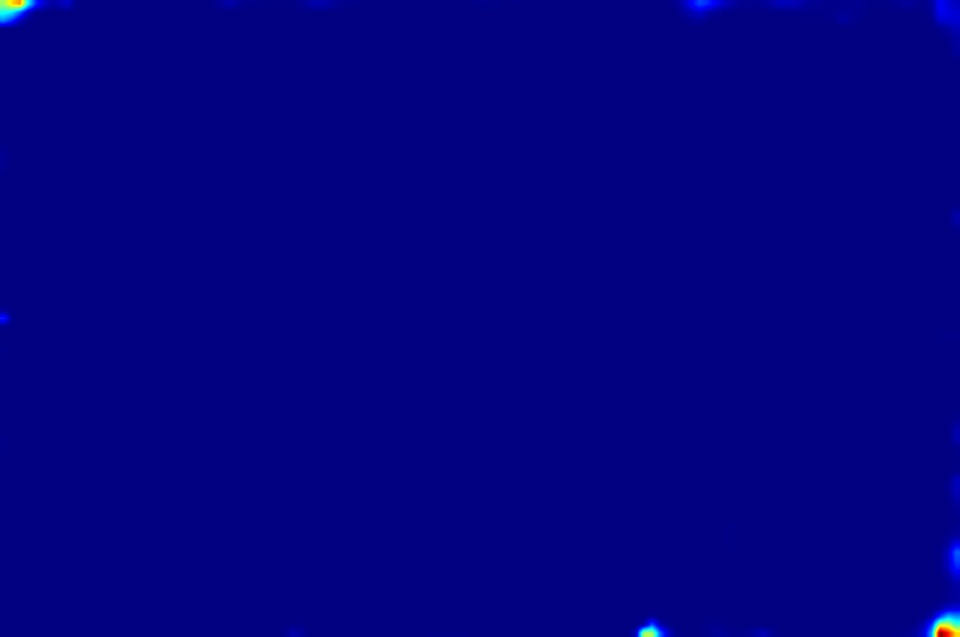}  &
			\includegraphics[width=0.14\linewidth]{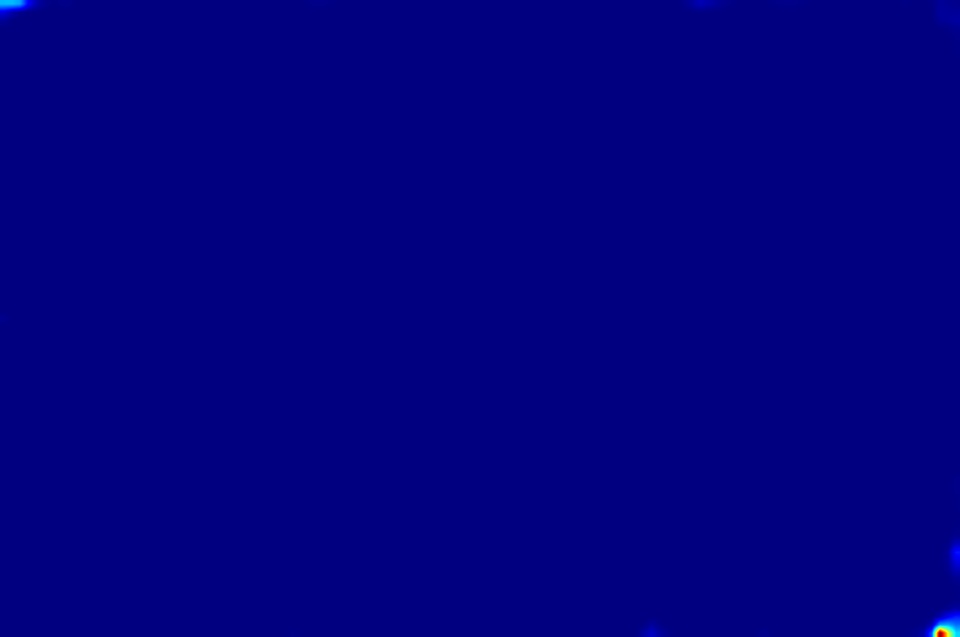} &
			\includegraphics[width=0.14\linewidth]{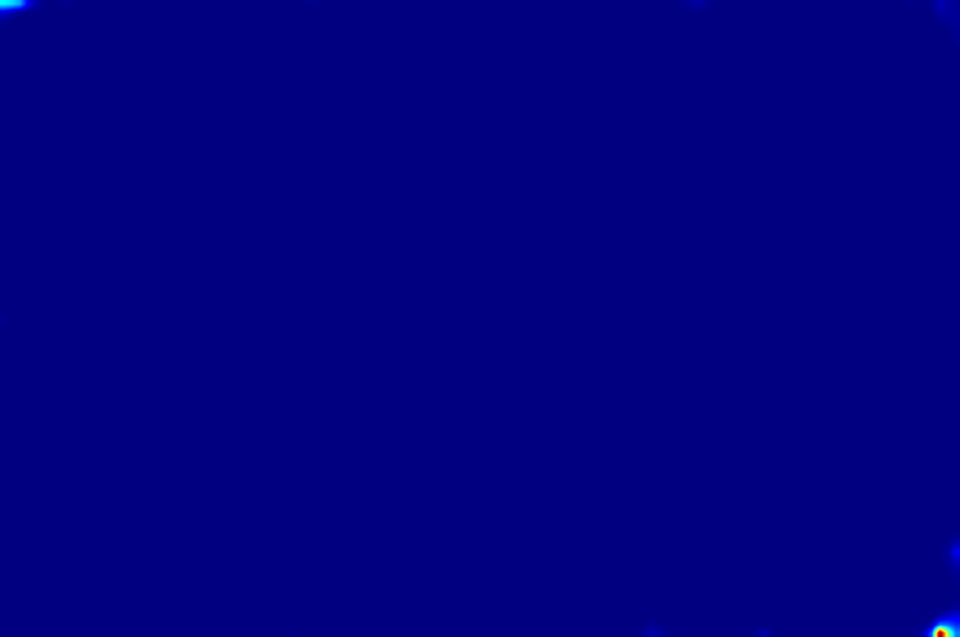} \\
			
			\footnotesize{B17} & \footnotesize{B18} & \footnotesize{B19} & \footnotesize{B20} & \footnotesize{B21} & \footnotesize{B22} \\
			
			\includegraphics[width=0.14\linewidth]{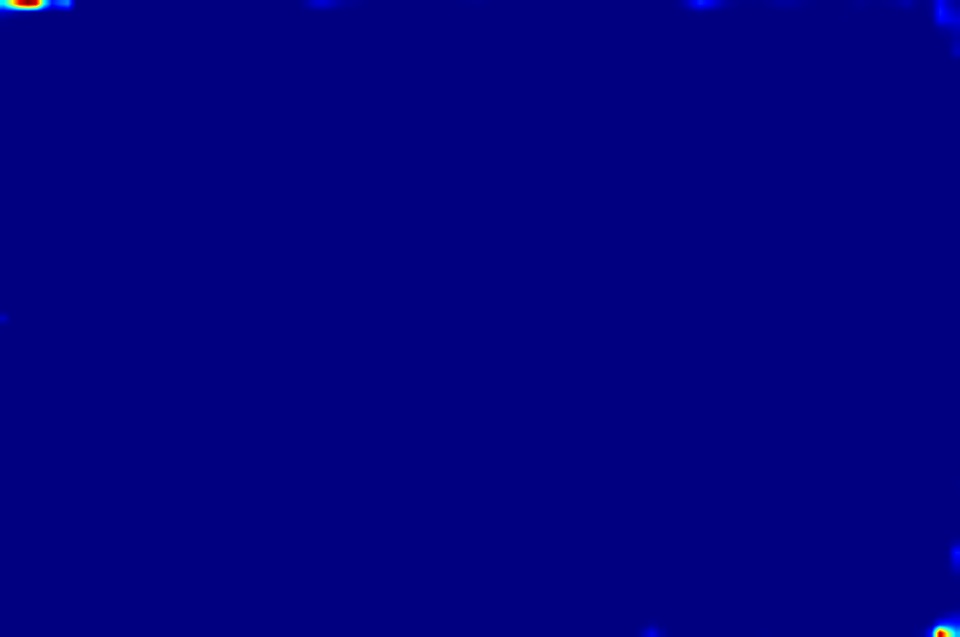}  &
			\includegraphics[width=0.14\linewidth]{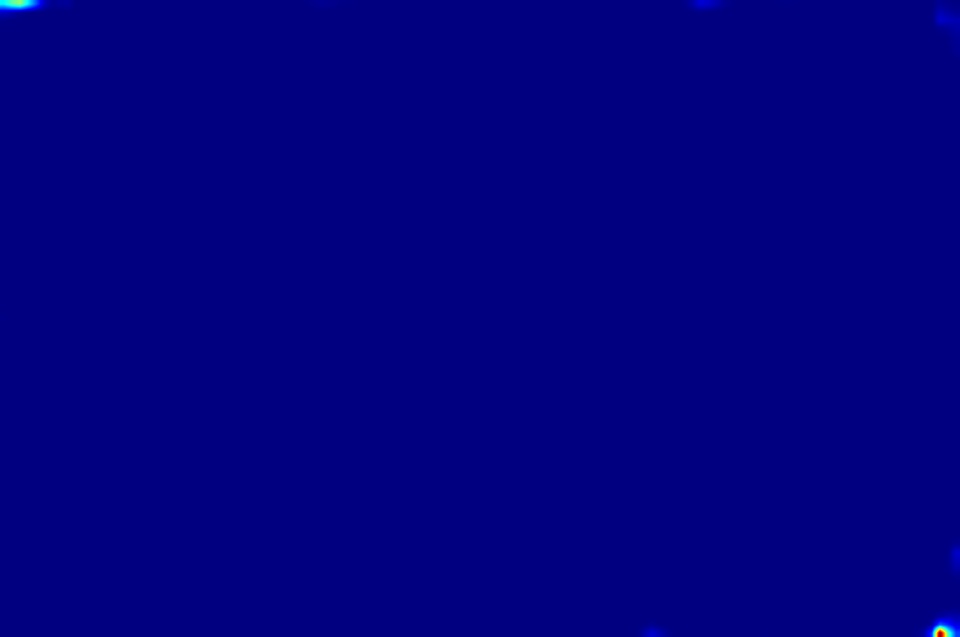}  &
			\includegraphics[width=0.14\linewidth]{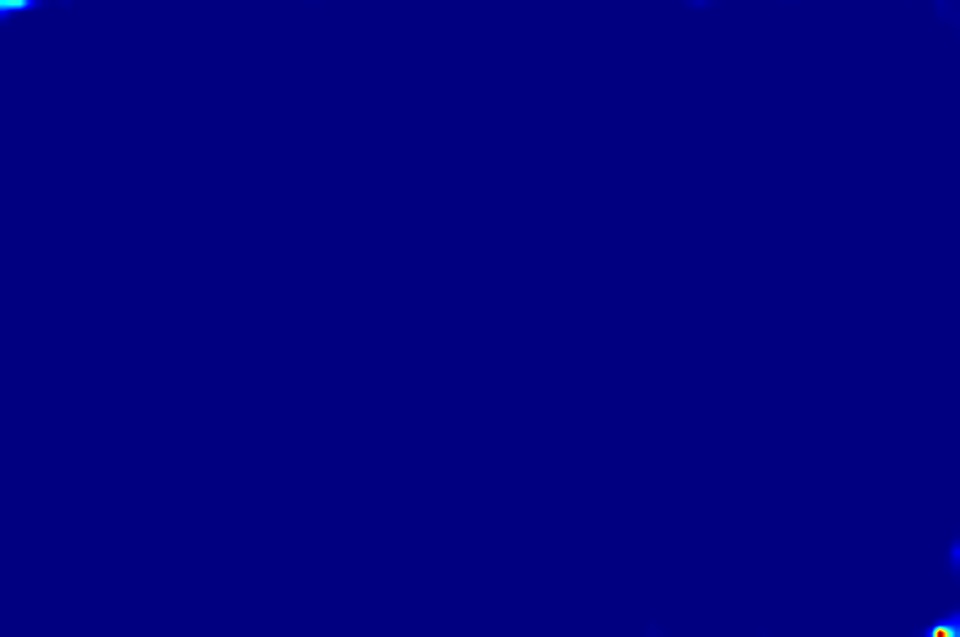} \\
			
			\footnotesize{B23} & \footnotesize{B24} & \footnotesize{B25} \\

		\end{tabular}
		\caption{Attention maps of each density token in dual branch for a labeled training image when training with a labeled ratio of $5\%$ on UCF-QNRF. A and B stand for different branches and the numbers represent the different tokens. Tokens with higher numbers specify the density interval with higher density.}
		\label{fig:viz}
	\end{center}

\end{figure*}}

\renewcommand{\tabcolsep}{5 pt}{
\begin{figure*}[t!]
	\begin{center}
		\begin{tabular}{cccccc}
			
			\includegraphics[width=0.14\linewidth]{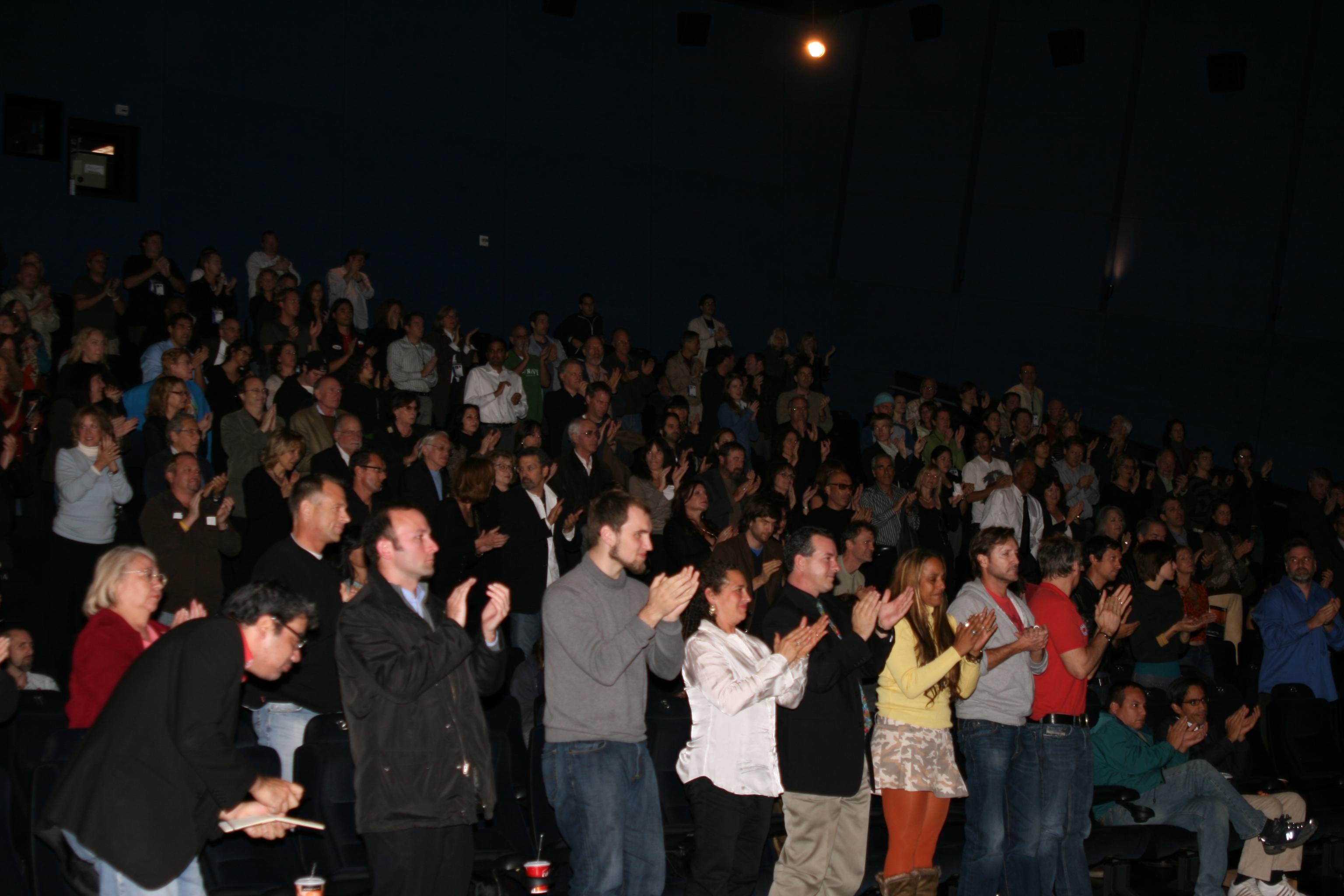}  &
			\includegraphics[width=0.14\linewidth]{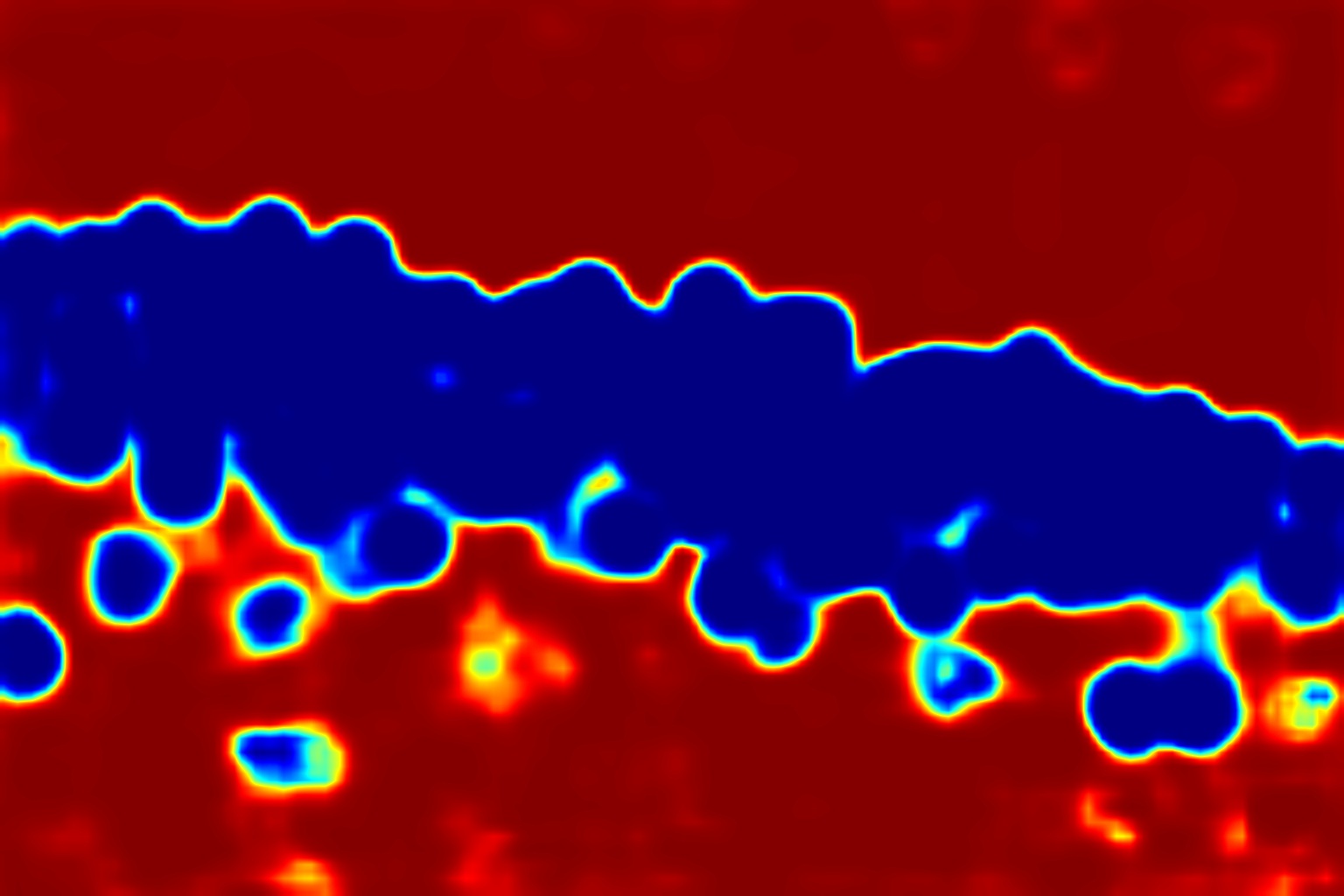}  &
			\includegraphics[width=0.14\linewidth]{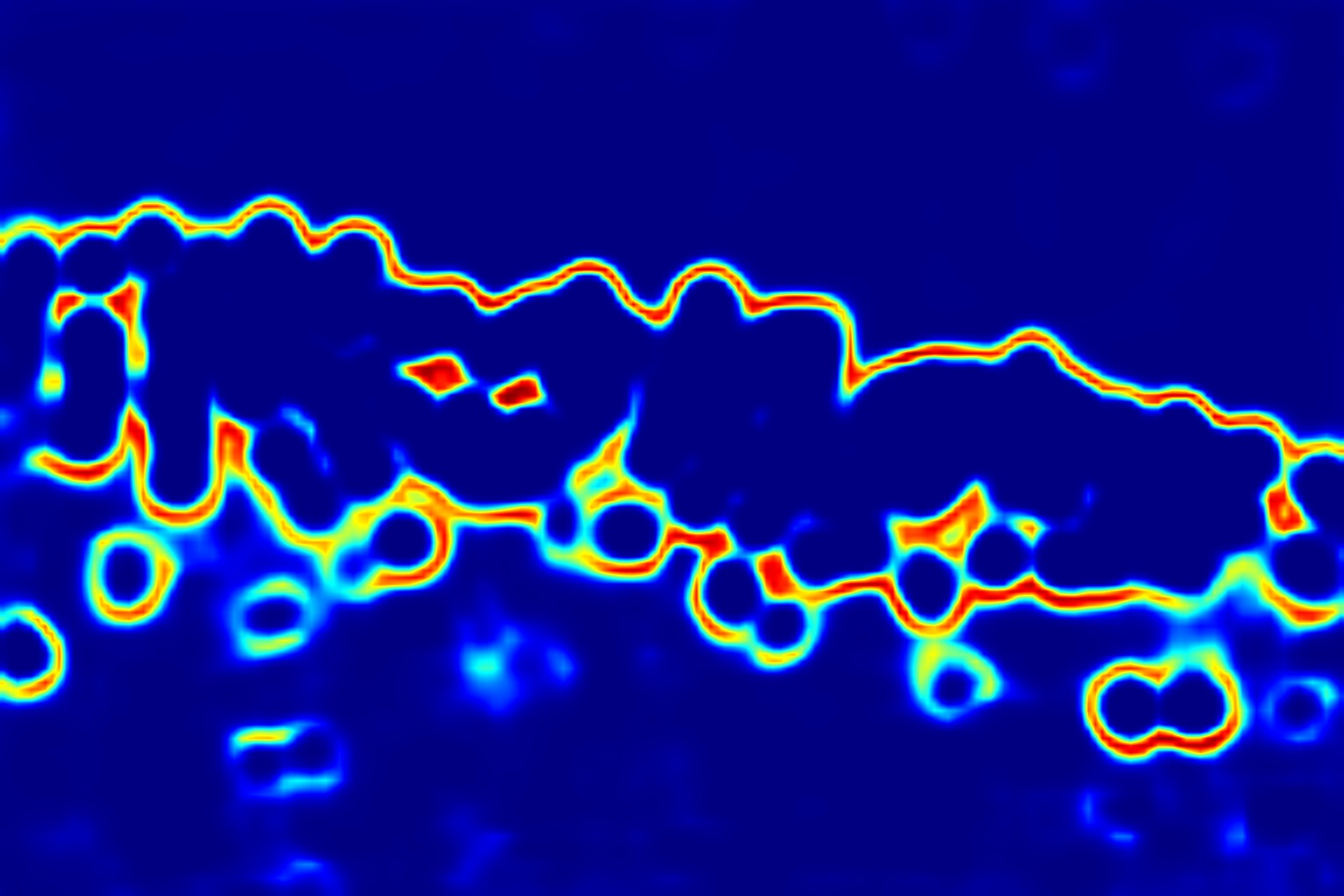} &
			\includegraphics[width=0.14\linewidth]{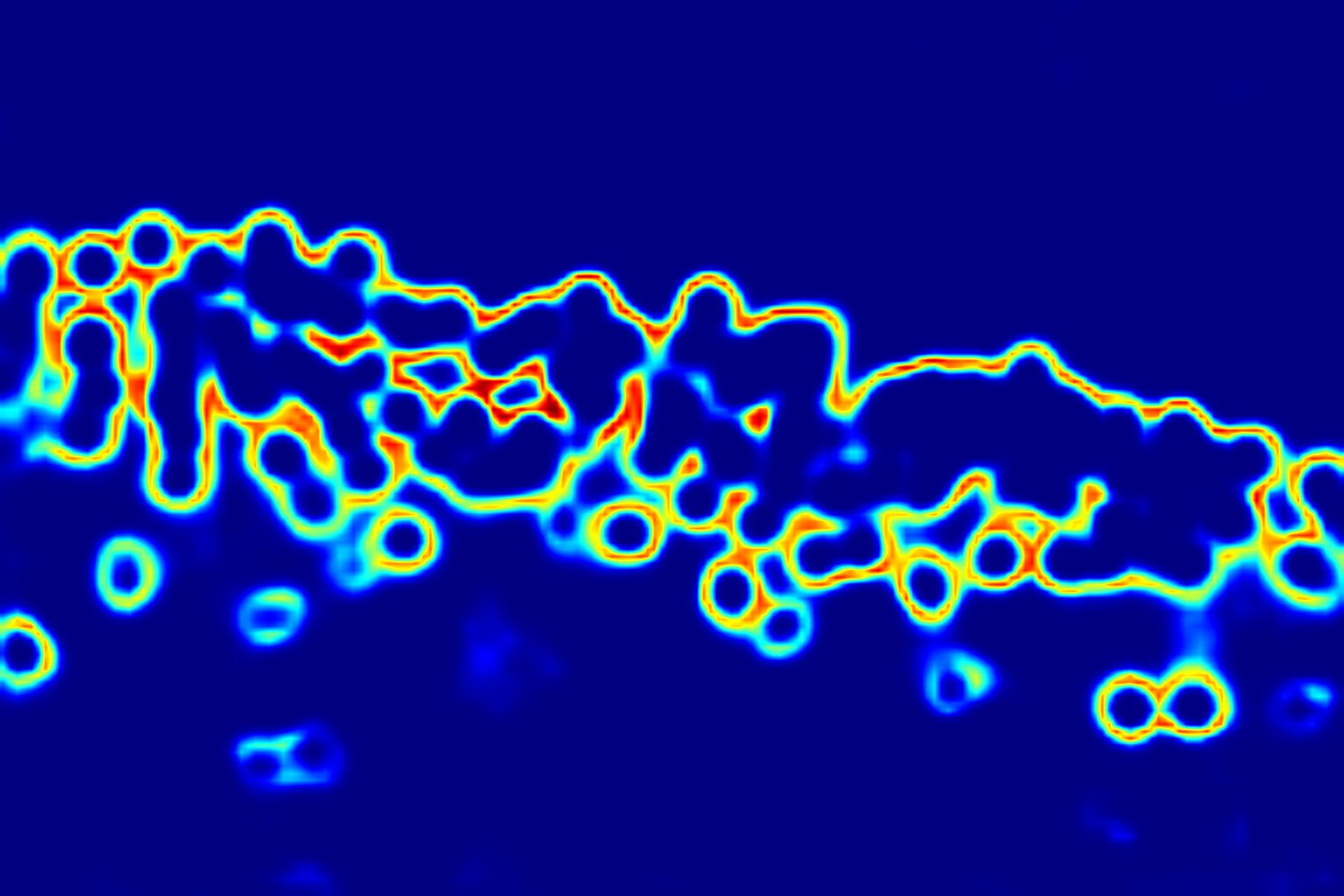}  &
			\includegraphics[width=0.14\linewidth]{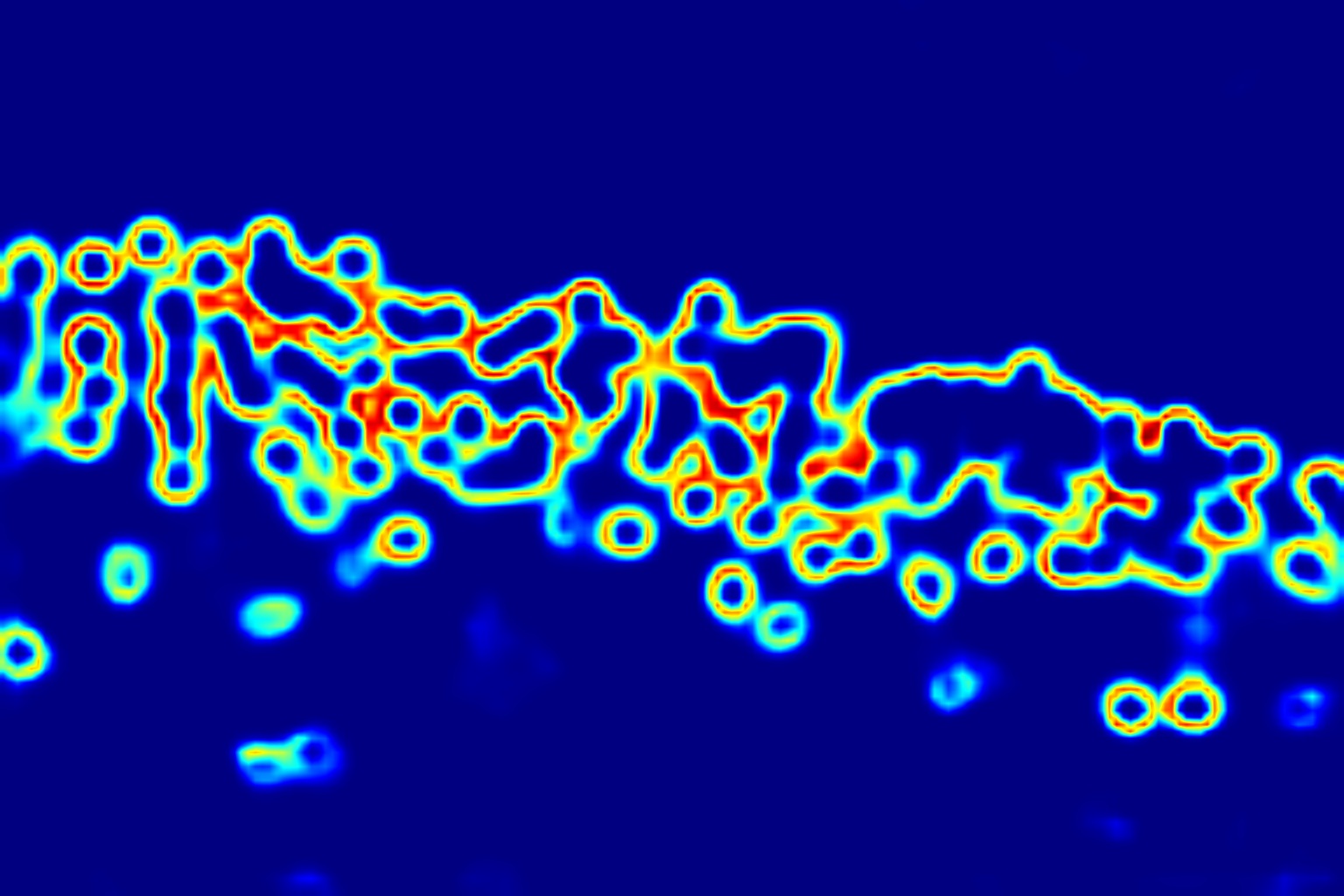} &
			\includegraphics[width=0.14\linewidth]{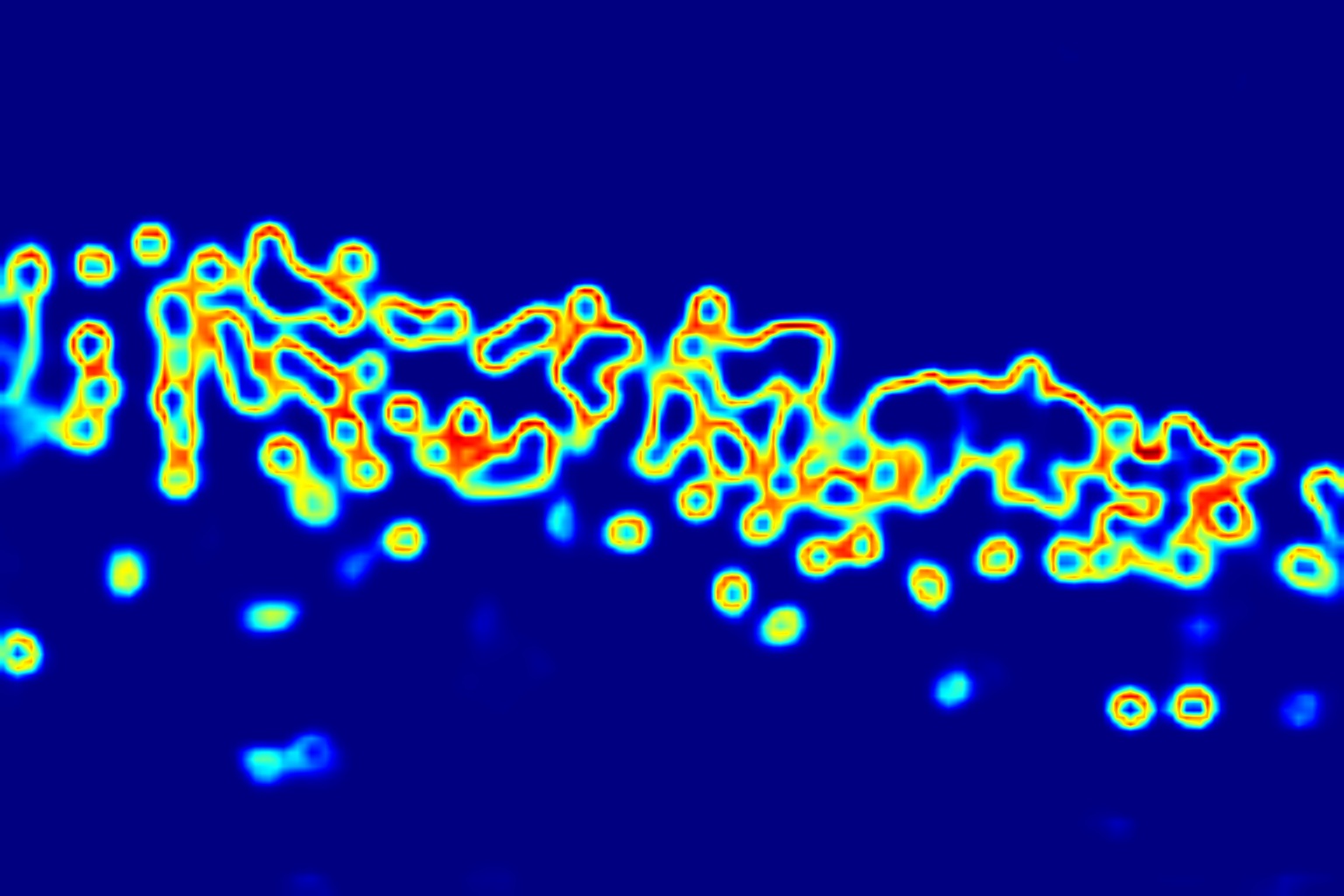} \\
			
			\footnotesize{Labeled Image} & \footnotesize{A1} & \footnotesize{A2} & \footnotesize{A3} & \footnotesize{A4} & \footnotesize{A5} \\
			
			\includegraphics[width=0.14\linewidth]{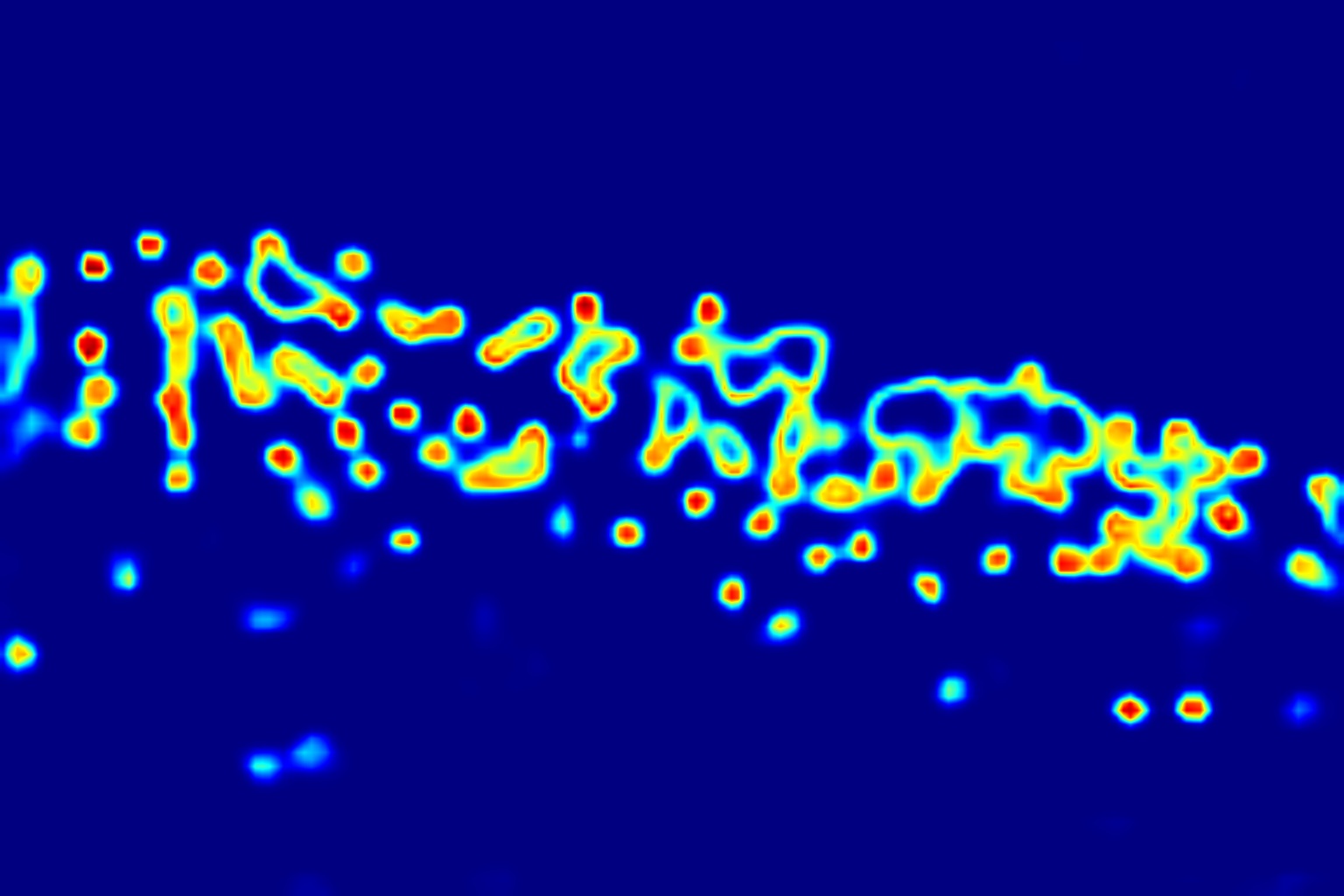}  &
			\includegraphics[width=0.14\linewidth]{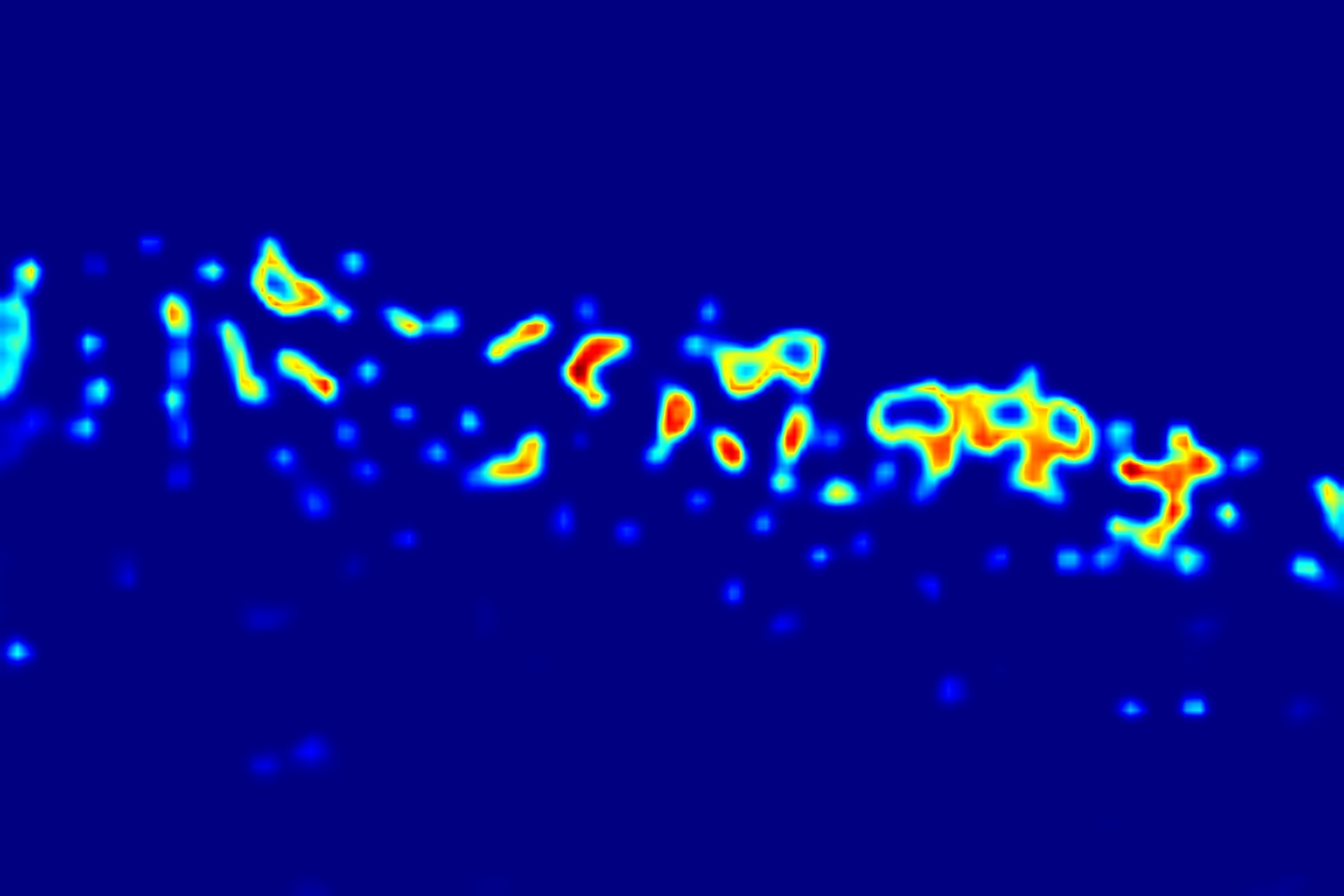}  &
			\includegraphics[width=0.14\linewidth]{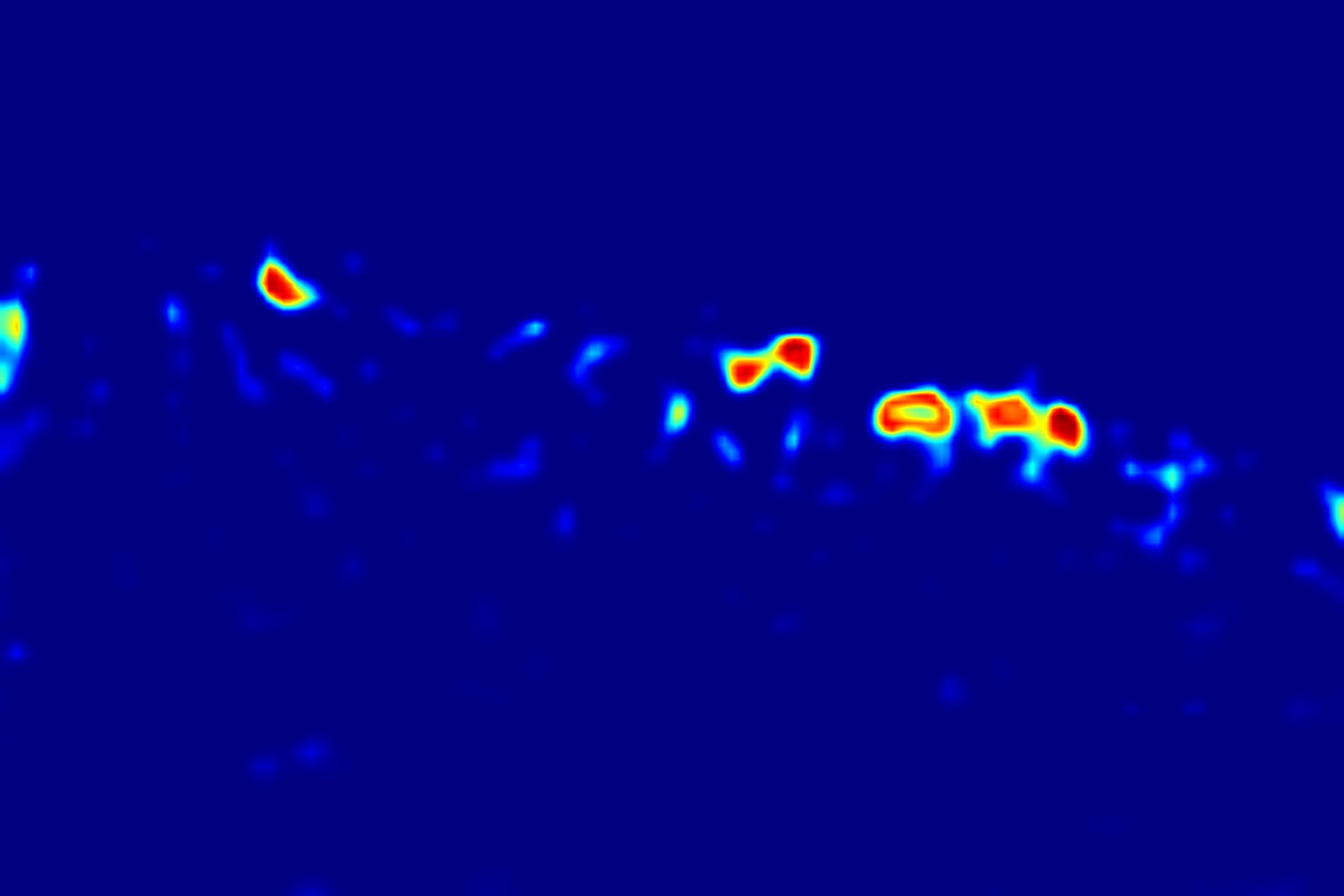} &
			\includegraphics[width=0.14\linewidth]{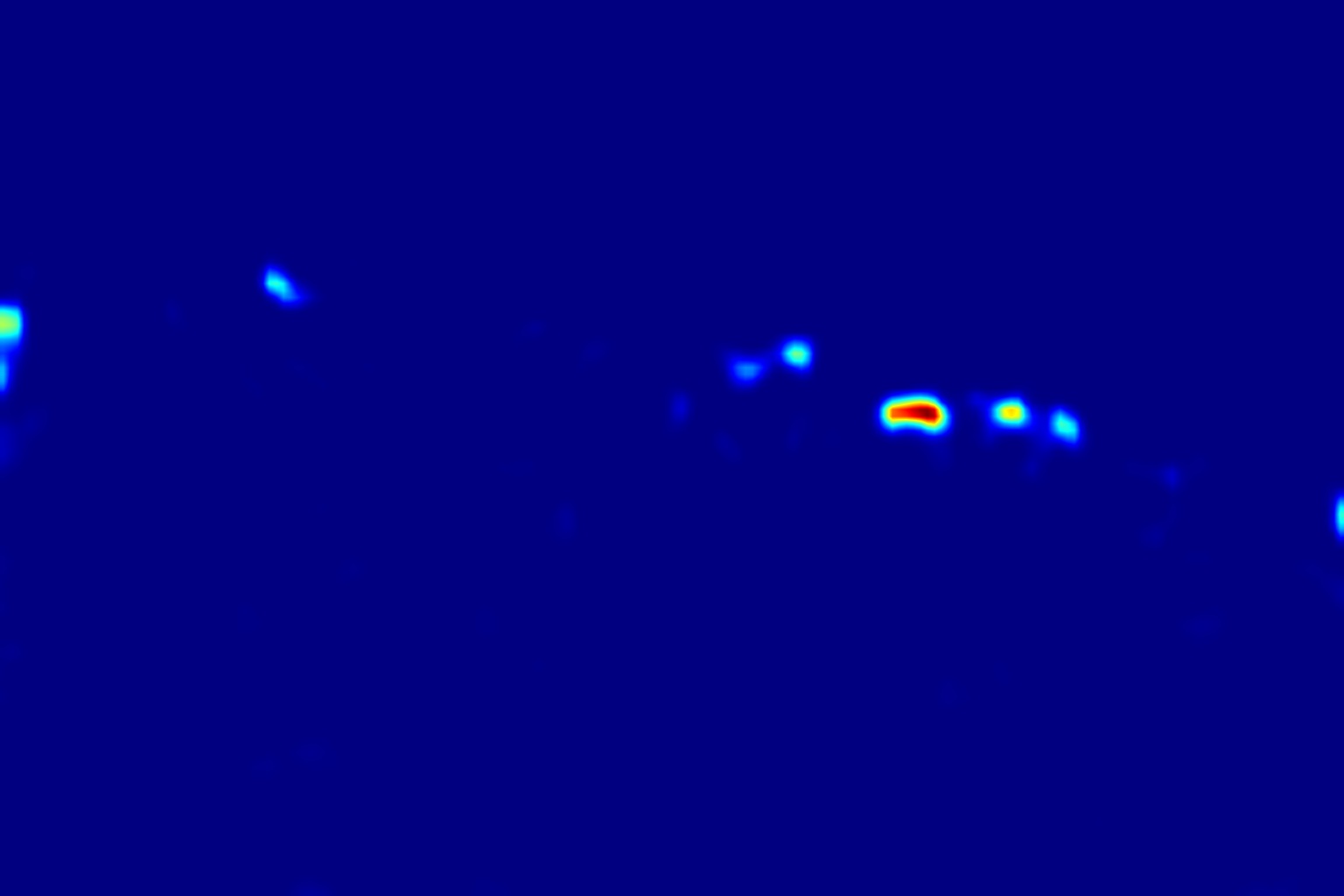}  &
			\includegraphics[width=0.14\linewidth]{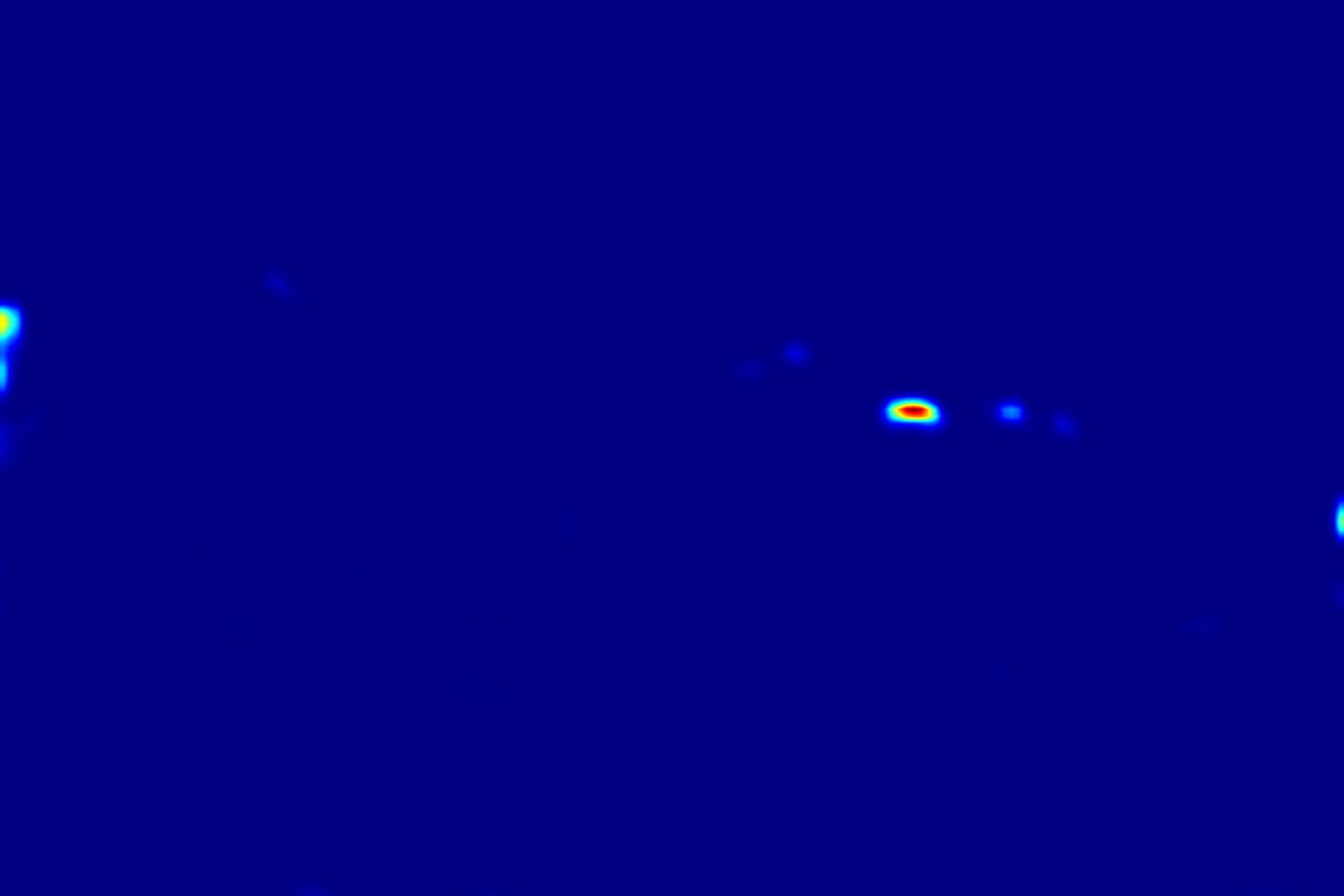} &
			\includegraphics[width=0.14\linewidth]{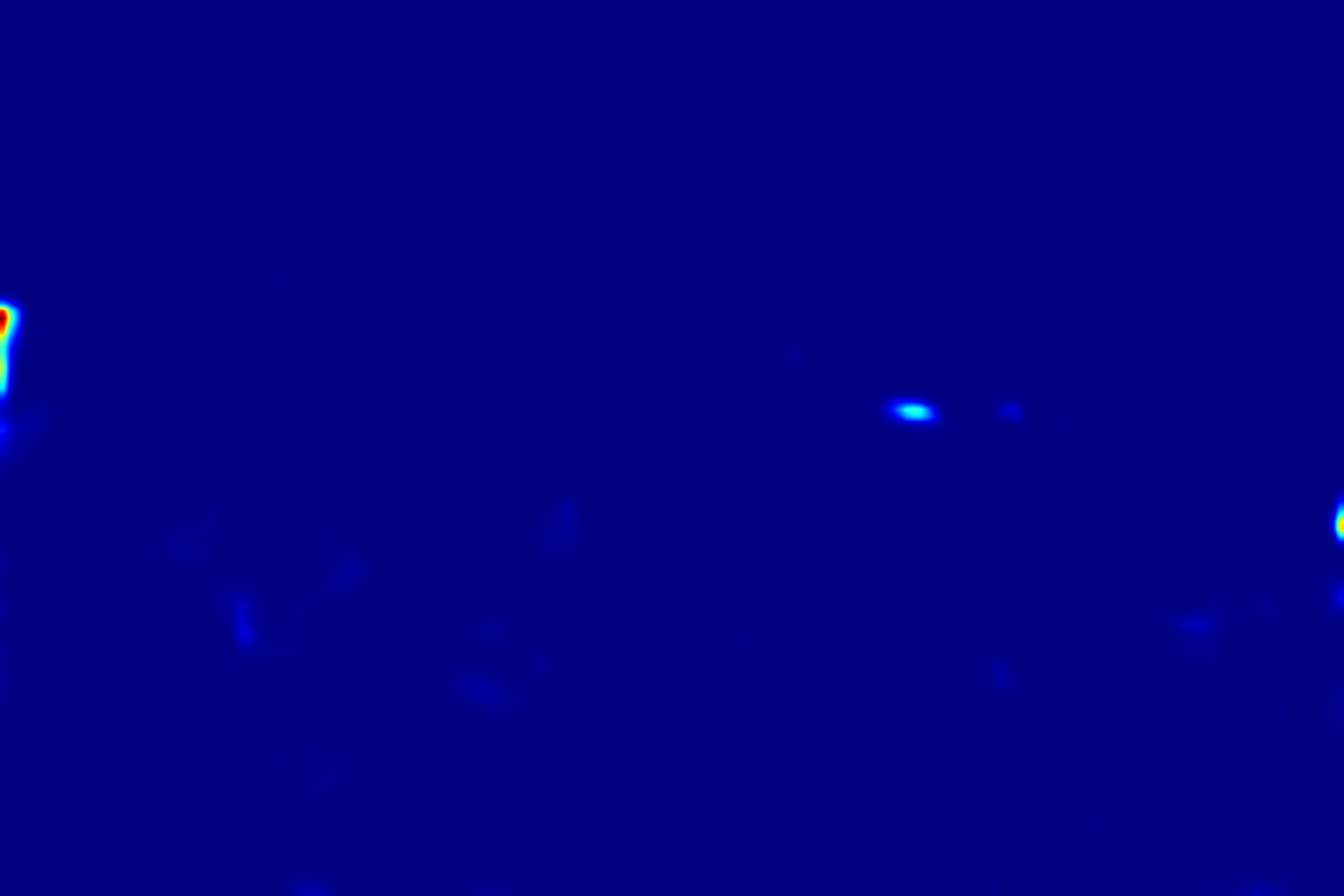} \\
			
			\footnotesize{A6} & \footnotesize{A7} & \footnotesize{A8} & \footnotesize{A9} & \footnotesize{A10} & \footnotesize{A11} \\
			
			\includegraphics[width=0.14\linewidth]{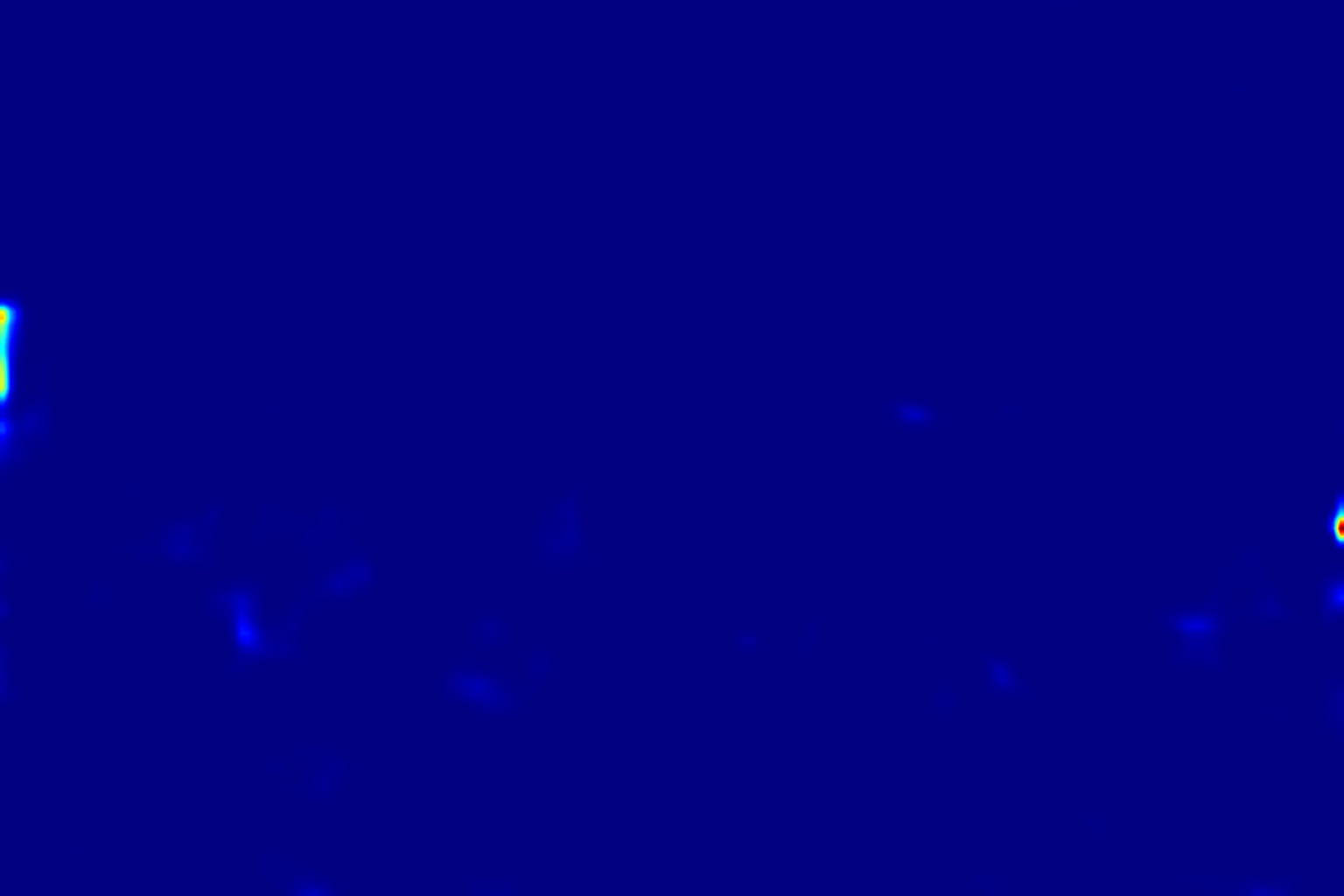}  &
			\includegraphics[width=0.14\linewidth]{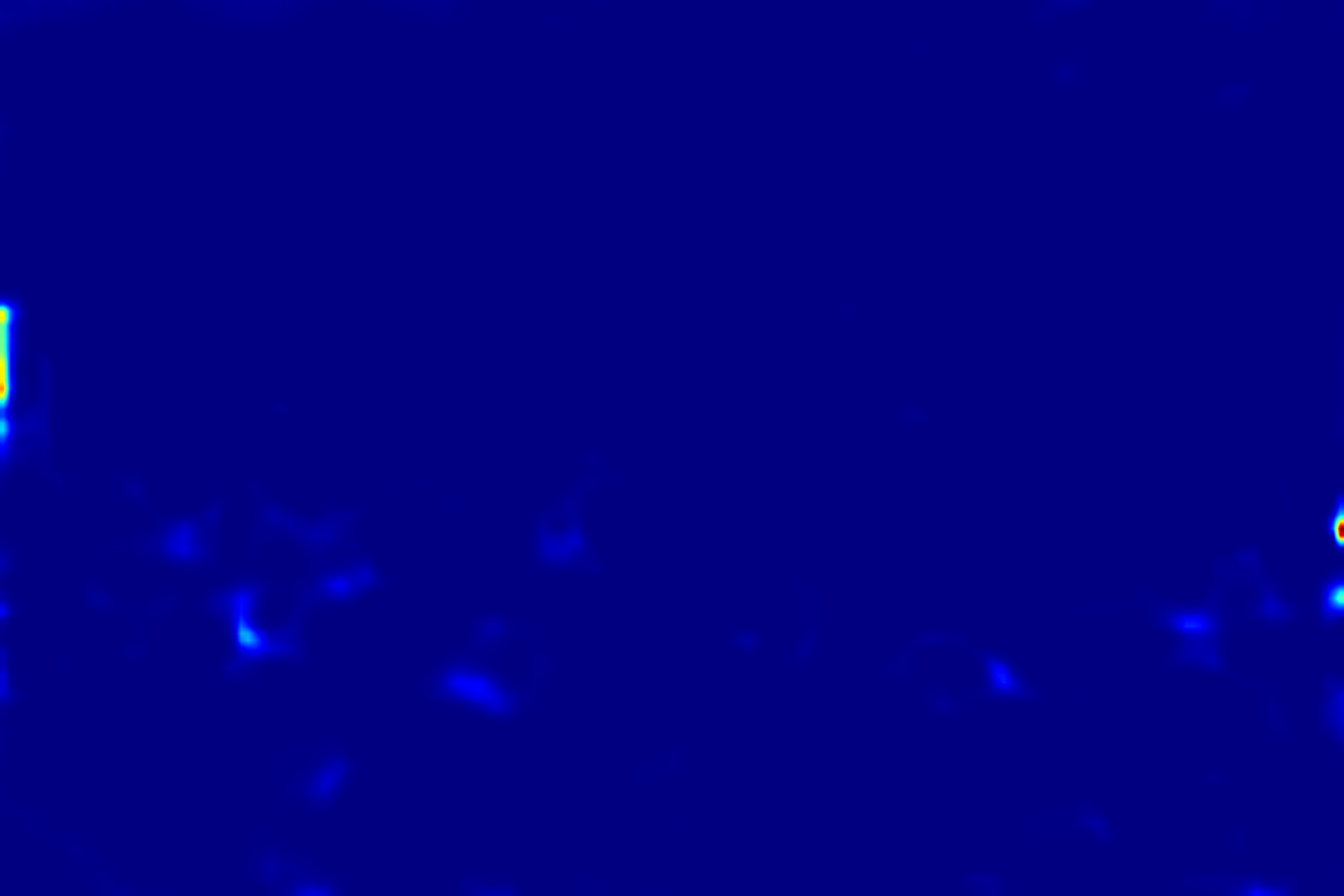}  &
			\includegraphics[width=0.14\linewidth]{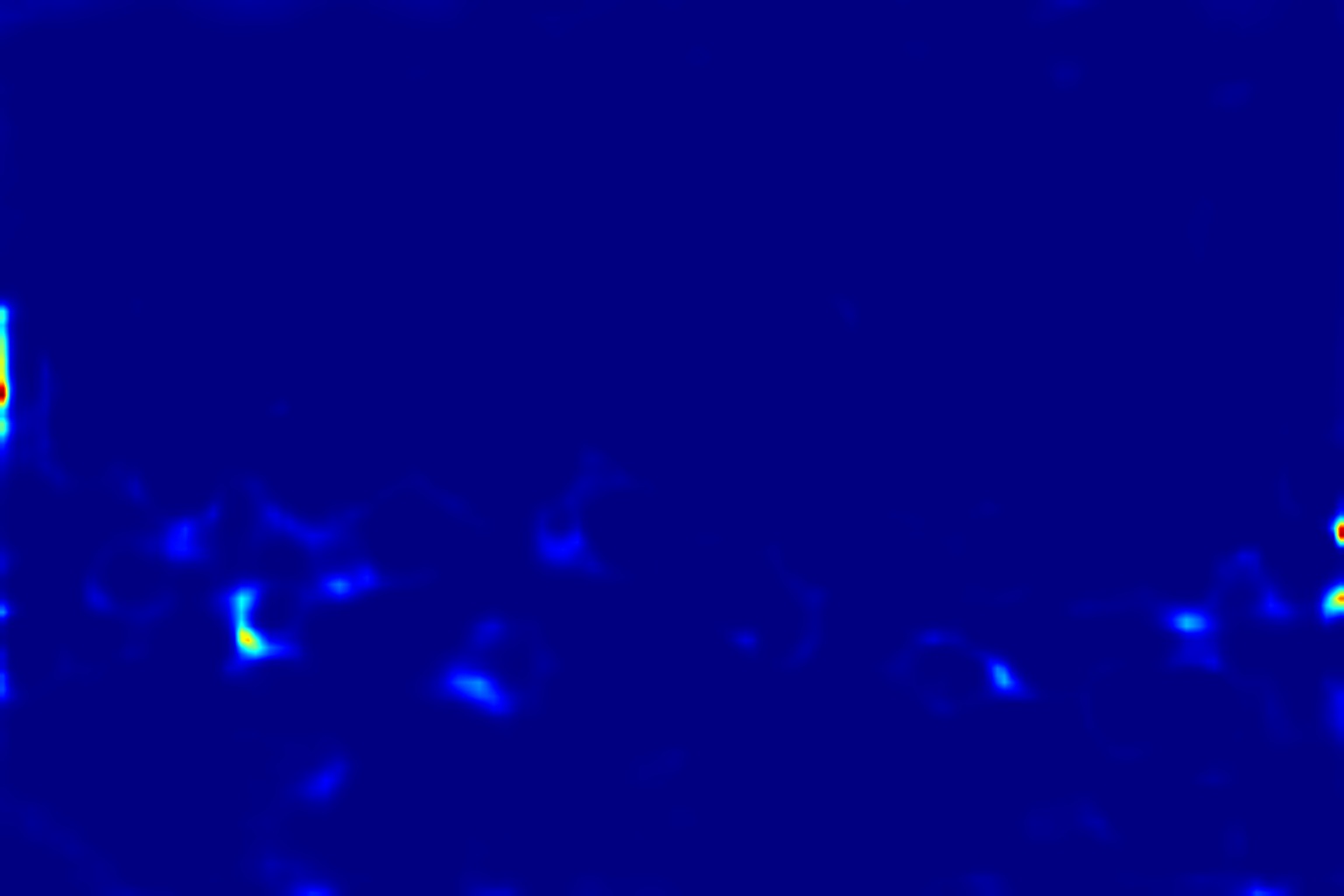} &
			\includegraphics[width=0.14\linewidth]{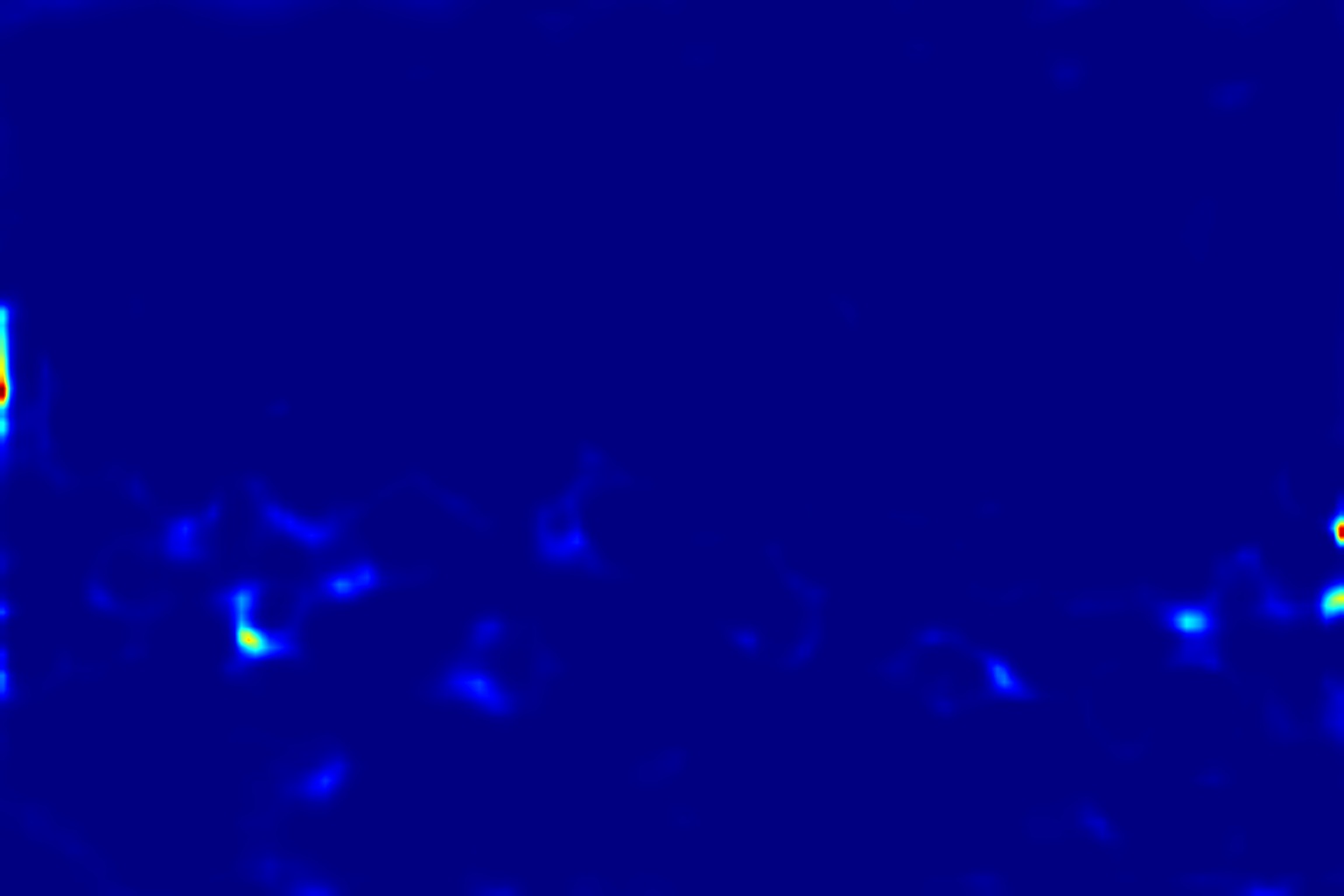}  &
			\includegraphics[width=0.14\linewidth]{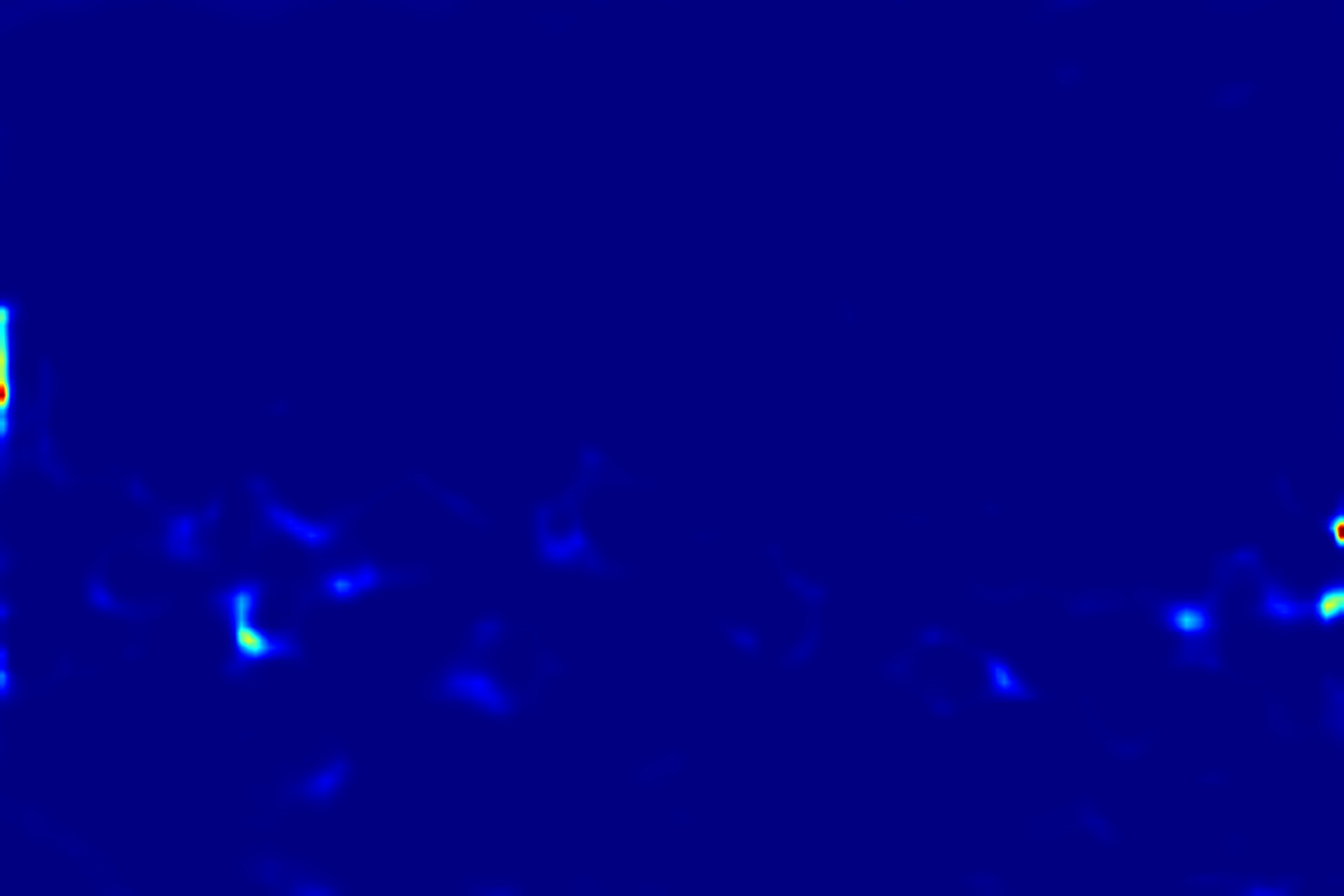} &
			\includegraphics[width=0.14\linewidth]{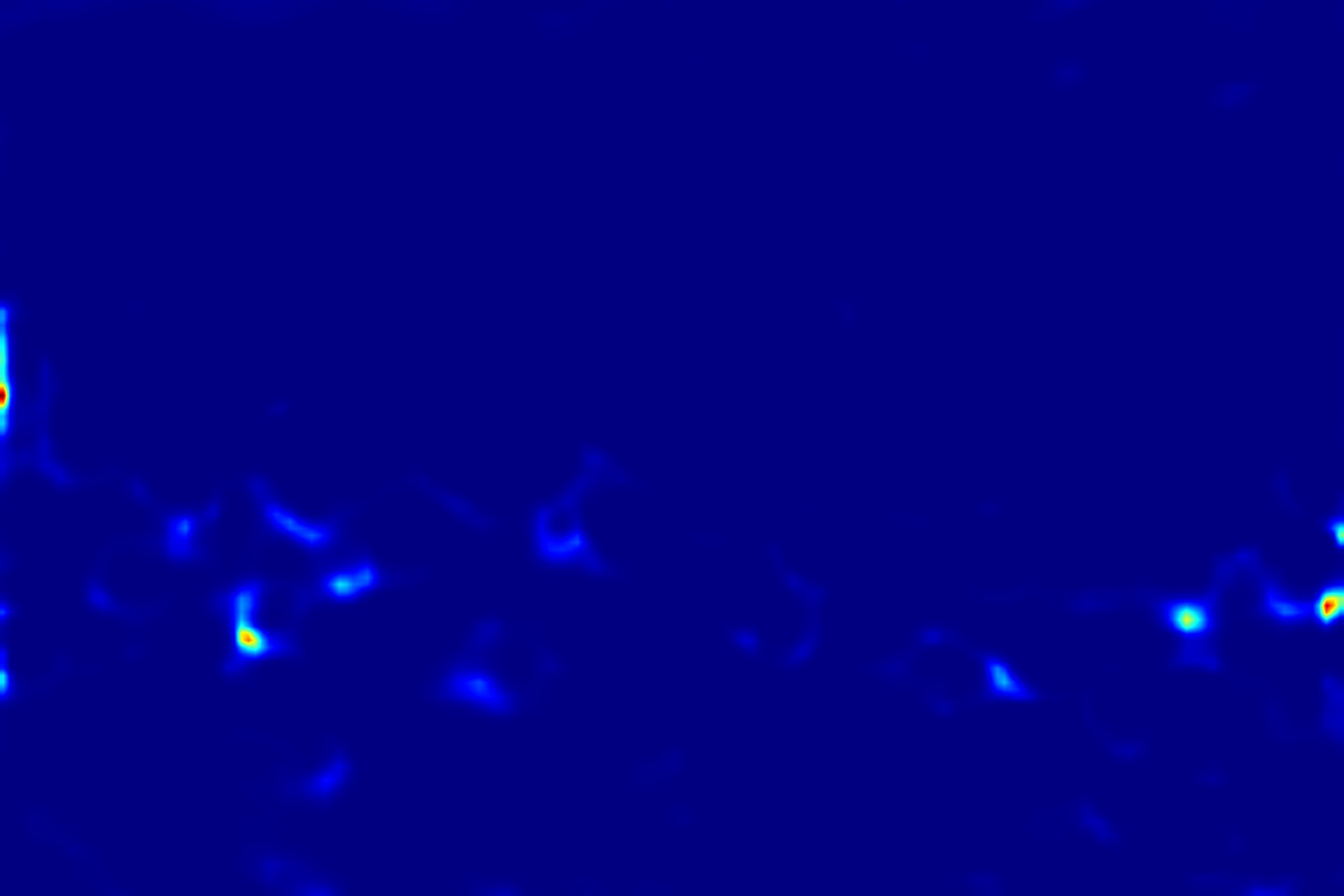} \\
			
			\footnotesize{A12} & \footnotesize{A13} & \footnotesize{A14} & \footnotesize{A15} & \footnotesize{A16} & \footnotesize{A17} \\
			
			\includegraphics[width=0.14\linewidth]{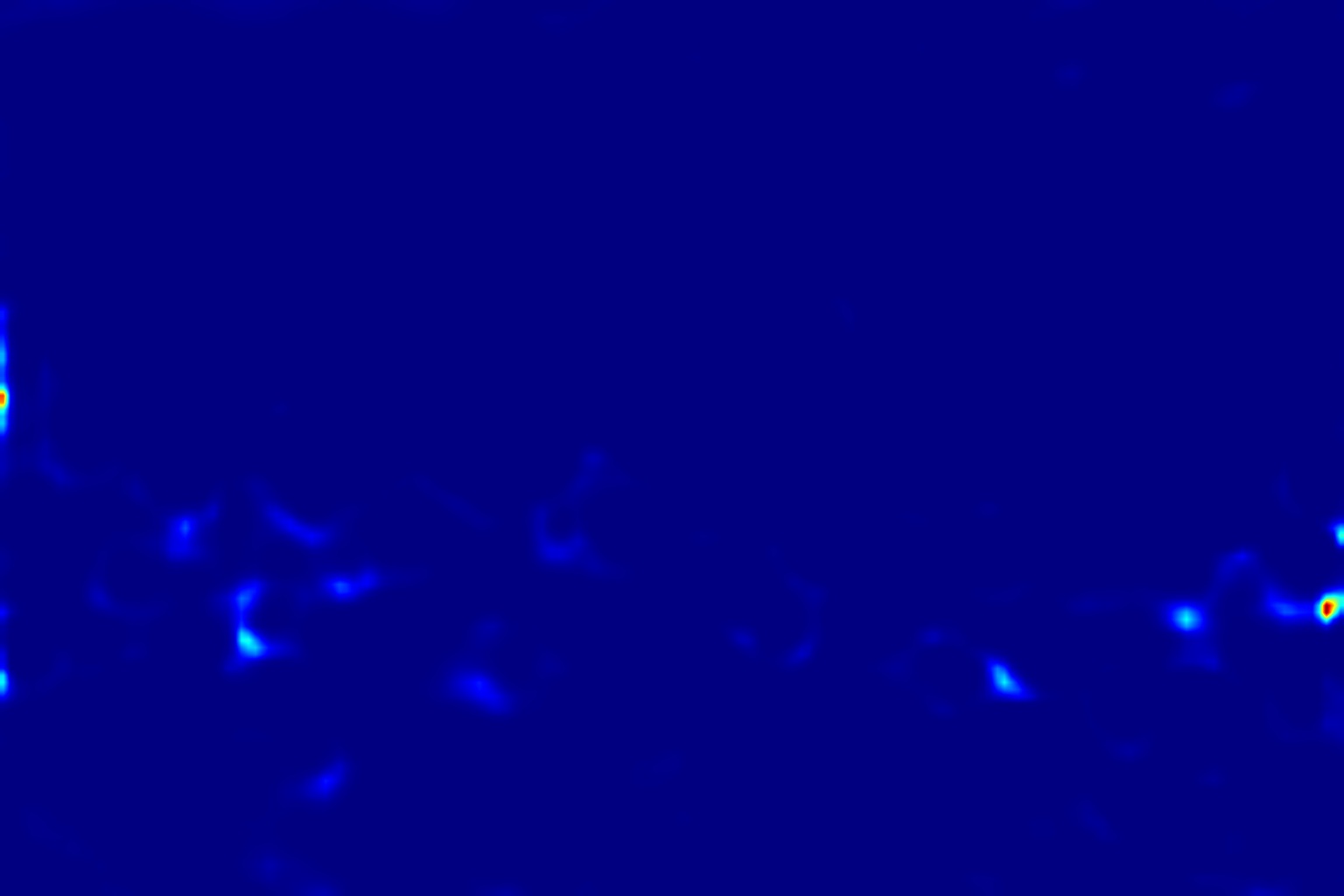}  &
			\includegraphics[width=0.14\linewidth]{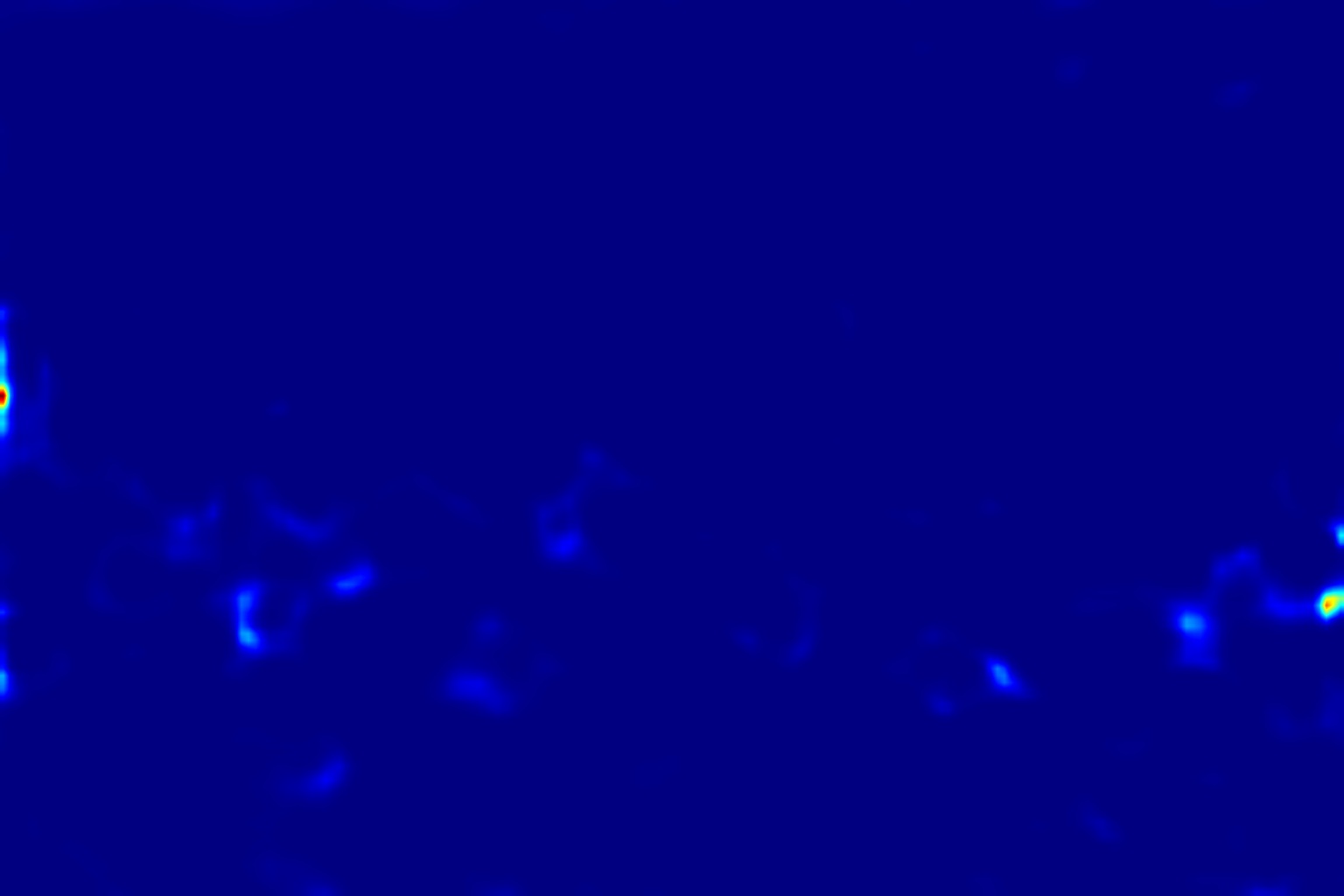}  &
			\includegraphics[width=0.14\linewidth]{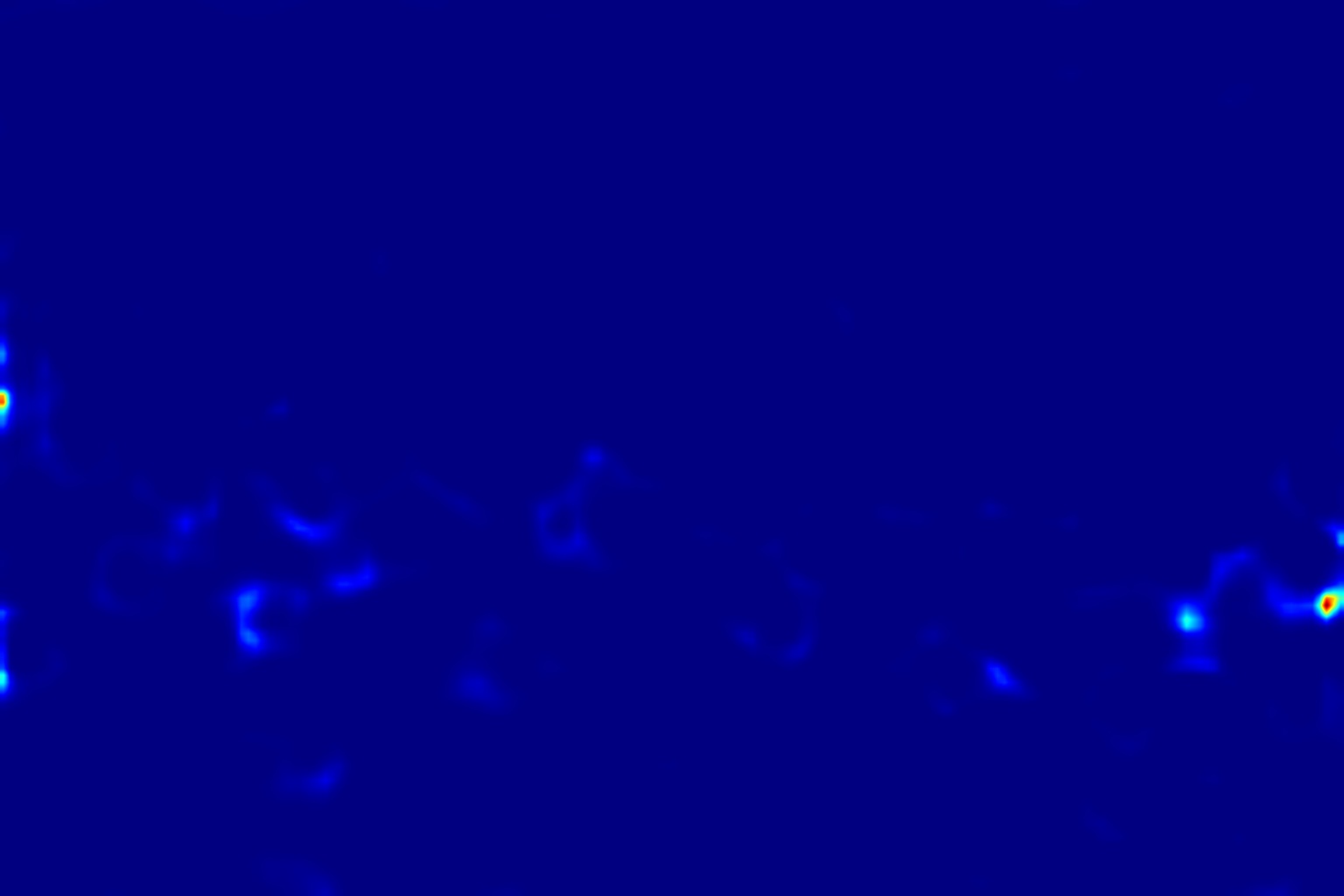} &
			\includegraphics[width=0.14\linewidth]{vis1_123_10.jpg}  &
			\includegraphics[width=0.14\linewidth]{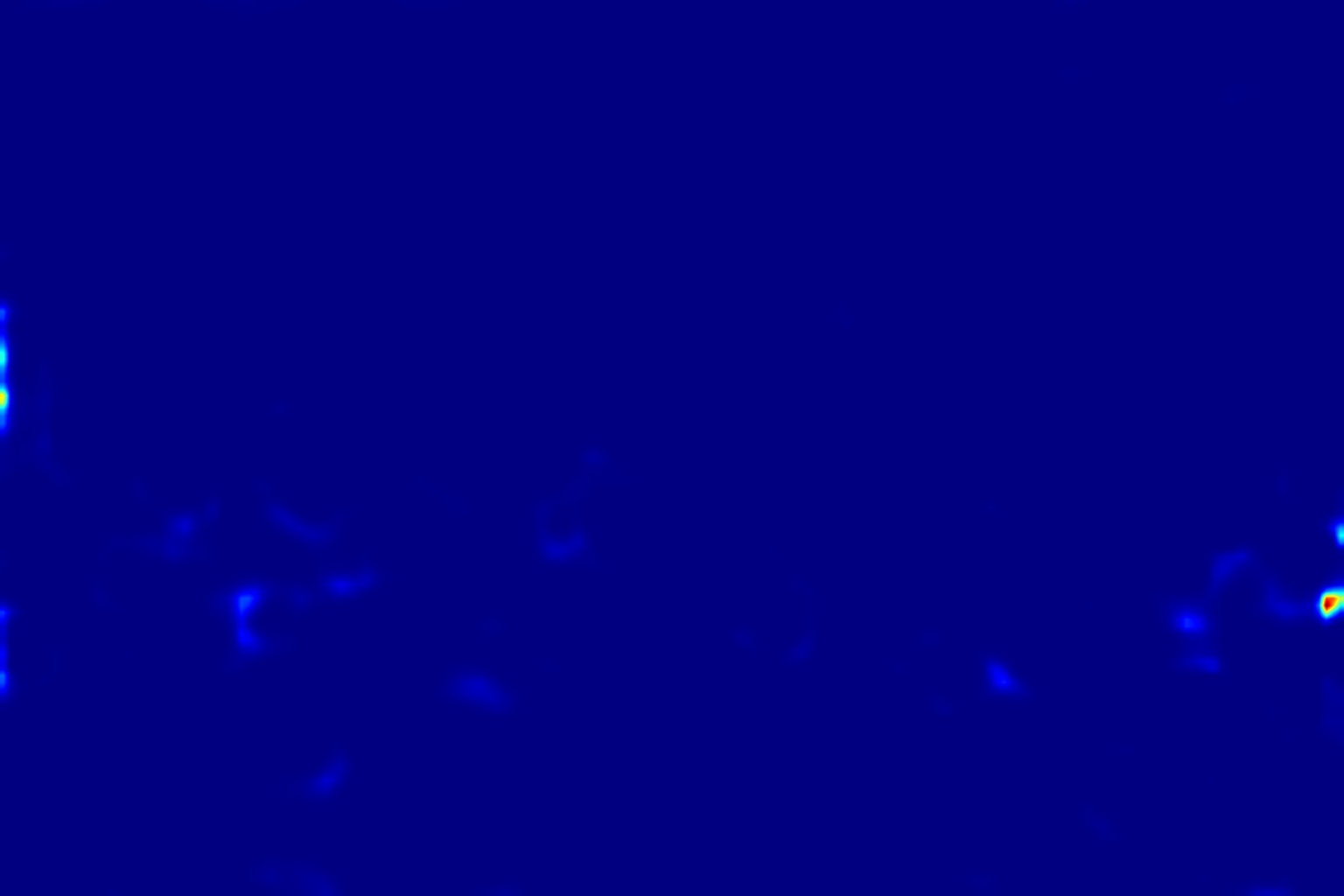} &
			\includegraphics[width=0.14\linewidth]{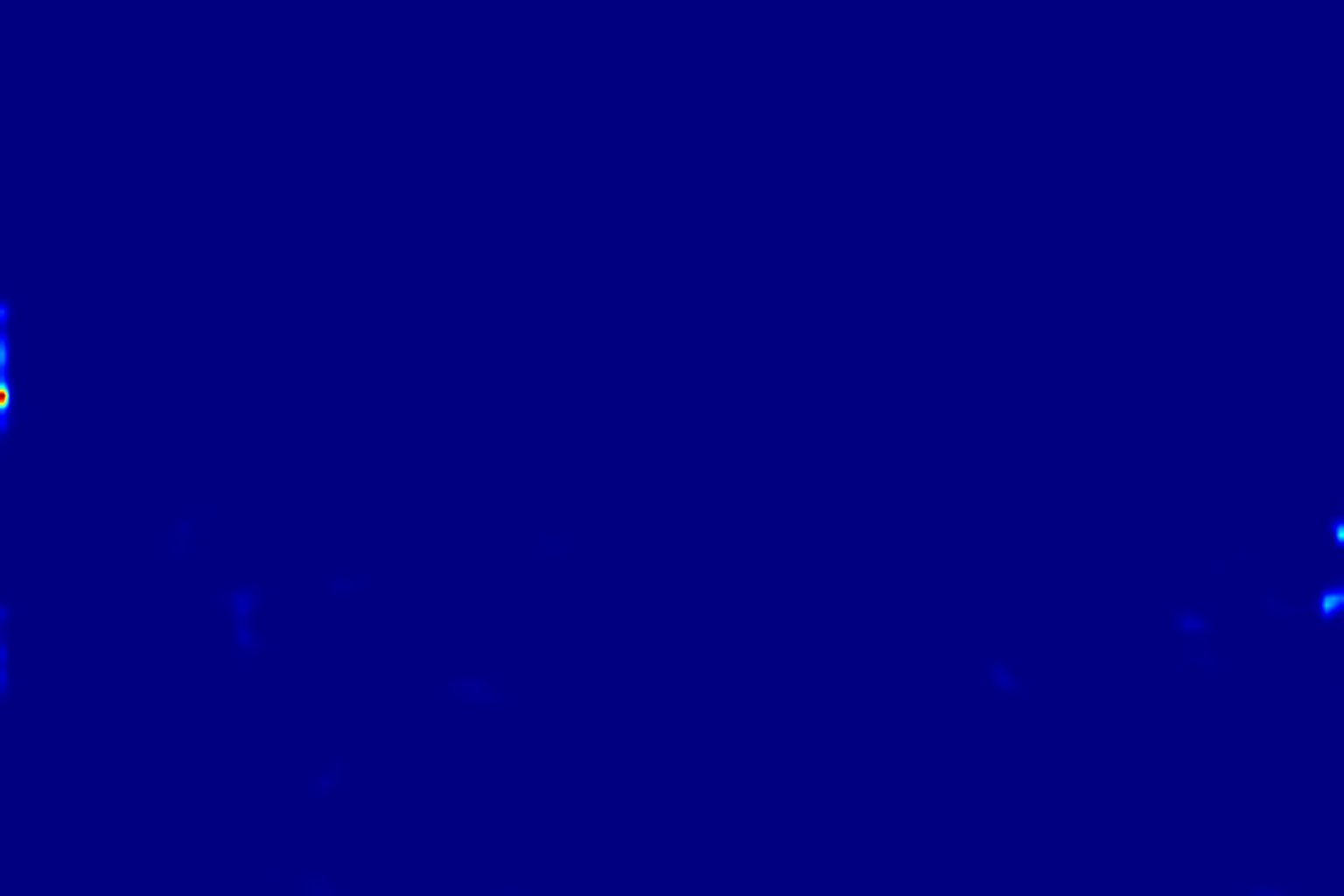} \\
			
			\footnotesize{A18} & \footnotesize{A19} & \footnotesize{A20} & \footnotesize{A21} & \footnotesize{A22} & \footnotesize{A23} \\
			
			\includegraphics[width=0.14\linewidth]{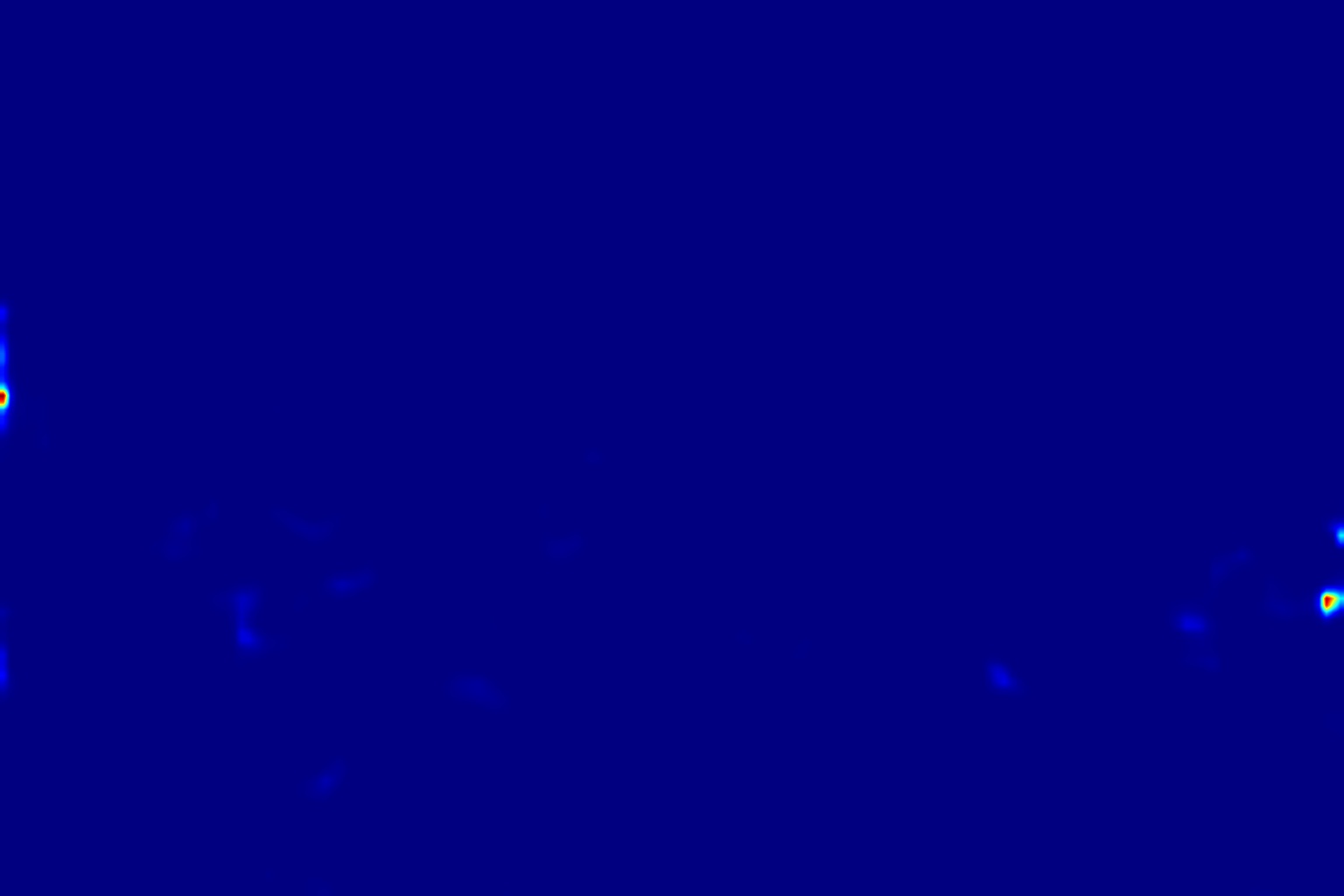}  &
			\includegraphics[width=0.14\linewidth]{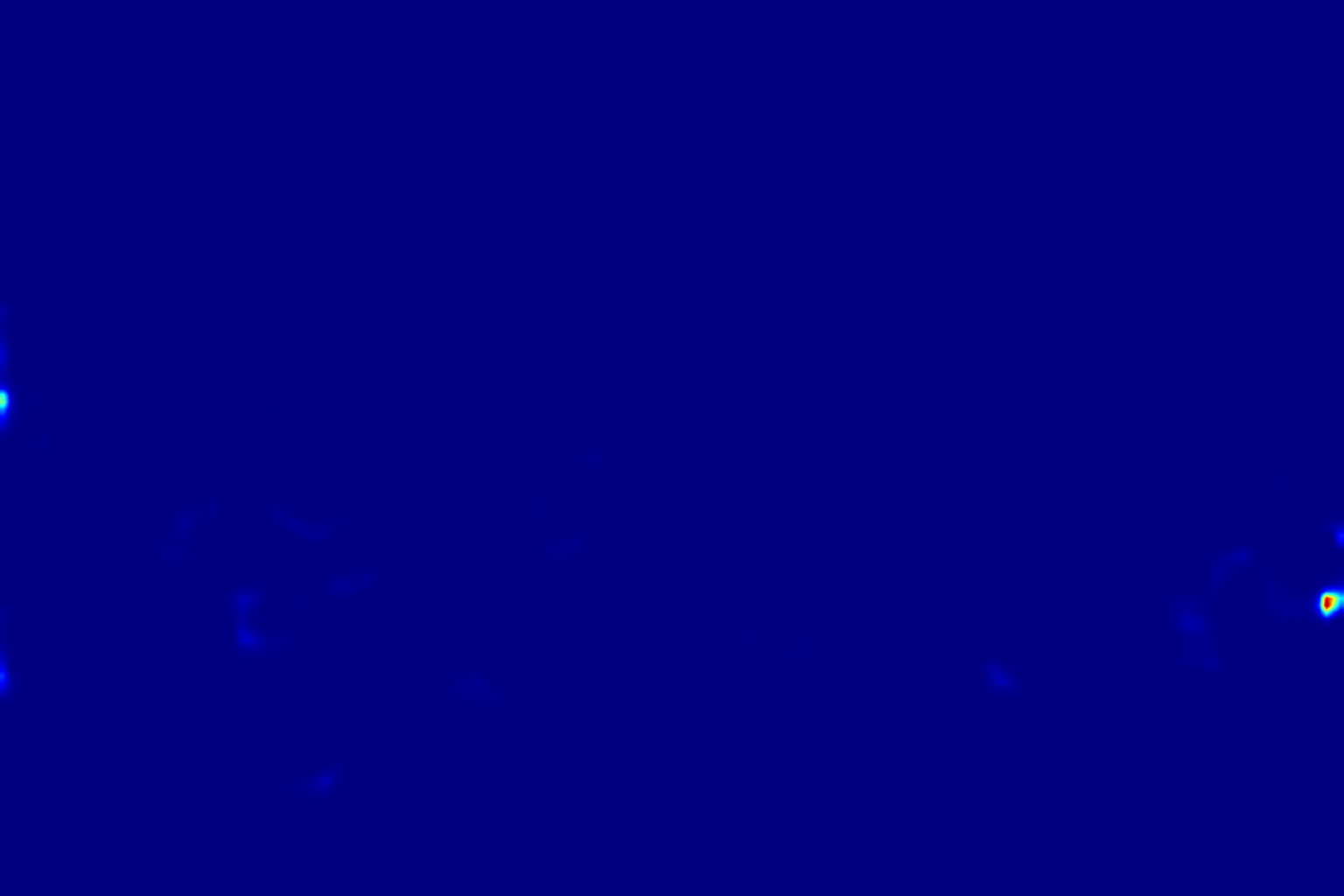}  &
			\includegraphics[width=0.14\linewidth]{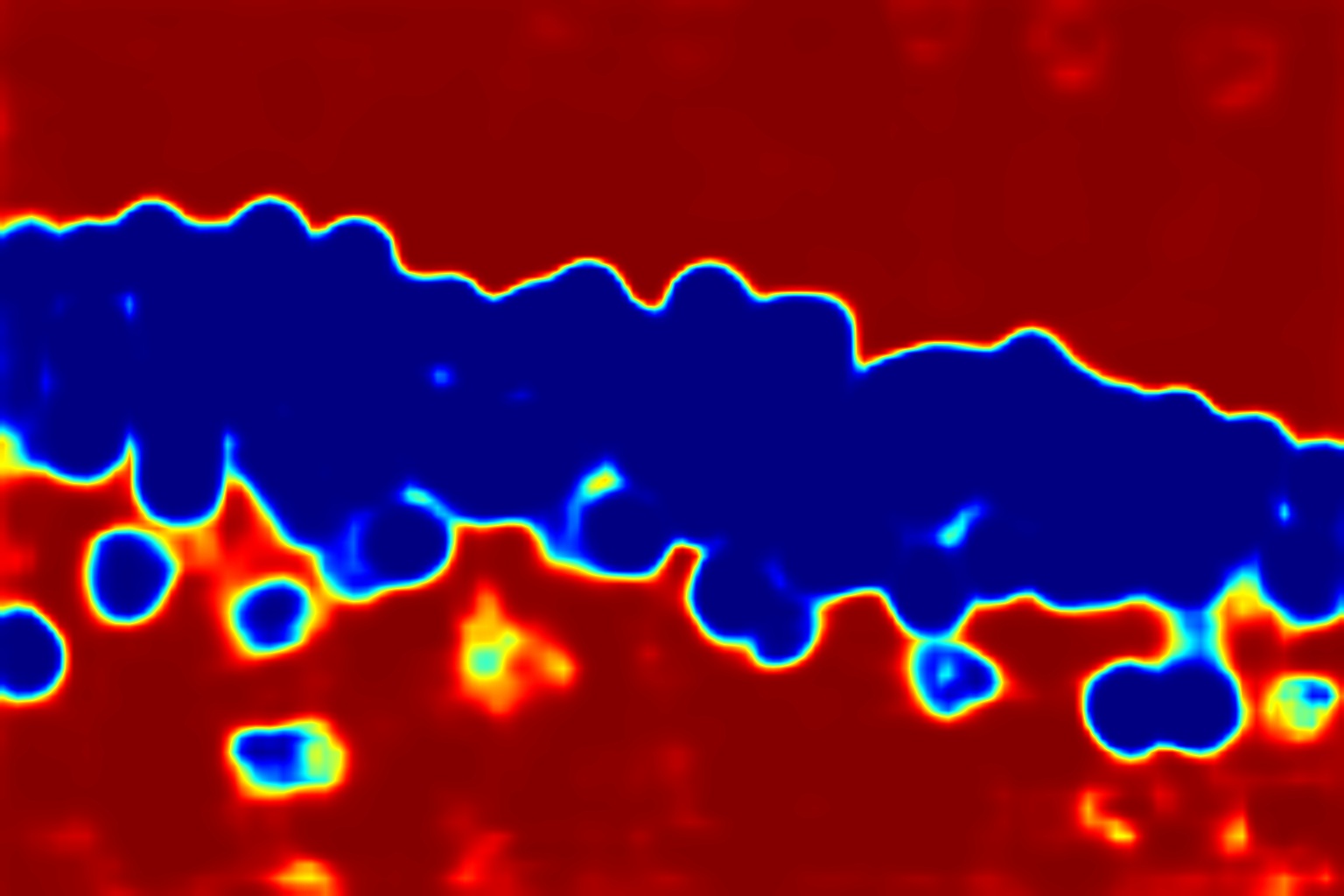} &
			\includegraphics[width=0.14\linewidth]{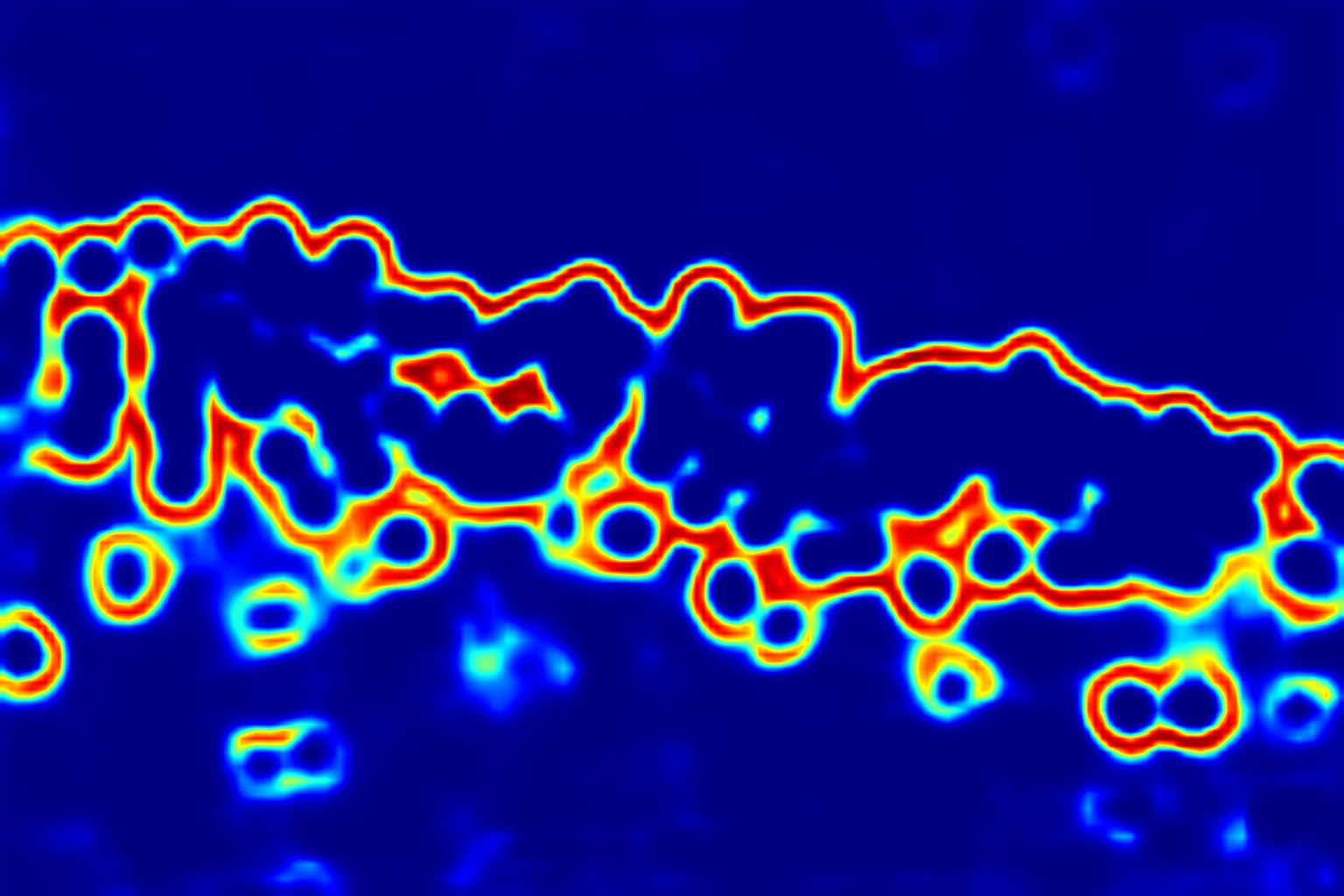}  &
			\includegraphics[width=0.14\linewidth]{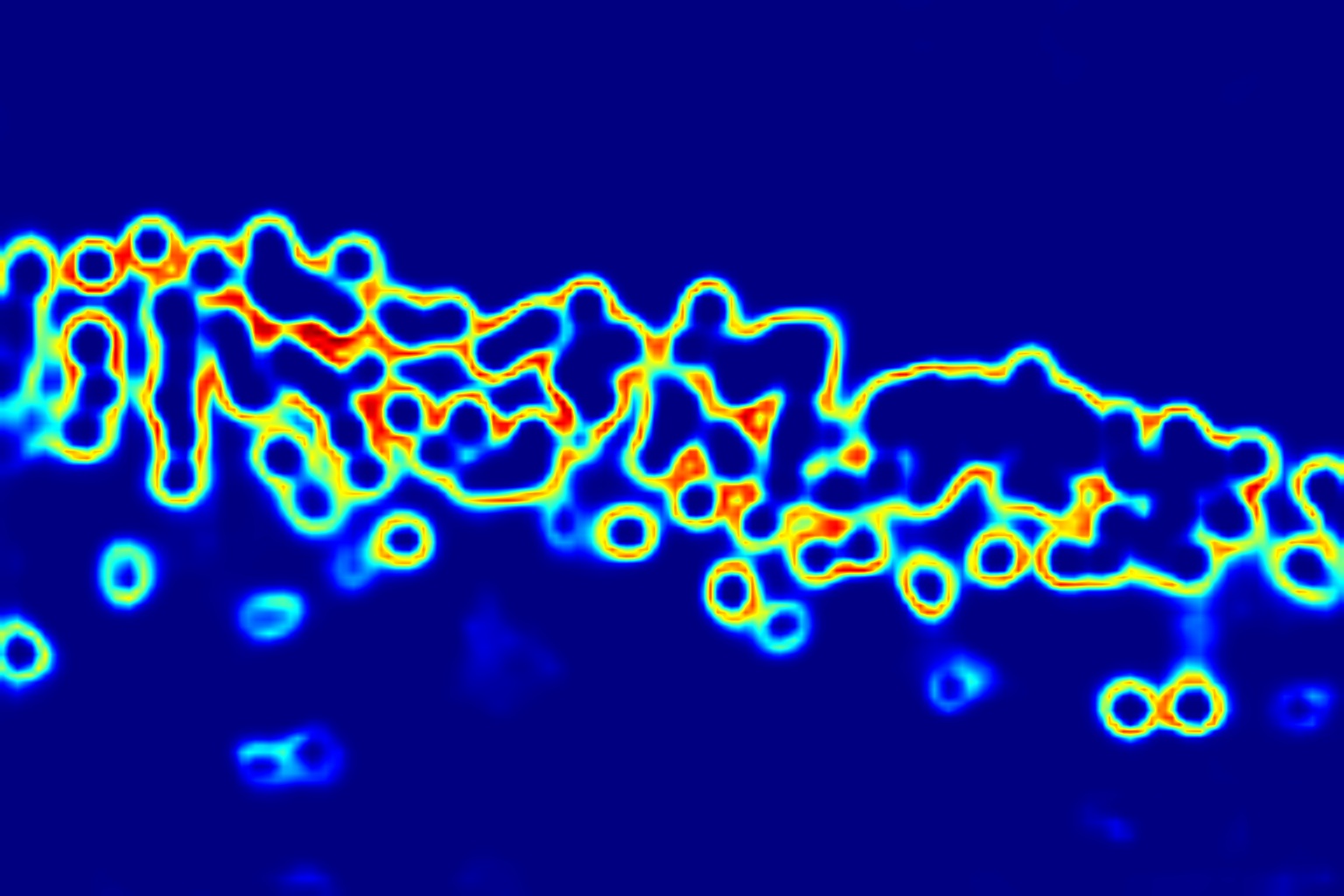} &
			\includegraphics[width=0.14\linewidth]{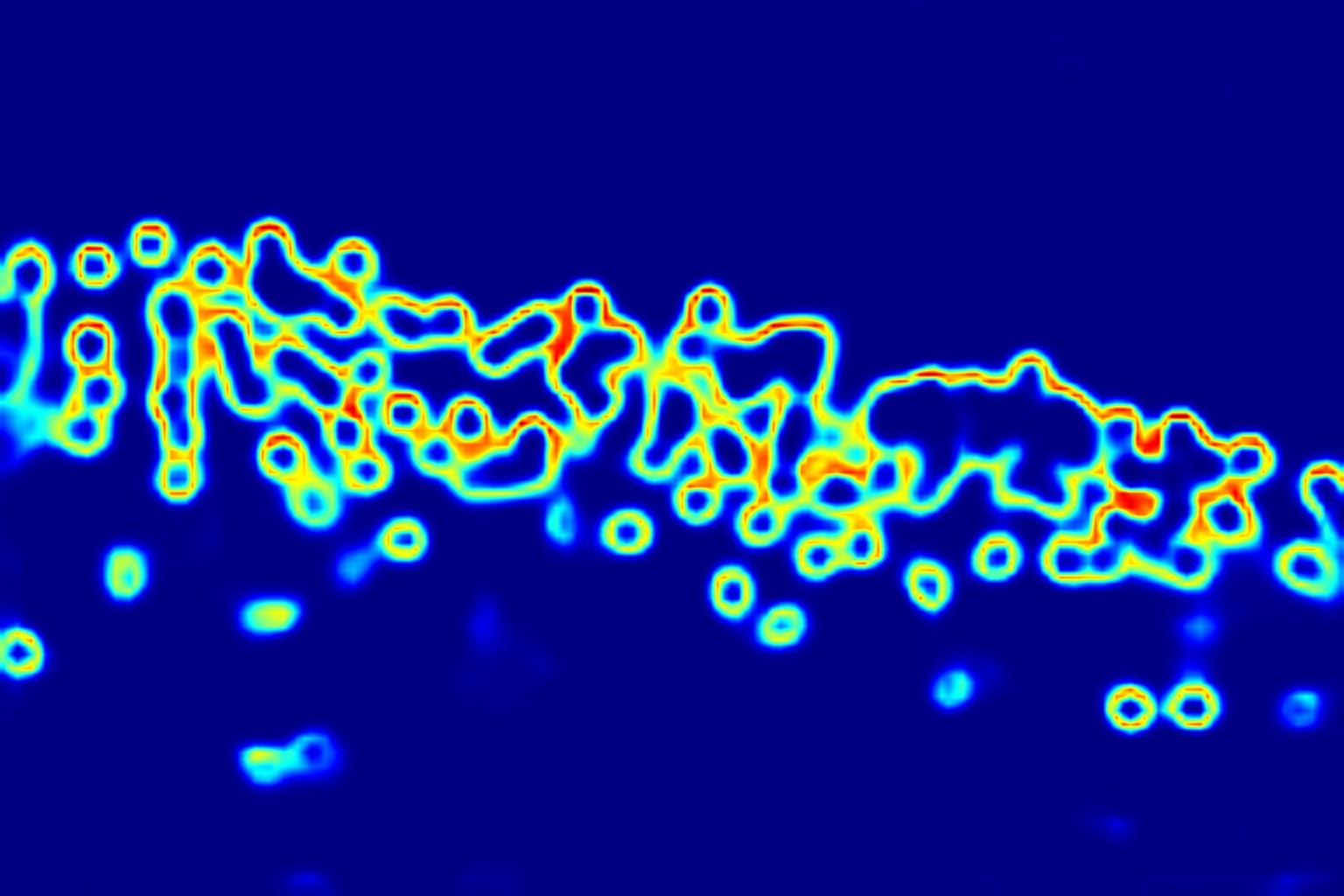} \\
			
			\footnotesize{A24} & \footnotesize{A25} & \footnotesize{B1} & \footnotesize{B2} & \footnotesize{B3} & \footnotesize{B4} \\
			
			\includegraphics[width=0.14\linewidth]{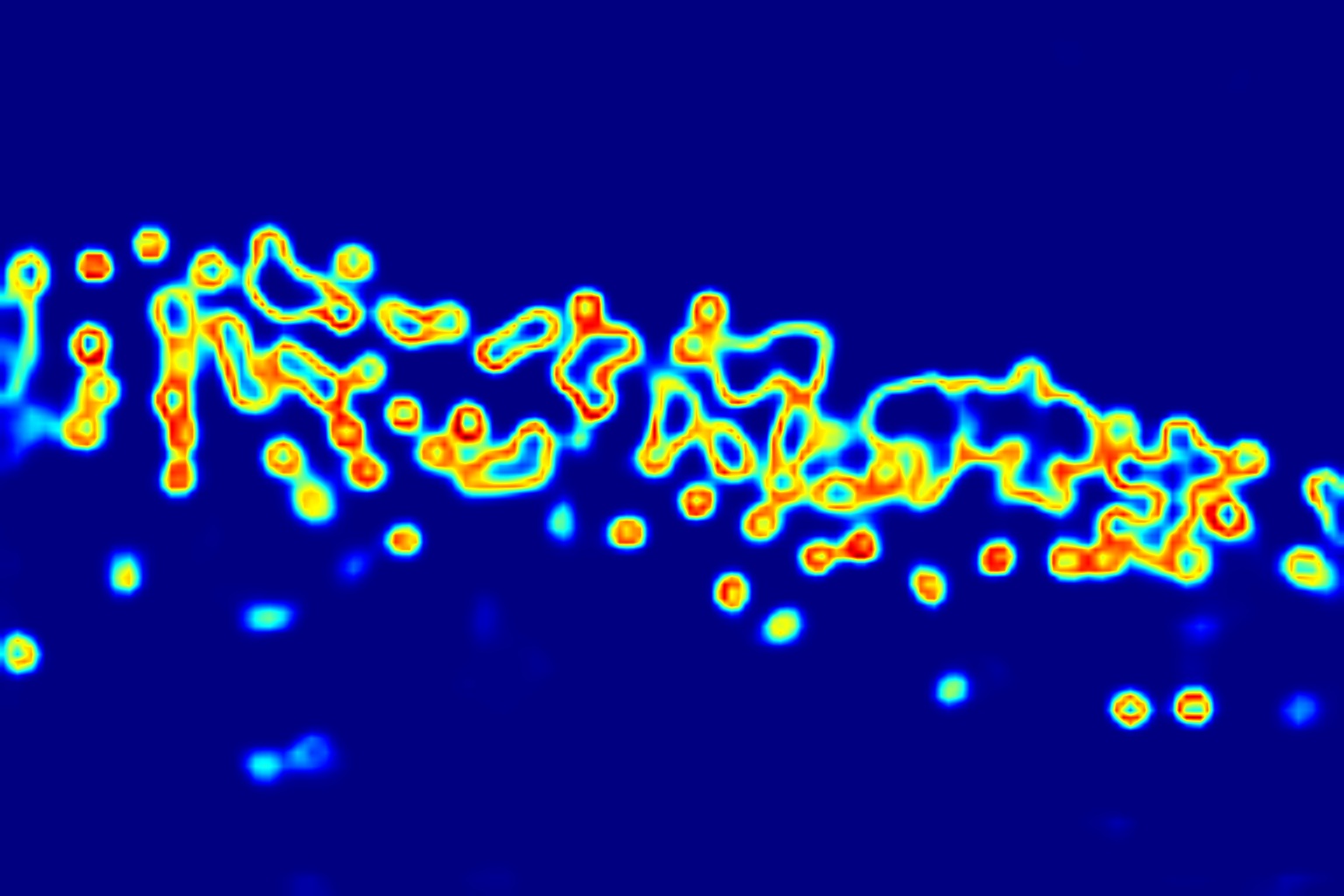}  &
			\includegraphics[width=0.14\linewidth]{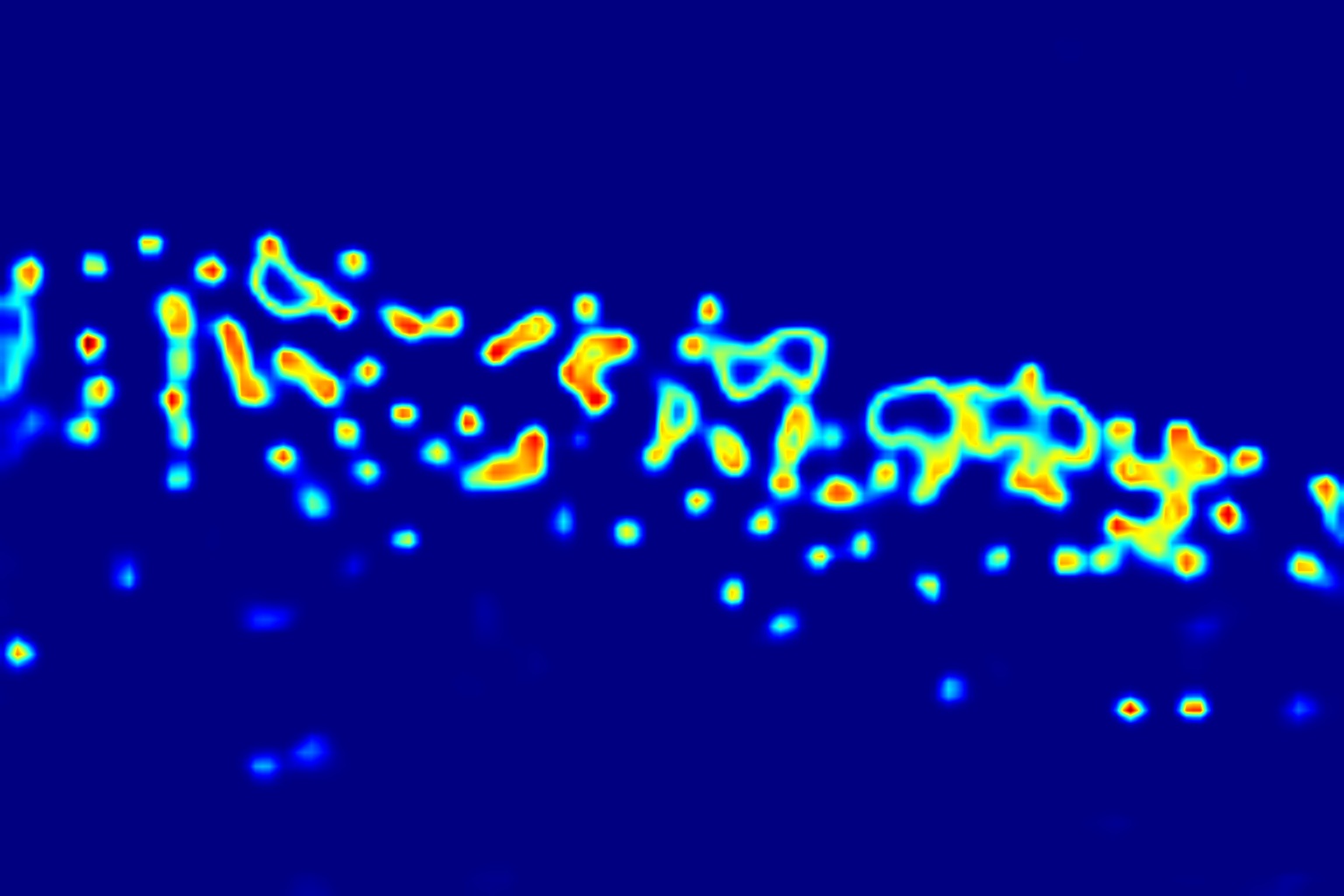}  &
			\includegraphics[width=0.14\linewidth]{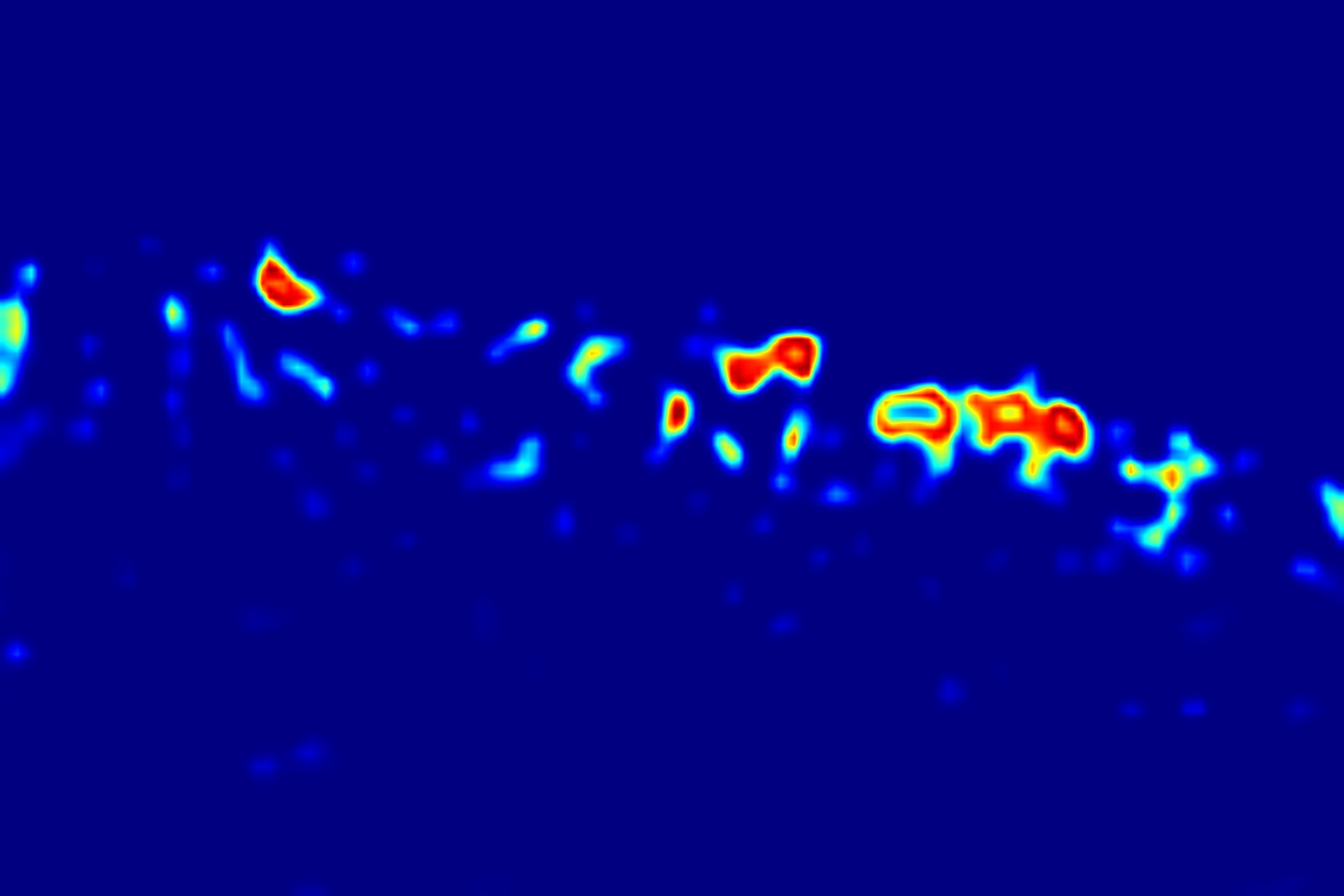} &
			\includegraphics[width=0.14\linewidth]{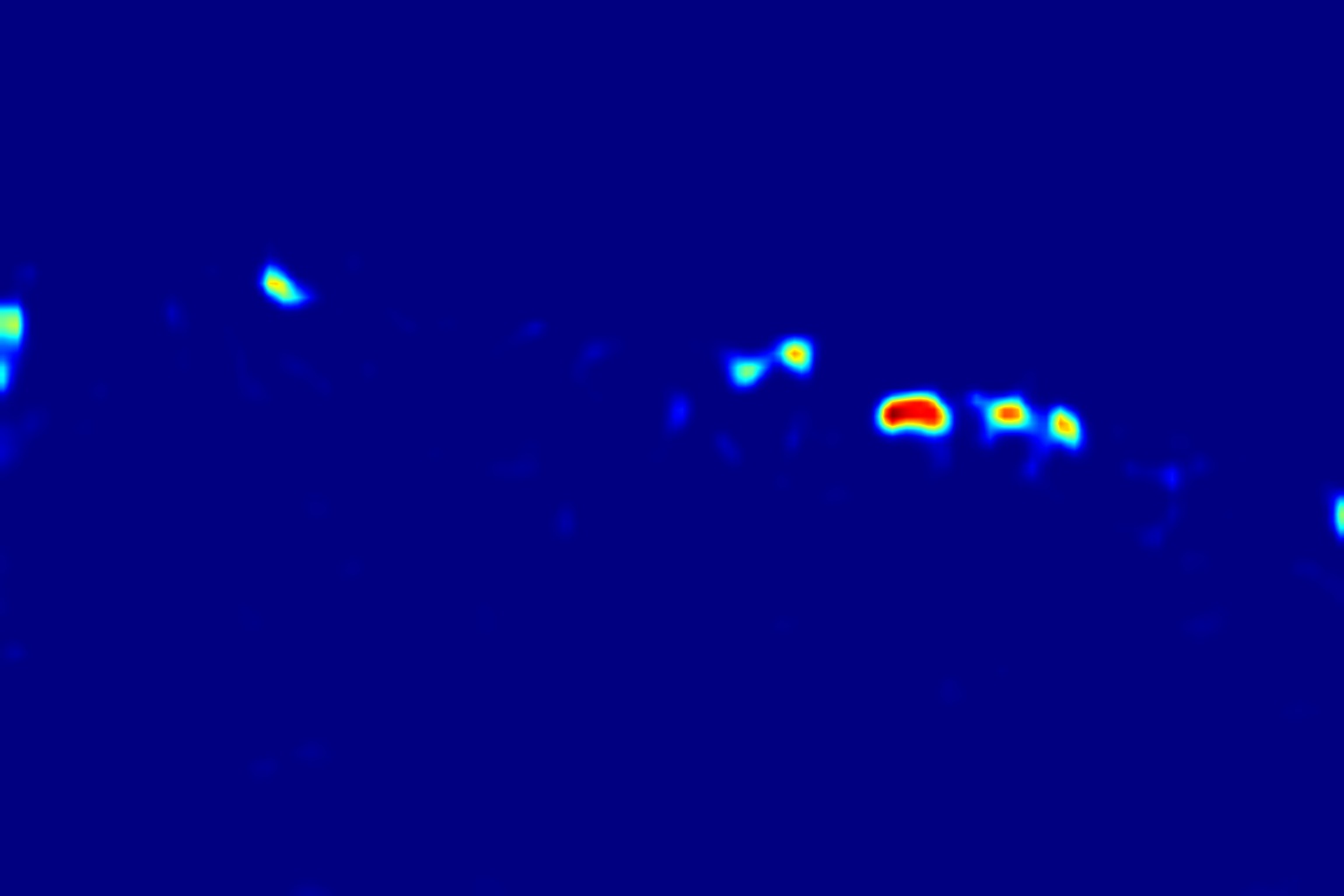}  &
			\includegraphics[width=0.14\linewidth]{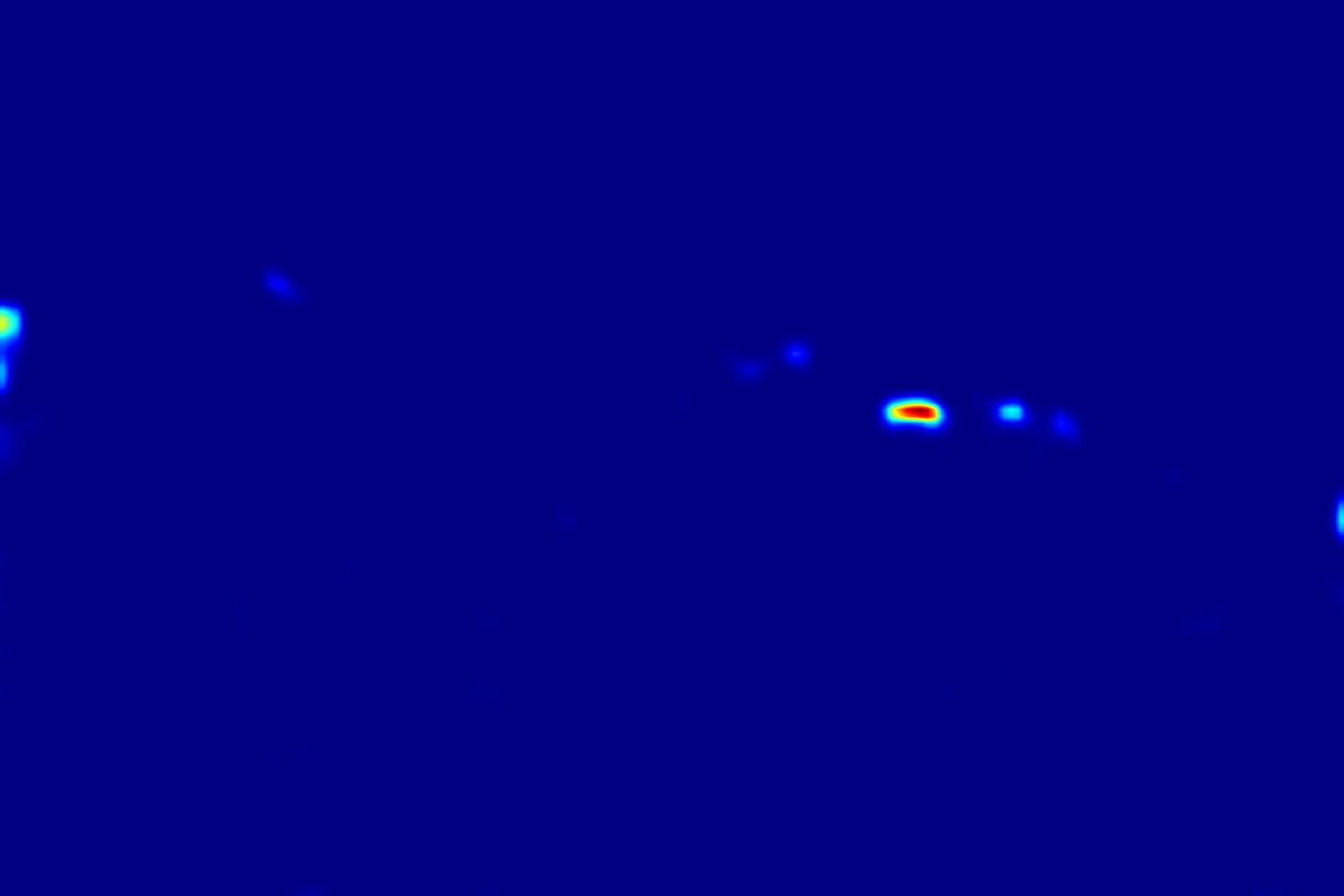} &
			\includegraphics[width=0.14\linewidth]{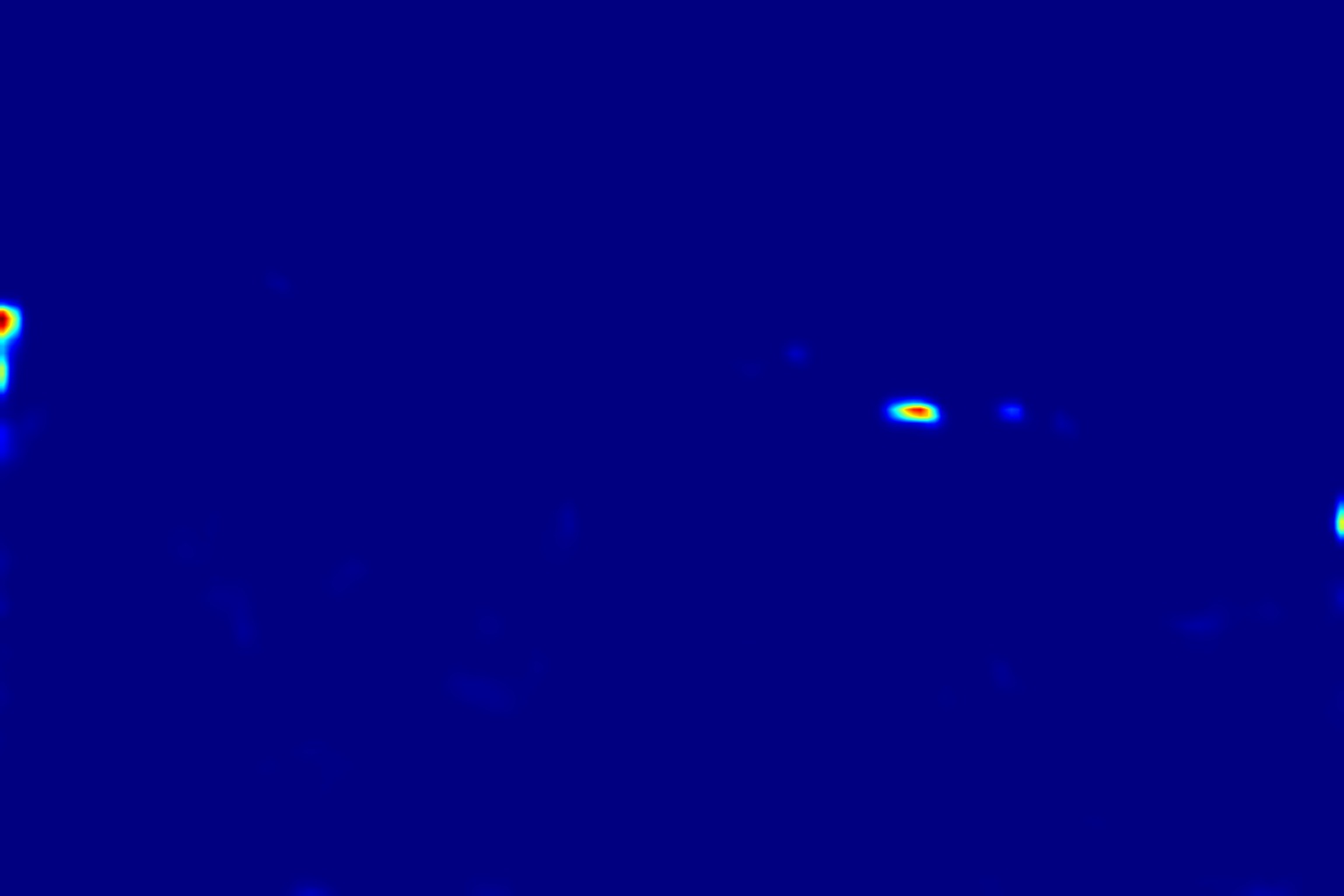} \\
			
			\footnotesize{B5} & \footnotesize{B6} & \footnotesize{B7} & \footnotesize{B8} & \footnotesize{B9} & \footnotesize{B10} \\
			
			\includegraphics[width=0.14\linewidth]{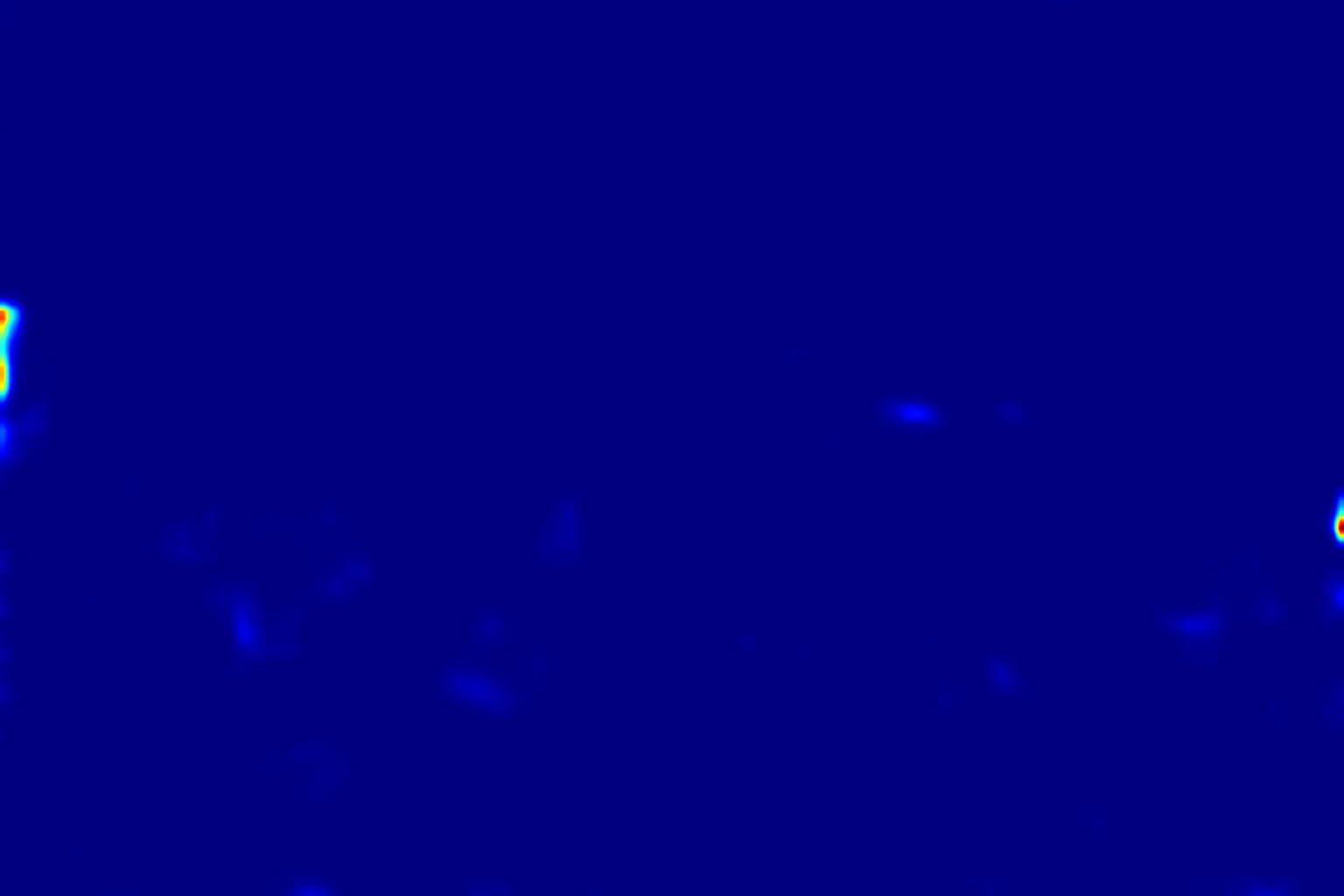}  &
			\includegraphics[width=0.14\linewidth]{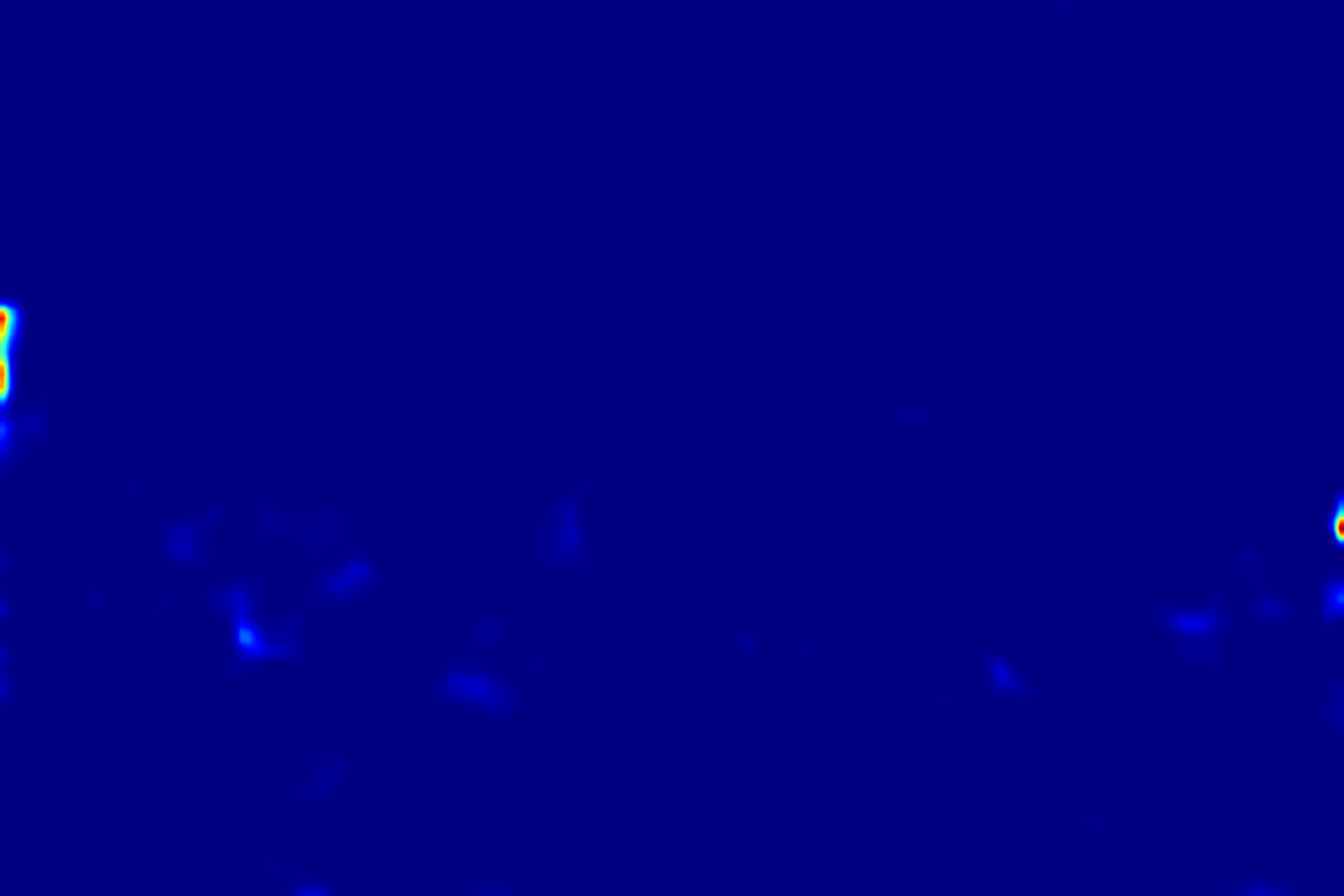}  &
			\includegraphics[width=0.14\linewidth]{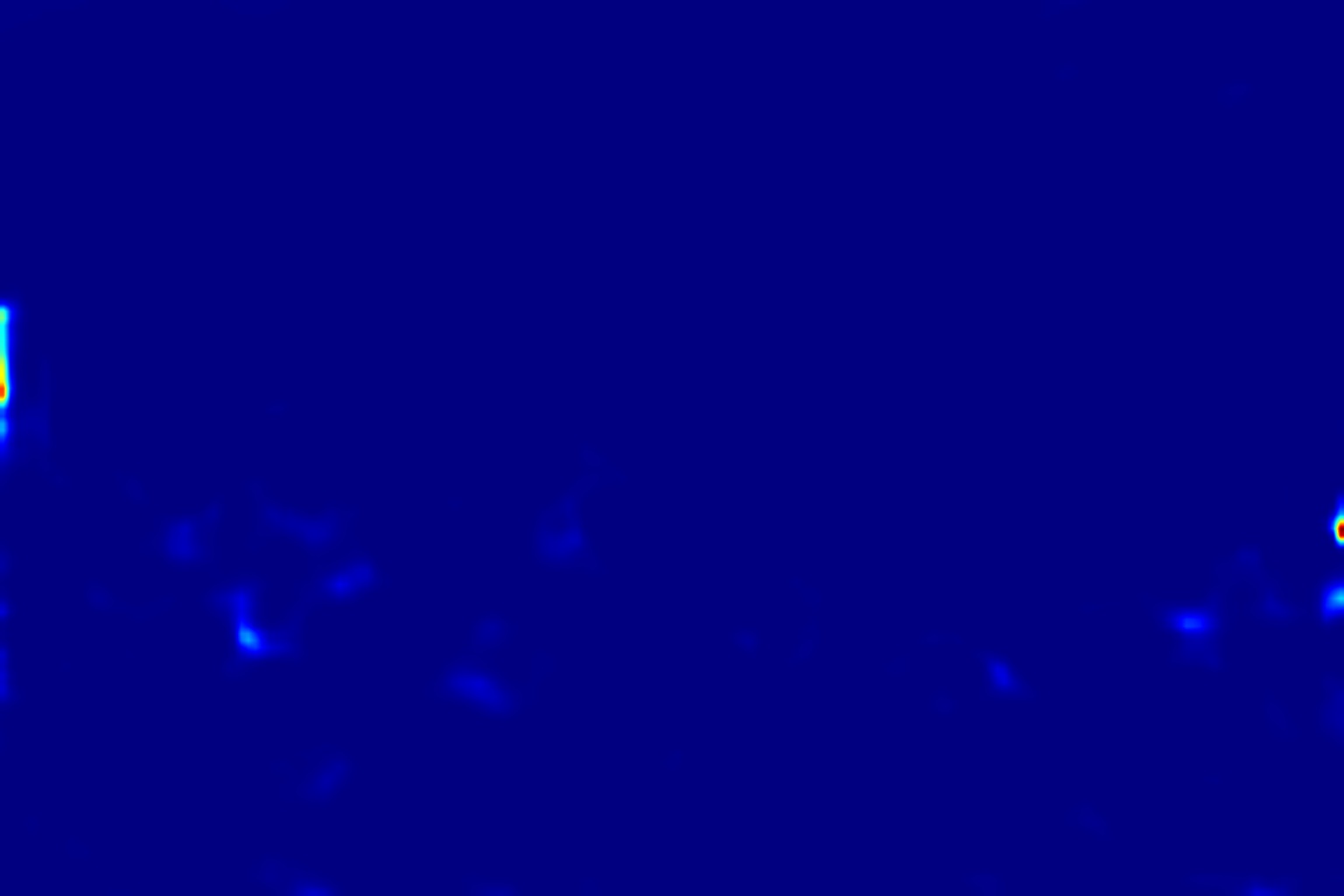} &
			\includegraphics[width=0.14\linewidth]{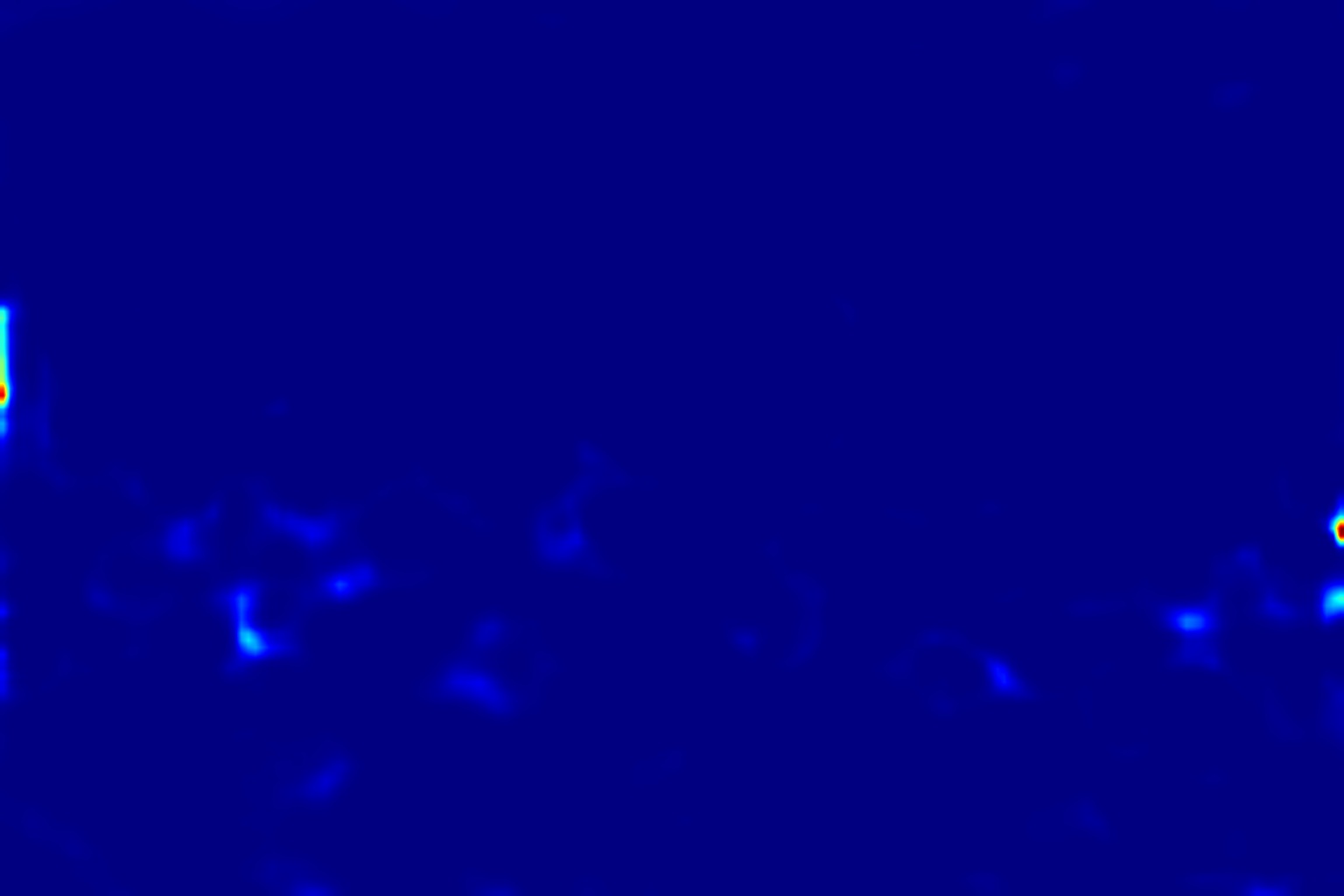}  &
			\includegraphics[width=0.14\linewidth]{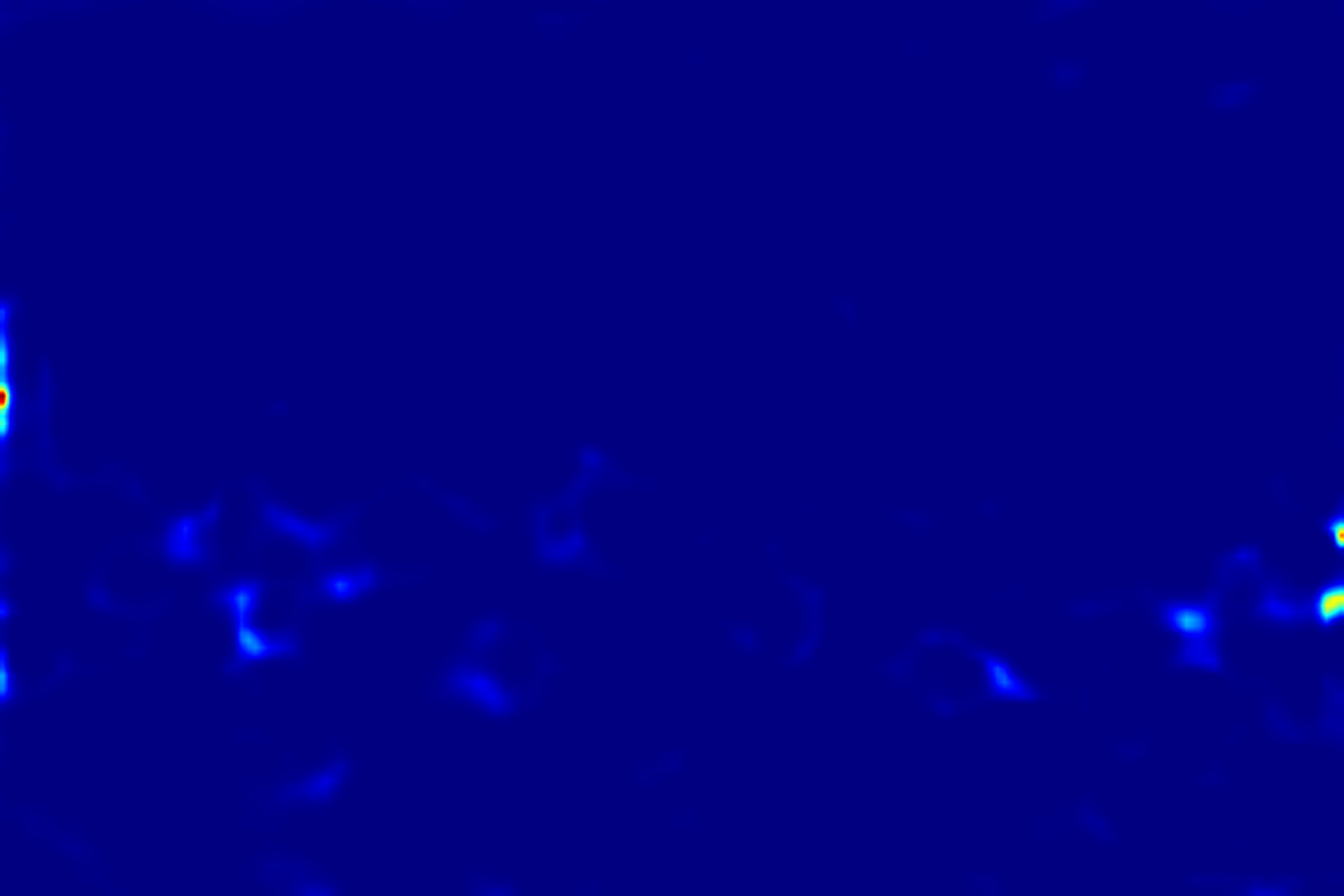} &
			\includegraphics[width=0.14\linewidth]{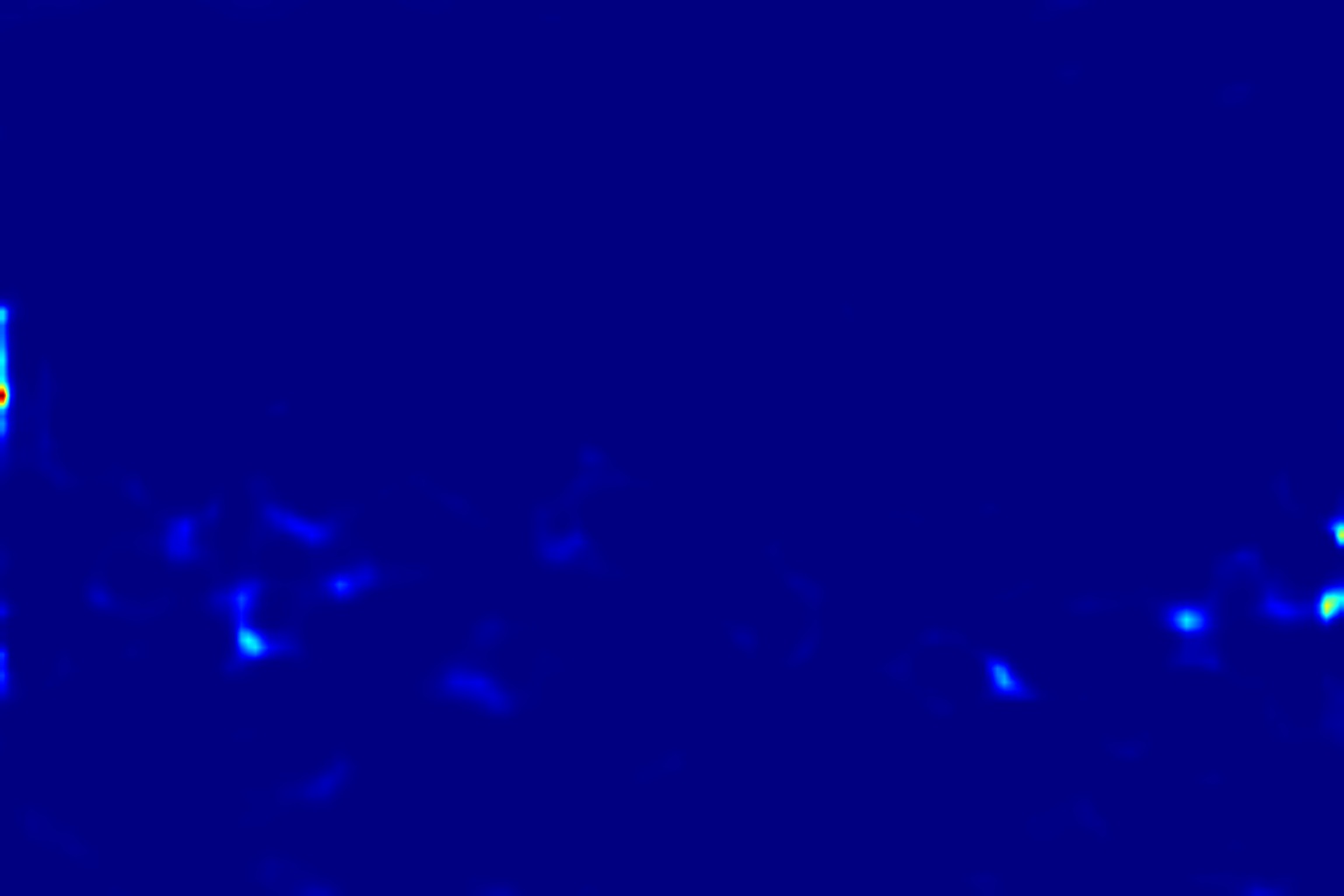} \\
			
			\footnotesize{B11} & \footnotesize{B12} & \footnotesize{B13} & \footnotesize{B14} & \footnotesize{B15} & \footnotesize{B16} \\
			
			\includegraphics[width=0.14\linewidth]{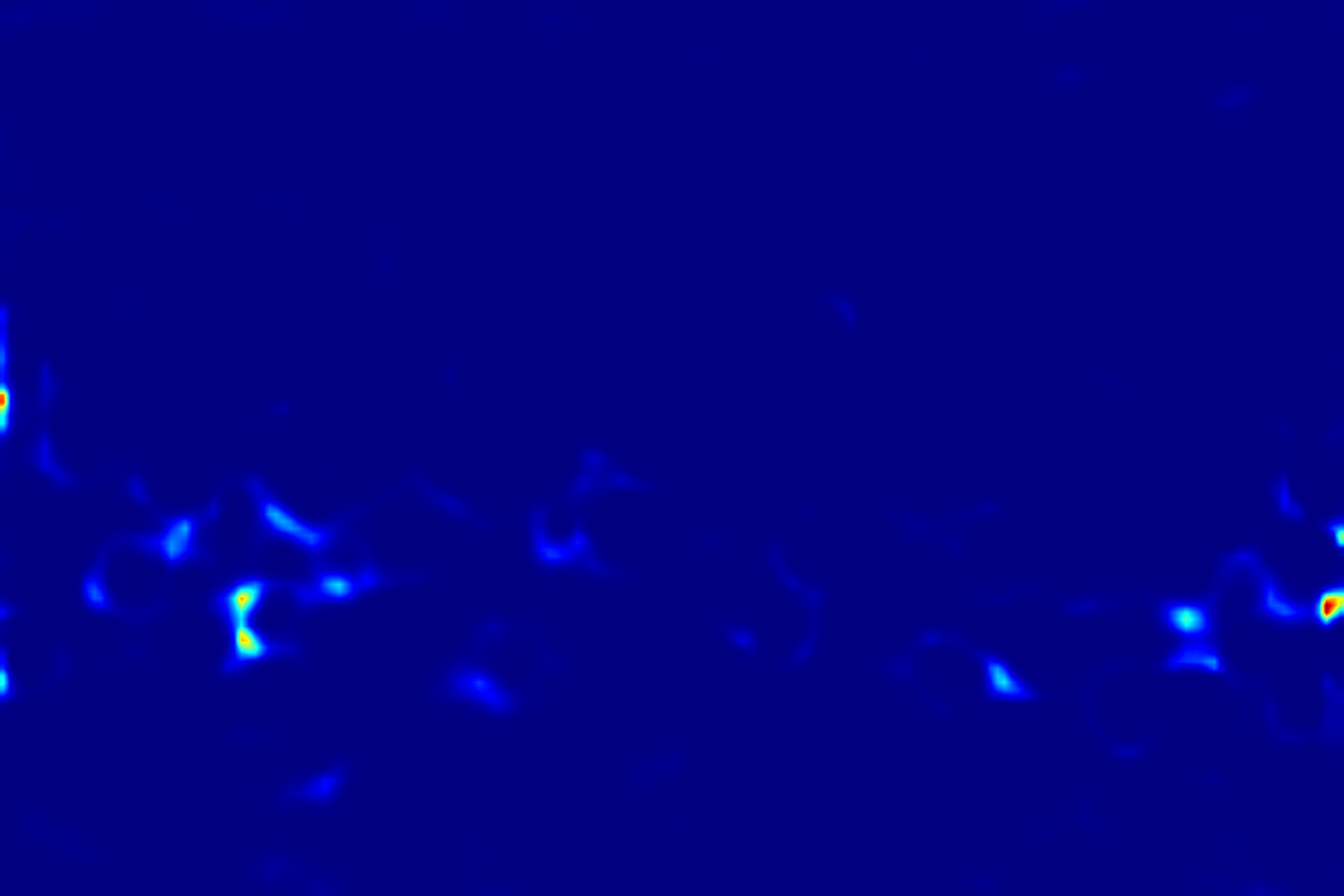}  &
			\includegraphics[width=0.14\linewidth]{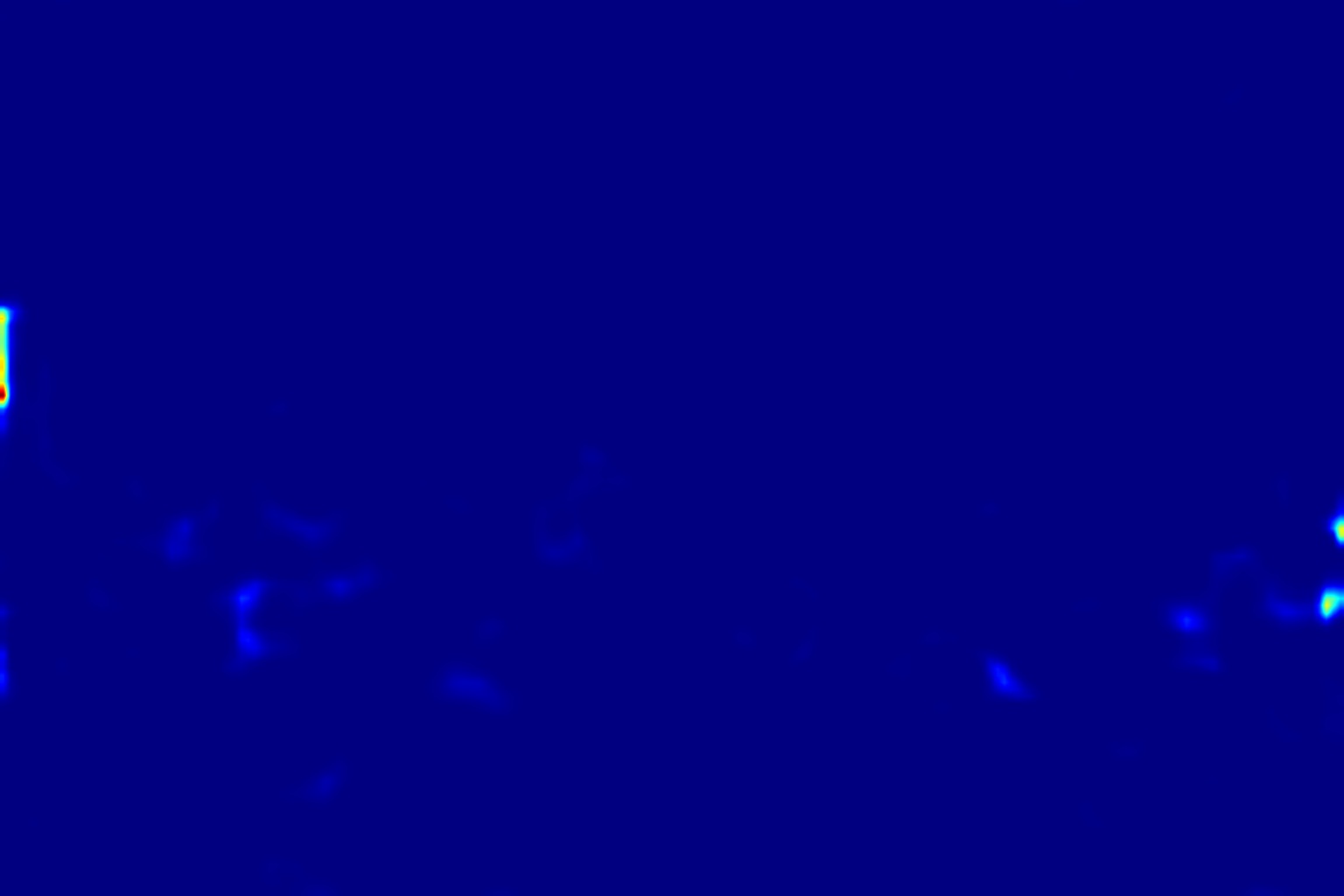}  &
			\includegraphics[width=0.14\linewidth]{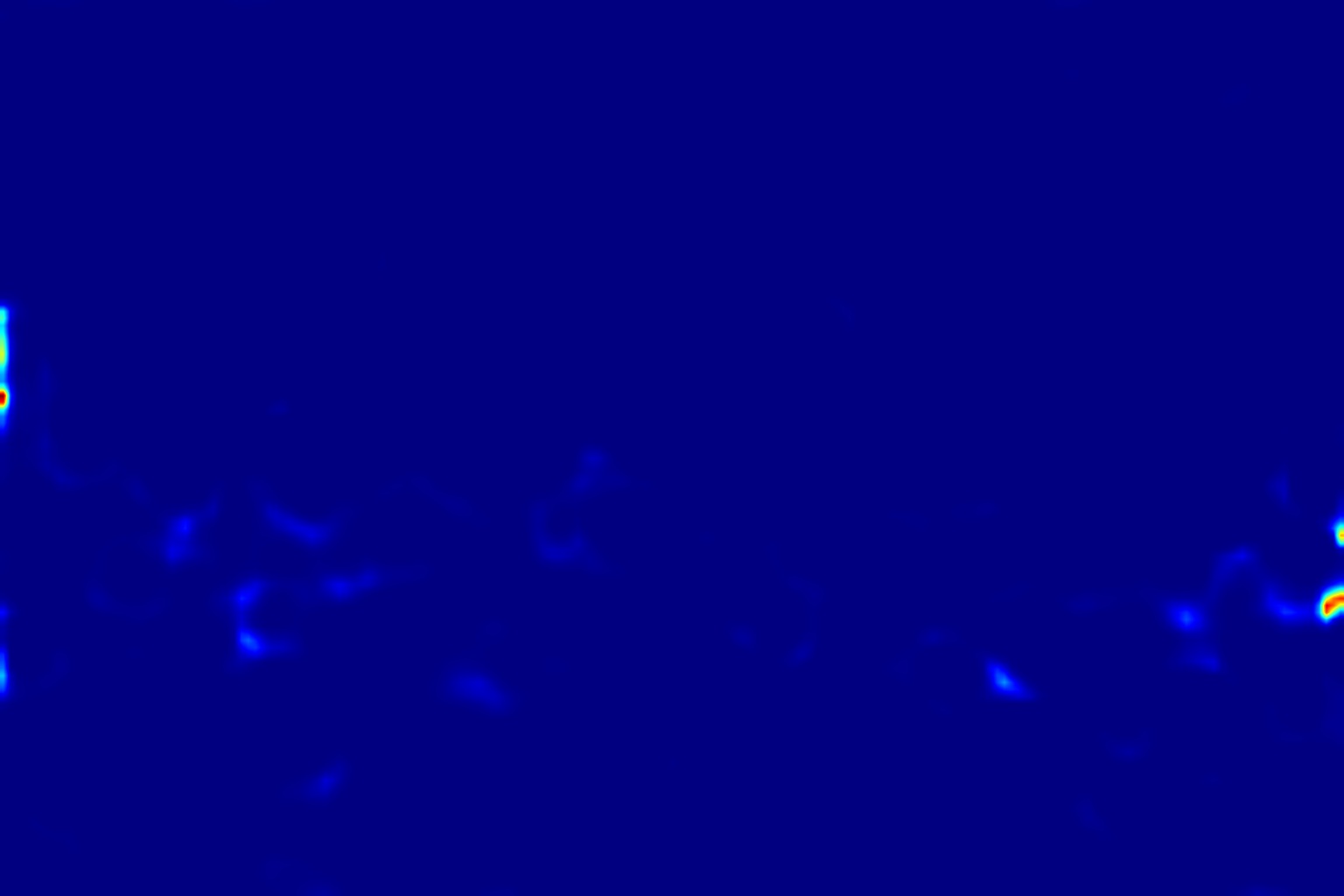} &
			\includegraphics[width=0.14\linewidth]{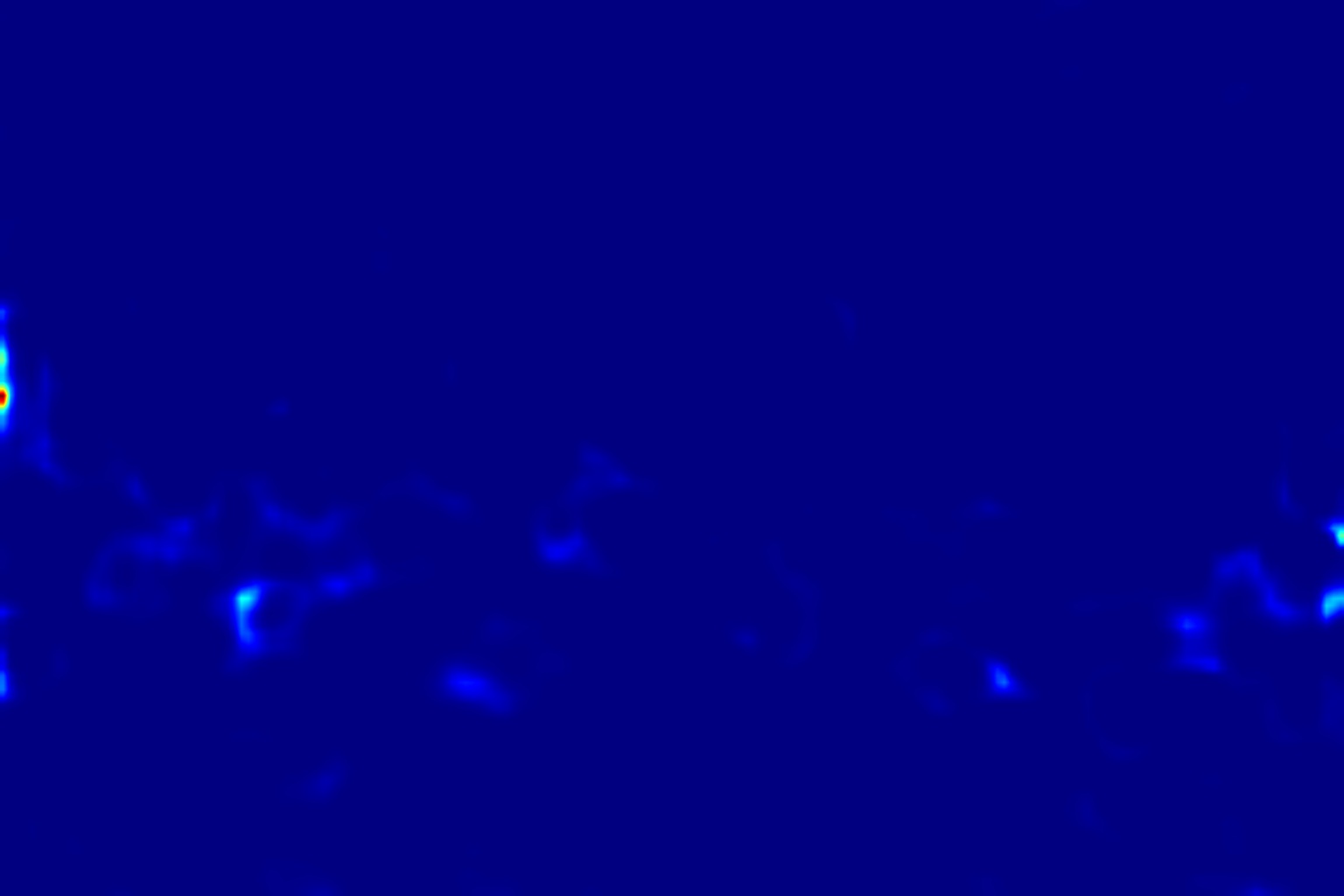}  &
			\includegraphics[width=0.14\linewidth]{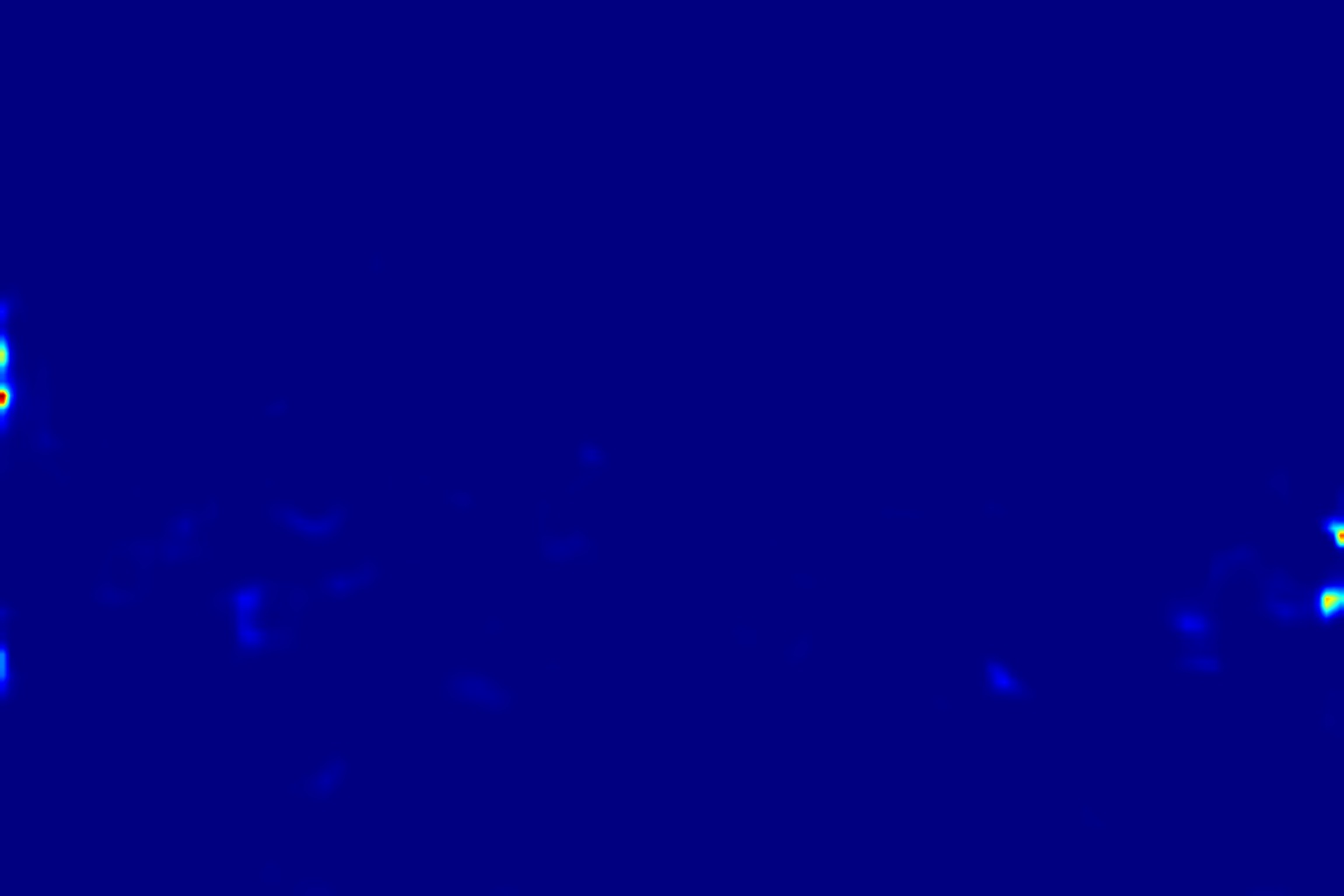} &
			\includegraphics[width=0.14\linewidth]{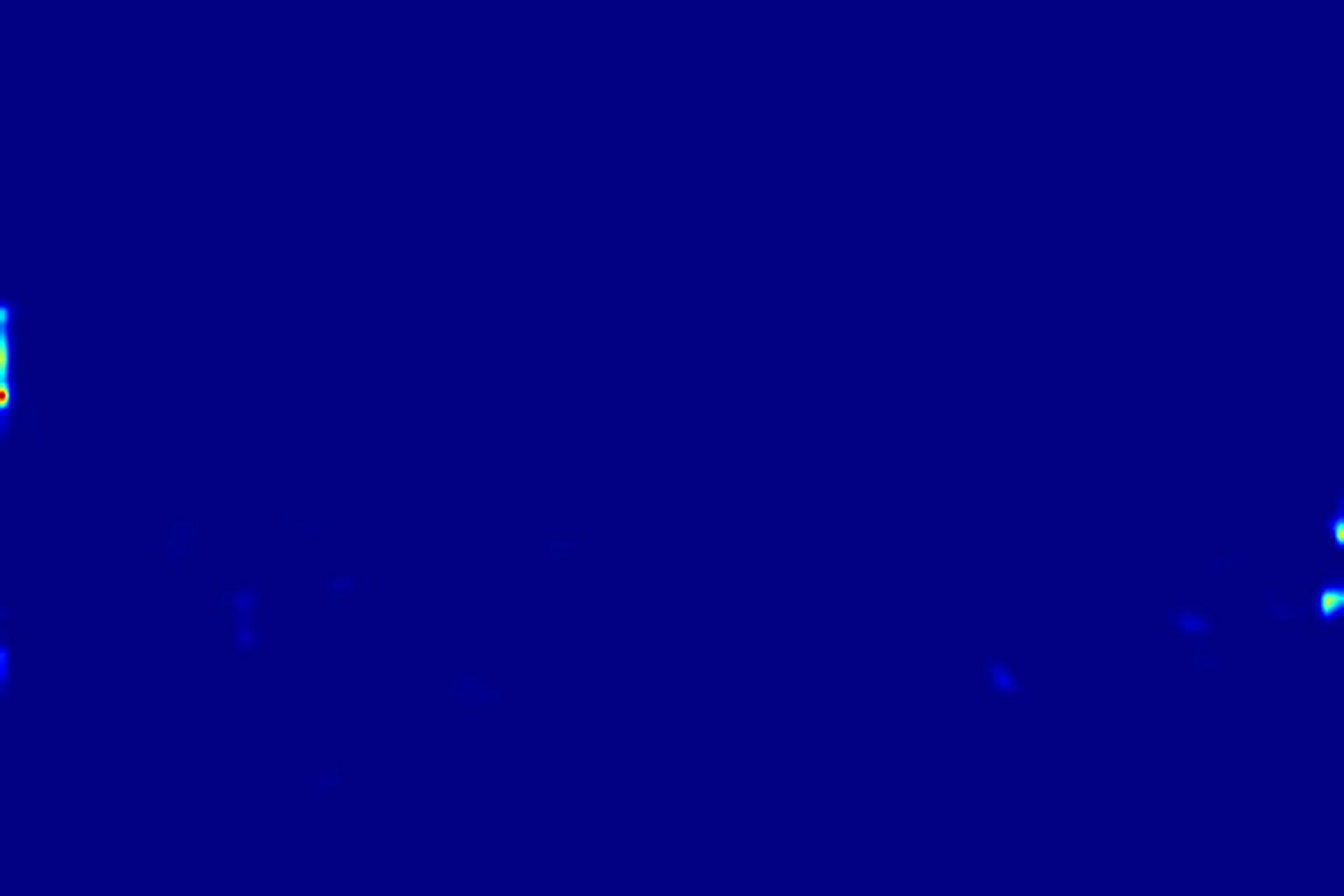} \\
			
			\footnotesize{B17} & \footnotesize{B18} & \footnotesize{B19} & \footnotesize{B20} & \footnotesize{B21} & \footnotesize{B22} \\
			
			\includegraphics[width=0.14\linewidth]{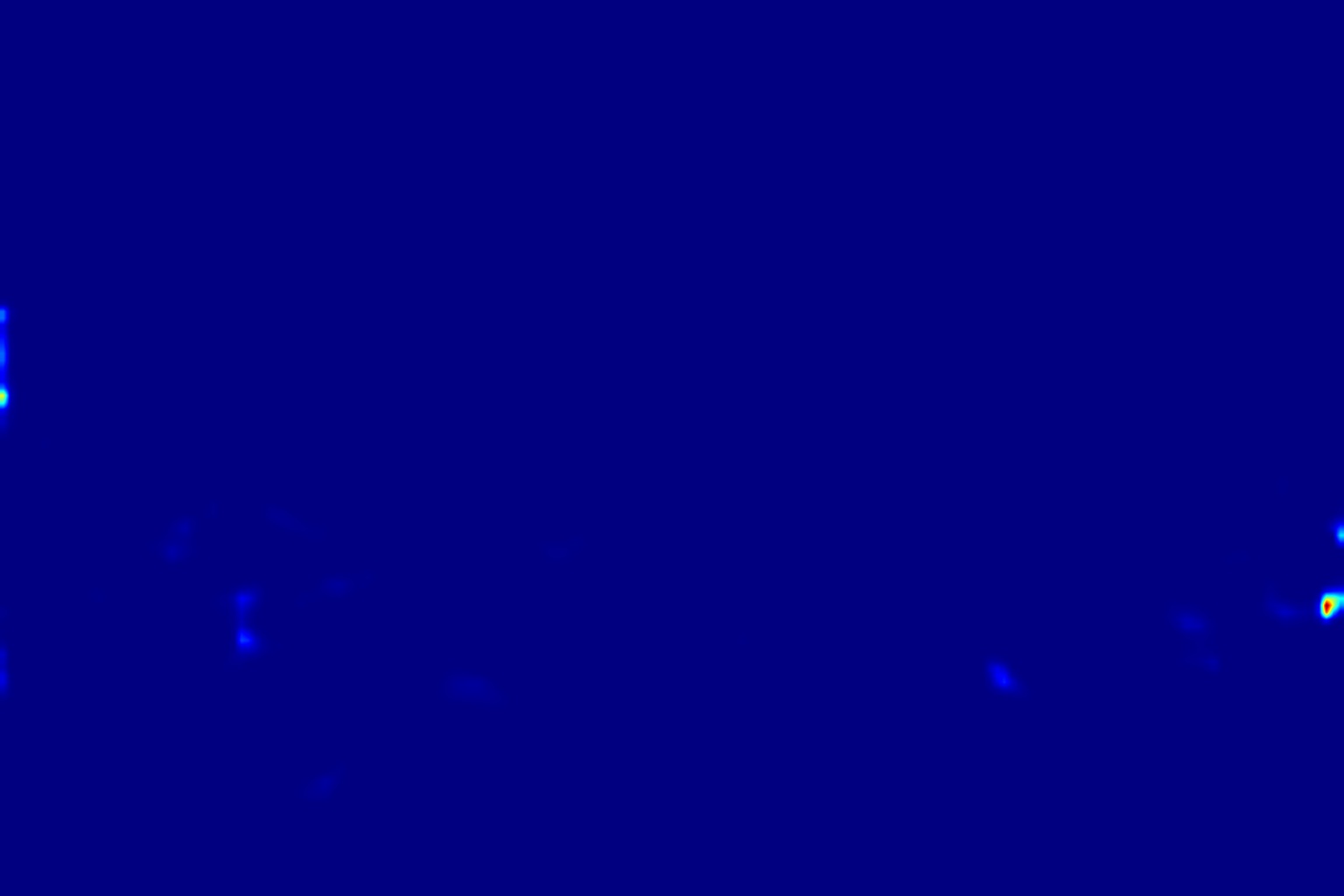}  &
			\includegraphics[width=0.14\linewidth]{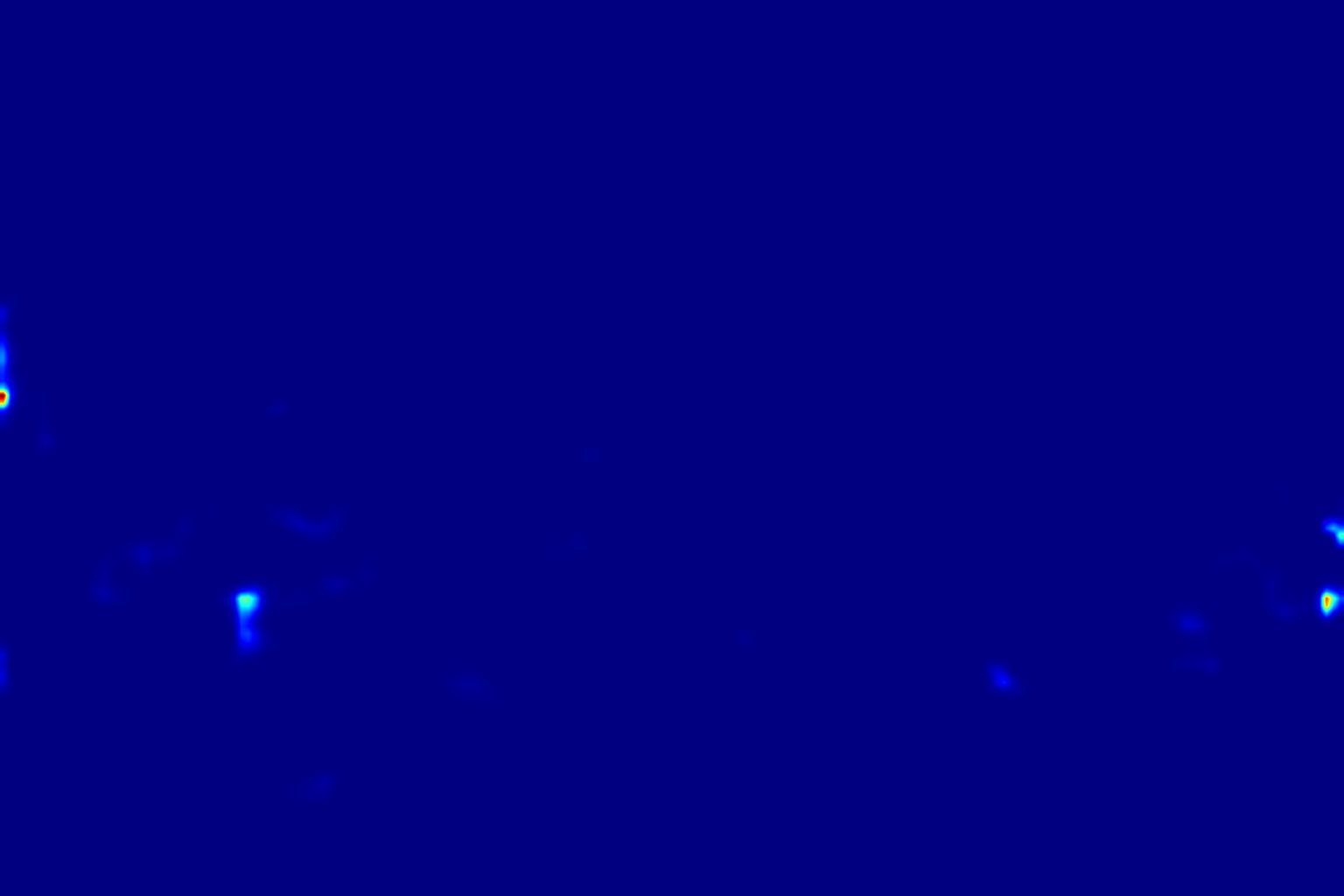}  &
			\includegraphics[width=0.14\linewidth]{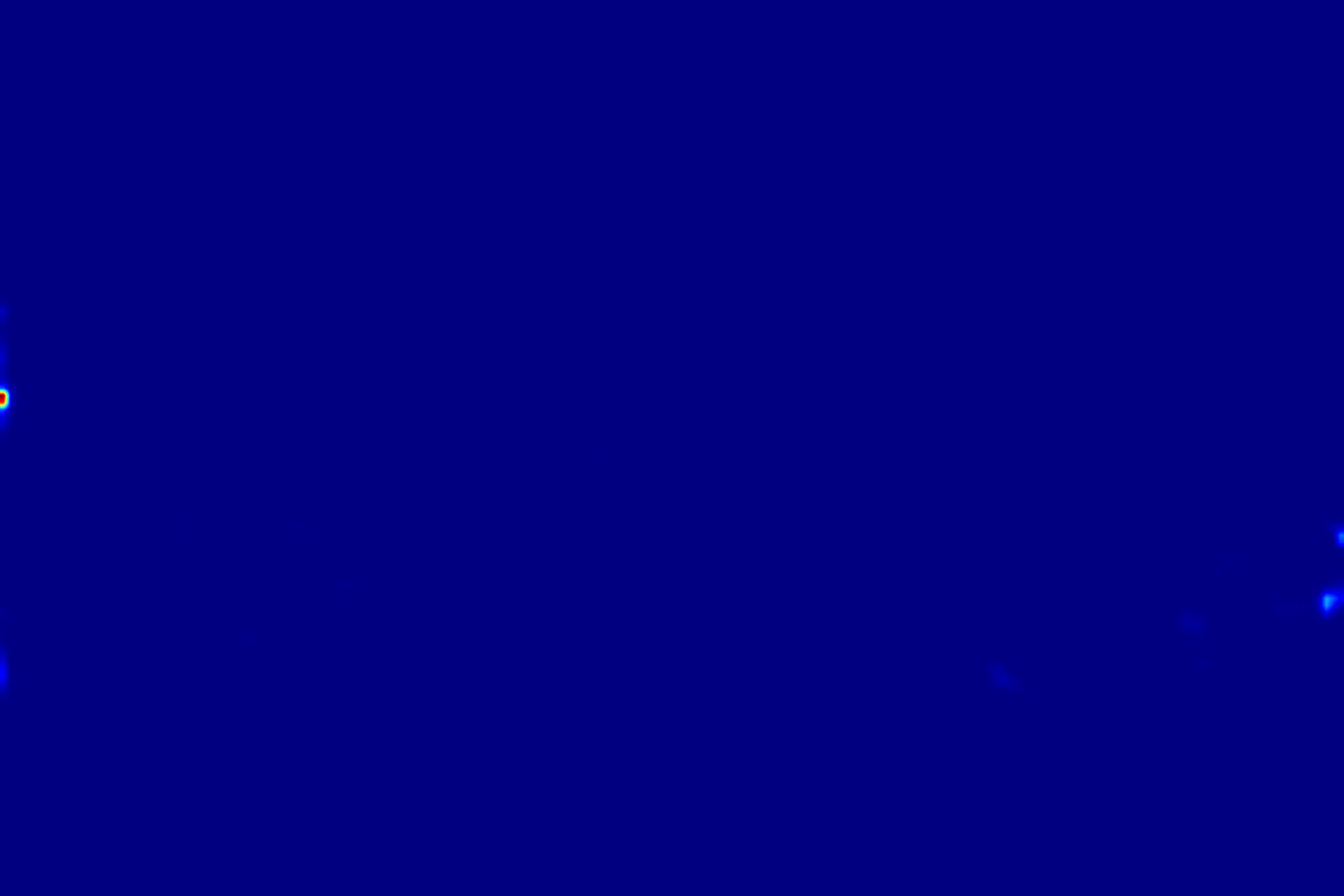} \\
			
			\footnotesize{B23} & \footnotesize{B24} & \footnotesize{B25} \\

		\end{tabular}
		\caption{Attention maps of each density token in dual branch for an unlabeled training image when training with a labeled ratio of $5\%$ on UCF-QNRF. A and B stand for different branch and the numbers represent the different tokens. A token with a higher number specifies the density interval with higher density.}
		\label{fig:viz2}
	\end{center}

\end{figure*}}

\end{document}